\documentclass{article}
\usepackage{jfrExamplee}
\usepackage{apalike}

\usepackage{graphicx} 

\usepackage[ampersand]{easylist}
\usepackage{amsmath} 
\usepackage{amssymb}  
\usepackage{mathtools}
\usepackage{xcolor}
\usepackage[export]{adjustbox}
\usepackage{xspace}
\usepackage{hyperref}
\usepackage{siunitx}
\usepackage{enumitem}

\usepackage{subcaption}
\usepackage{booktabs}

\newcommand{\reffig}[1]{Fig.~\ref{#1}}
\renewcommand{\refeq}[1]{(\ref{#1})}
\newcommand{\reftab}[1]{Table~\ref{#1}}
\newcommand{\refsec}[1]{Section~\ref{#1}}

\newcommand{\etal}{\textit{et al.}\xspace}

\DeclareMathOperator*{\argmin}{\arg\!\min} 

\renewcommand{\vec}[1]{\mathbf{#1}}
\newcommand{\mat}[1]{\mathbf{#1}}

\newcommand{\norm}[1]{\left\lVert #1 \right\rVert}

\renewcommand{\deg}{^{\circ}}

\ListProperties(Space=-0.25cm,Space*=-0.25cm)

\title{An Open-Source System for Vision-Based Micro-Aerial Vehicle Mapping, Planning, and Flight in Cluttered Environments}

\author{
  \begin{minipage}{35em}
    \begin{center}
      Helen Oleynikova\thanks{Work performed while at ETH, now with Microsoft; \texttt{helenoleynikova@gmail.com}}, Christian Lanegger, Zachary Taylor\thanks{Work performed while at ETH, now with Oculus.}, Michael Pantic, Alexander Millane, Roland Siegwart, and Juan Nieto\\
    \end{center}
  \end{minipage} \\
  \\
  Autonomous Systems Lab\\
  ETH Z\"{u}rich, Switzerland \\
  \texttt{firstname.lastname@mavt.ethz.ch} \\
}

\begin{document}
  
\maketitle

\begin{abstract}
We present an open-source system for Micro-Aerial Vehicle autonomous navigation from vision-based sensing.
Our system focuses on dense mapping, safe local planning, and global trajectory generation, especially when using narrow field of view sensors in very cluttered environments.
In addition, details about other necessary parts of the system and special considerations for applications in real-world scenarios are presented.
We focus our experiments on evaluating global planning, path smoothing, and local planning methods on real maps made on MAVs in realistic search and rescue and industrial inspection scenarios. We also perform thousands of simulations in cluttered synthetic environments, and finally validate the complete system in real-world experiments.
\end{abstract}

\section{Introduction}
Autonomous navigation from on-board sensing is essential for Micro-Aerial Vehicles (MAVs) in many applications.
In this work, we specifically target applications where MAVs can assist human operators in difficult tasks, such as search and rescue (S\&R) and industrial inspection applications.
Core challenges in these areas include lack of GPS or other external sensing, unreliable communication (forcing the MAV to be self-sufficient without off-board processing), and having to operate close to structure with comparatively small platforms, limiting payload and sensing capabilities.

Two specific examples of applications we would like to address are inspection of earthquake-damaged buildings and structures, and of repeated inspection in industrial scenarios.
First, in the disaster relief case, there are no prior maps available of a space, as the structure of the space may change entirely.
Currently, firefighters and relief workers either need to enter a building themselves to inspect its structural integrity and whether any victims were trapped inside, or send a ground robot, which is often unable to overcome rubble.
An MAV is a perfect robot for this scenario, but it needs to be able to navigate safely in unknown space, and very close to obstacles.
Therefore a significant portion of our work focuses specifically on safety in cluttered environments: where the robot's narrow field-of-view sensors can frequently have obstacles in unknown space.

After an initial inspection of an earthquake-damaged structure, the MAV may be needed to perform more follow-up inspections to find missing persons, deliver (small) supplies, or explore more of the area than a single battery charge allows.
This would require the MAV to be able to use previously-built maps to globally plan through a space and return to its previous location.
Several of the maps used for evaluation and experiments (shed and rubble) are created at a simulated disaster site to validate these cases.

The second example, industrial inspection, requires a very similar set of capabilities.
It is more likely that the environment is previously mapped, but may contain significant changes and unmapped obstacles over time.
We focus some of our industrial inspection experiments specifically on these scenarios: being able to detect and avoid new obstacles not present in a global map, even if using a global map to plan.

Aiming to present solutions for those challenges, this paper introduces an open-source system capable of performing autonomous online planning and mapping.
Our previous work proposes dense mapping~\cite{oleynikova2017voxblox} and online replanning~\cite{oleynikova2016continuous-time,oleynikova2018safe} methods for safe navigation of previously-unexplored spaces.
We focus on explicitly mapping free space in very cluttered environments, and exploring planning strategies that are inherently conservative: that is, they only allow traversal of space that is confirmed to be free.
Here we aim to extend our local planning work to also cover global planning scenarios, where a map is already available (either on the return route of the current mission or from a previous mission), and explore interactions between local and global planning frameworks.

Furthermore, we compare various global planning methods, including improving on our previously-proposed topological graphs built from Euclidean Signed Distance Fields (ESDFs)~\cite{oleynikova2018sparse}.
We also extend our local replanner to be used as a path-smoothing method for converting lists of waypoints from global planners to dynamically-feasible timed trajectories.
All global planning methods are evaluated on three realistic scenarios, two from a search and rescue training area, and one from an industrial environment with large machinery, recorded with the MAV system described in this paper.

We additionally improve our local planning architecture, and discuss how it fits in from a system perspective. We unify how global plans and single waypoints are handled and tracked, improve the performance of the local planner when goals are in unknown space, and validate the complete framework both in thousands of simulated scenarios and in real-world experiments.

The aim of this work is to introduce an open-source \textit{complete system} dealing with the aforementioned challenges.
This set-up allows the robot to be robust to loss of communication (which is typical in for example S\&R scenarios), and behave intelligently and safely even when not under direct control of a human.
We discuss the control, state estimation, sensor, and hardware concerns and requirements for such systems, and make the software for all parts available open-source.

\begin{figure}[tbp]
  \centering
  \includegraphics[width=0.6\columnwidth,trim=0 0 0 0 mm, clip=true]{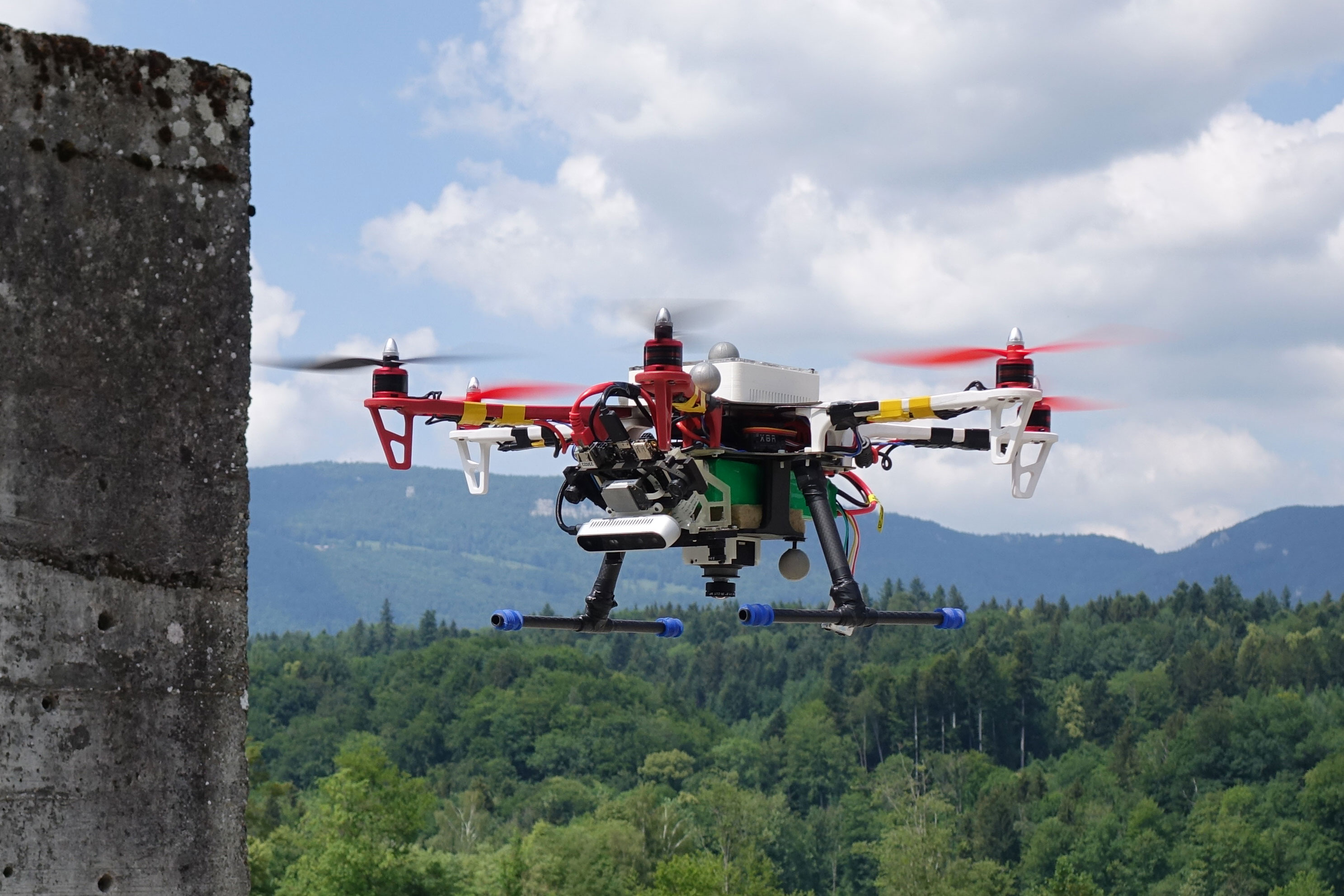}
  \caption{The platform used for collecting the test datasets, a custom-built drone using the DJI FlameWheel F550 frame, an Intel NUC for on-board processing, Pixhawk for flight control, VI Sensor for monocular state estimation and stereo depth, and an Intel Realsense D415 for RGB-D input.}
  \label{fig:jay_system}
\end{figure}

In summary, the main contributions of this work are as follows:

\begin{easylist}[itemize]
  \ListProperties(Space=-0.25cm,Space*=-0.25cm)
  & A complete open-source system for autonomous GPS-denied navigation.
  & A thorough treatment of considerations in 3D mapping for planning applications, expanding on our previous work~\cite{oleynikova2017voxblox} with regards to unknown space and generation methods.
  & We extend our previous work on topological sparse graph generation~\cite{oleynikova2018sparse} to create more useful graphs, faster.
  & A discussion and comparison of various global planning methods on real maps.
  & We extend our previous work in local planning~\cite{oleynikova2016continuous-time,oleynikova2018safe} to also be usable for path smoothing and compare several methods.
  & Finally, we unify our previously-proposed local planner architecture to create a system that can both smooth global trajectories, locally explore in unknown space, and deal with changes in a map by actively replanning.
\end{easylist}

\section{Related Work}

In this section we will mostly focus in reviewing previous work presenting similar complete systems. For a more thorough discussion of state of the art in the individual components we refer the reader to our previous work~\cite{oleynikova2016continuous-time,oleynikova2017voxblox,oleynikova2018safe,oleynikova2018sparse}, and more detailed comparisons to other works are offered in each individual section of the system.

We aim to show a complete system for mapping and planning on-board an autonomous UAV, using vision-based sensing.
Lin \etal \cite{lin2017autonomous} presented a similar complete system, spanning visual-inertial state estimation, local replanning, and control.
However, there are a few key differences between the frameworks proposed: ours focuses strongly on the map representation we use and exploiting all the information within, while theirs uses a standard occupancy map.
More importantly, our planning is \textit{conservative}, meaning we will only traverse known free space, while theirs assumes unknown space is free.
Therefore, we must make more considerations about the contents of our map with our restrictive assumptions.
We also offer an evaluation of global and path smoothing planning methods.
The motivation for these differences are the scenarios we consider: flying very close to structure in cluttered environments in earthquake-damaged buildings and inside industrial spaces.
Treating unknown space as free, as in Lin's work, is an excellent solution in environments where obstacles are sparse or the sensor field of view covers most of the environment, but is a safety issue in our applications.

Mohta \etal \cite{mohta2018fast} also propose an autonomous system for fast UAV flight through cluttered environments.
There are also a few key differences with their work, especially on mapping and planning.
They use a LIDAR as the main sensor, which gives 360$\deg$ field of view to collision detection, removing many of the issues with narrow field of view sensors which we attempt to address in this work.
They also only keep a small local 3D map, and use a global 2D map to escape local minima, whereas we use a full global 3D approach at comparable computation speeds.
For how the mapping is used, they attempt to break the world into overlapping convex free-space regions, which grows in complexity and is increasingly more limited as the space gets more cluttered, while we always plan directly in our map representation.
They also make no considerations for how drift will affect the map other than to keep only a local 3D map.

Finally, the system we propose is conceptually similar to the original system in our previous work~\cite{burri2015real-time}.
The core differences are that we improved every individual component, designed and evaluated a custom mapping system, and proposed a way to do local replanning as well (whereas the previous work was only global planning).
This makes the system proposed in this work much more robust and able to deal with changes in the environment. Additionally, in this work we discuss practical requirements for real-world applications and make every part of the system available open-source.

\section{System Overview}
We describe a complete MAV hardware and software system capable of supporting autonomous flight with only on-board vision-based sensing. 
All of the software described in the system has been made available open-source with provided links.

\subsection{Overall Architecture}

\begin{figure}[tbp]
  \centering
  \includegraphics[width=1.0\columnwidth]{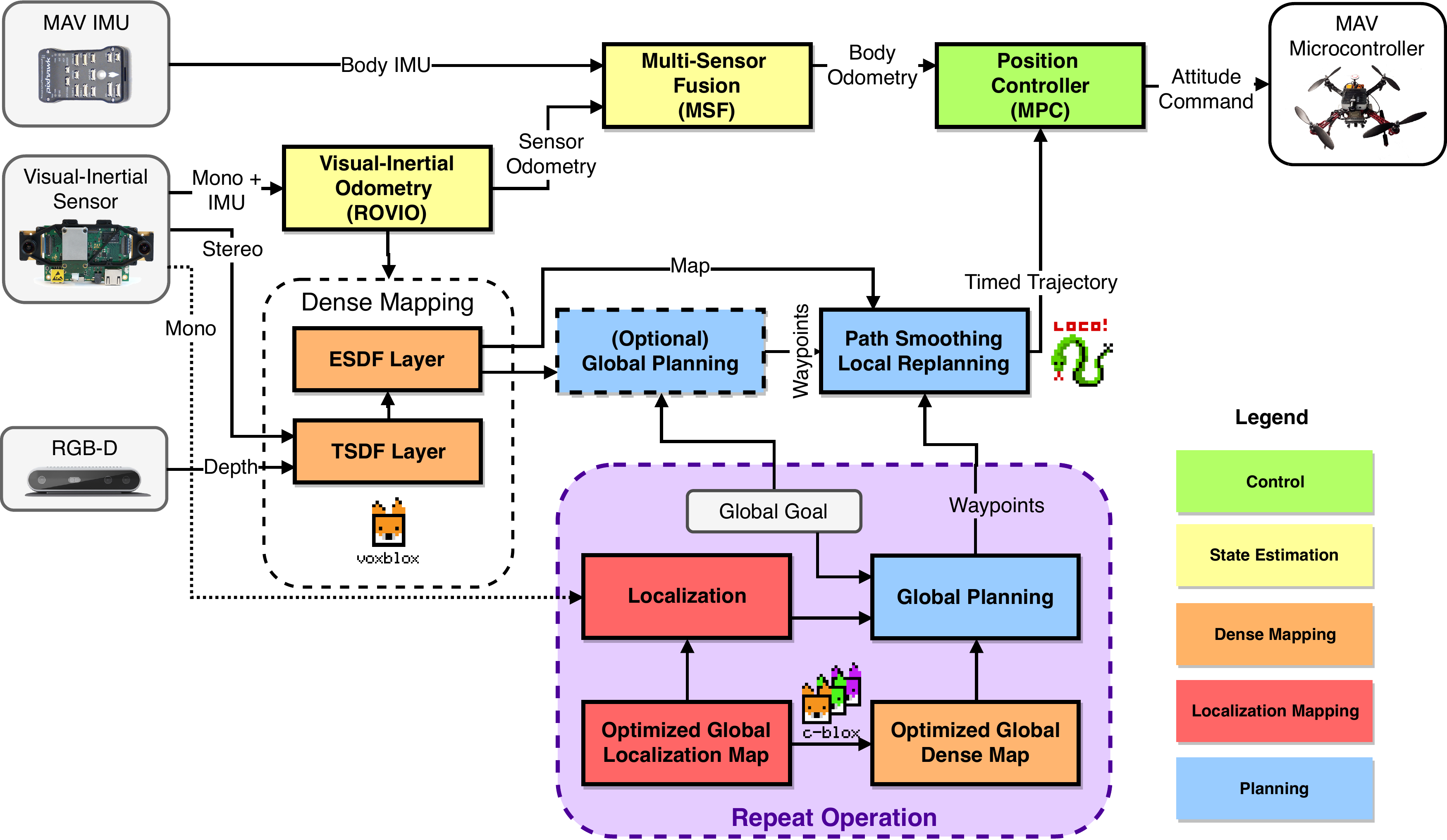}
  \caption{Overall system architecture, showing most key components of the system, and the data flow between especially global and local components. The initial inspection of an environment takes place using local replanning and avoidance through unknown space, and repeated missions use mostly the global planning framework. Though to allow our system to deal with dynamic obstacles, changes in the environment, and potentially drifting maps, the local planner will always collision-check and re-plan any trajectories before execution.}
  \label{fig:system_diagram}
\end{figure}

We show an overview of the complete system in \reffig{fig:system_diagram}, focusing on the data flow between mapping and planning processes.
Stereo and depth images can be used interchangeably for the mapping, combined with a pose estimate from visual-inertial monocular odometry.
The odometry is then fused with the body IMU of the MAV to create a high-rate pose estimate used for control.
Position control runs on the on-board computer, and gives roll, pitch, yaw-rate, and thrust commands to the attitude controller running on the flight controller.
The output of the planning stack are timed trajectories, sampled at the position controller rate (typically 100 Hz).
Since we use a model-predictive controller (described in more detail below), this allows us to have a much higher trajectory tracking accuracy due to the long prediction horizon.

The diagram shows operation set up over two stages: the first is online flight through completely or partially unexplored space, in white, and in purple global planning in previously-built maps.
This system allows us to do an initial flight, optimize a map using off-line tools, then perform repeat inspections in the same area.
Using a local planner on the output of the global planners guarantees that even if the environment changes between missions, we are able to safely navigate through it by replanning locally.
Likewise, we can use the same tools without stopping to optimize a map -- for instance, using a global planner to return home at the end of a mission.

\subsection{Hardware}
\label{sec:hardware}

\reffig{fig:jay_system} shows the MAV used to collect the evaluation datasets and test the overall system.
It is built around a DJI FlameWheel F550, with 6 motors for actuation.
The low-level control is performed with a pixhawk\footnote{\url{http://pixhawk.org}}, using custom firmware that accepts attitude and yaw-rate commands\footnote{\url{http://github.com/ethz-asl/ethzasl_mav_px4}}.
This is necessary as we do not use GPS and fly indoors and in other environments where magnetometer readings are unreliable, therefore giving an absolute yaw reference (as is typical in such systems~\cite{meier2015px4}) is both undesirable and meaningless in our local coordinate frame.

On-board processing, which runs everything shown in \reffig{fig:system_diagram}, is performed on an Intel NUC.
The main sensor is a custom-made visual-inertial sensor~\cite{nikolic2014synchronized}, with two monochromatic cameras in a stereo configuration, hardware-synced to an ADIS448 IMU.
It is used for mono visual-inertial state estimation, and also for stereo depth for mapping.
Optionally, we also use an Intel Realsense D415 for an additional source of RGB-D data.

Though a dedicated visual-inertial sensor is a nice-to-have for such platforms, there are only a few available the shelf that is suitable for MAV flight, such as the Intel T265\footnote{\url{http://www.intelrealsense.com/tracking-camera-t265/}}.
However, the Intel sensor and similar sensors generally do not offer any hardware synchronization abilities to other sensors or to flight computers, which may be necessary for some applications or multi-camera set-ups.
Instead, we recommend using a USB machine vision camera, and hardware time-synchronizing it to a flight controller.
We make a sample driver implementing this for the FLIR Blackfly or Chameleon 3\footnote{\url{http://ptgrey.com/blackfly-usb3-vision-cameras}} and the pixhawk available\footnote{\url{http://github.com/ethz-asl/flir_camera_driver}}.
We make our instructions for setting up such a system and considerations with regard to timing accuracy available online\footnote{\url{http://github.com/ethz-asl/mav_tools_public/wiki/Visual-Inertial-Sensors}}.
This set-up is also extendable to multi-camera systems, as multiple cameras can be triggered from the same pulse.

\subsection{Control}
We use a cascaded control architecture, with an inner loop that controls attitude and runs at a minimum of \SI{100}{\hertz} on the MAV autopilot, and an outer position control loop that runs on the on-board computer at \SI{100}{\hertz}.
For the outer loop, we use a non-linear Model Predictive Control (MPC), proposed by Kamel \etal~\cite{kamel2017nonlinear} and available open-source\footnote{\url{http://github.com/ethz-asl/mav_control_rw}}.
Previous work compares the non-linear MPC formulation to a linear MPC, finding that the non-linear MPC is actually faster to compute and performs better than the linear approximation~\cite{kamel2017linear}, and a detailed description of how to implement and tune both for an arbitrary MAV is available as part of a recent ROS book~\cite{kamel2017model}.

The MPC takes in the body odometry estimates from Multi-Sensor Fusion (MSF, described in \refsec{sec:state_estimation}) and a timed full state trajectory to track.
We exploit the properties of the flat state for MAVs to only need to specify position, yaw, and their derivatives~\cite{mellinger2011minimum}.

One of the advantages of using an MPC over a PID loop for trajectory tracking is that the MPC is able to look ahead at future trajectory points, and minimize tracking error over the complete prediction horizon.
This means that overall trajectory tracking performance is improved significantly, and there are advantages to planning high-fidelity, dynamically-feasible trajectories, as they will be executed almost perfectly.

\begin{figure}[tbp]
  \centering
  \includegraphics[width=0.8\columnwidth]{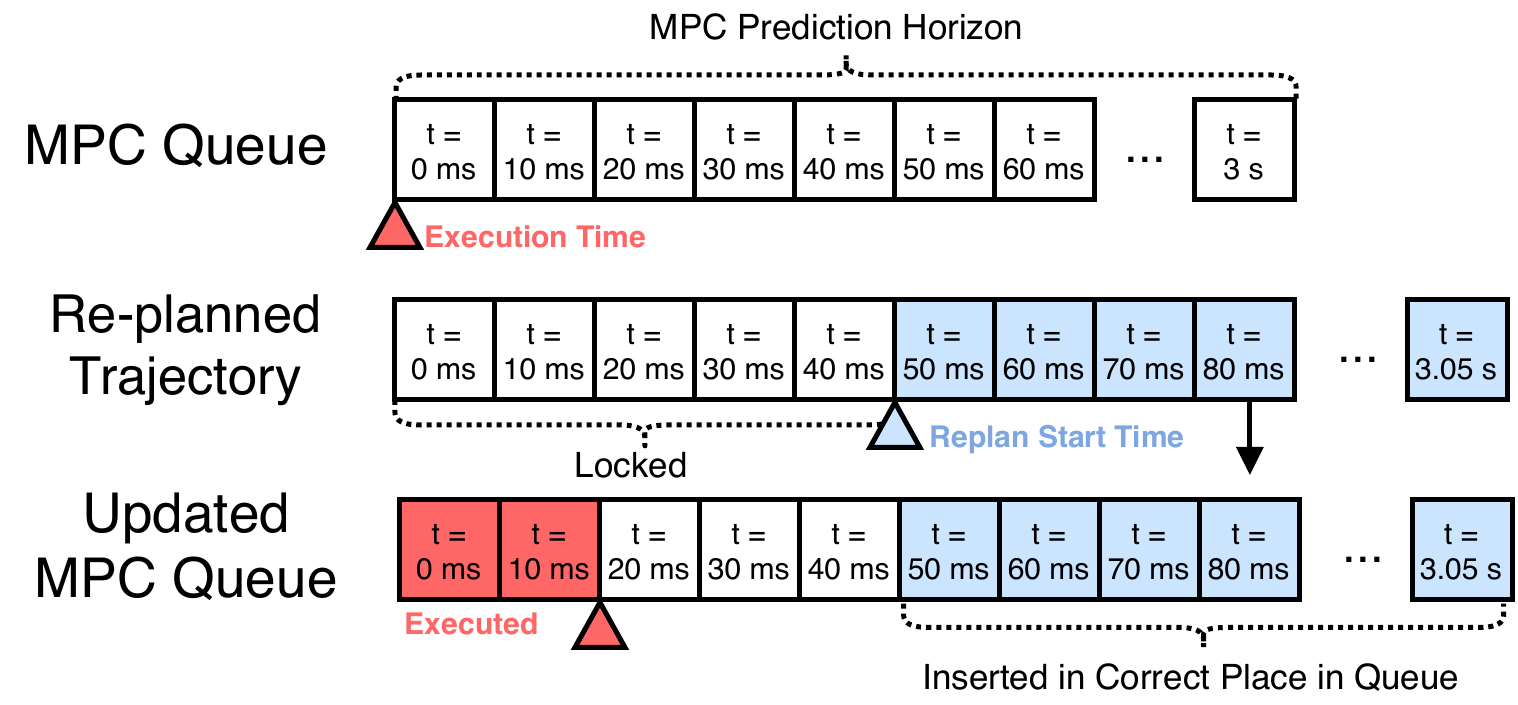}
  \caption{Diagram showing the MPC queue, and how it is updated when a section of the trajectory is replanned. Note that slightly more of the trajectory is locked down than is executed during the planning time. This allows the MPC to not have very abrupt changes in reference. For illustration purposes, the MPC is shown here running at 50 Hz, while it normally runs at 100 Hz on the real platform.}
  \label{fig:mpc_queue}
\end{figure}

The non-linear MPC also has a \textit{very} long horizon of 3 seconds, or 300 timesteps.
While this is very convenient for executing long complex global trajectories, special care must be taken when using it for online replanning.
Namely, we need to timestamp our entire trajectory to be monotonically increasing, and trajectory updates must be inserted into the correct place in the MPC queue.
The queue is cleared if a trajectory with a time before the current execution time is sent, and the new trajectory replaces the complete queue.
\reffig{fig:mpc_queue} shows an example of a replan cycle, happening for the purposes of the illustration at \SI{50}{\hertz}.
The MPC queue is initialized with a starting trajectory.
The local replanning ``locks'' the beginning of the initial trajectory, including the first \SI{20}{\milli\second} which is when the controller will receive the updated trajectory, and also another \SI{30}{\milli\second} look-ahead so that the reference does not change too quickly, then replans starting at \SI{50}{\milli\second}.
A 3 second chunk, starting at the \SI{50}{\milli\second}, is then sent to the controller queue, which inserts the updated trajectory at the correct time, even though it has only executed up to \SI{20}{\milli\second}.

This queueing scheme allows us to replan at any given rate, while making sure that the controller always has a reasonable trajectory within its time horizon.

\subsection{State Estimation}
\label{sec:state_estimation}
All state estimation is done on-board and not using external sensing (i.e., no vicon or GPS).
This gives our system flexibility to be used in complex GPS-denied environments.
Our main state estimator is Rovio\footnote{\url{http://github.com/ethz-asl/rovio}}, which is a robust visual-inertial odometry framework~\cite{bloesch2015robust}, \cite{bloesch2017iterated}. 
Rovio is a filter-based estimator, which uses direct photometric error on a small number of patch features in the image.
For our application, this design has a few distinct advantages over more traditional, keypoint-based methods like Okvis~\cite{leutenegger2015keyframe}: a filter with a small state-space (using only 25 features) is fast to compute, even on the on-board CPU, and using direct photometric error makes the method resistant to motion blur.
In our experience, Rovio is comparably accurate to other methods such as Okvis and VINS-mono~\cite{qin2018vins}, but more robust under real conditions.

However, Rovio only gives us the odometry in the \textit{sensor} frame, which for our system is usually the IMU that is part of the visual-inertial (VI) sensor.
Rovio also only outputs updated poses at the camera frame rate.
Depending on the hardware set-up, we have two solutions to receive body-IMU-frame odometry at \SI{100}{\hertz}.

If using a separate VI sensor, then the odometry estimate from Rovio must be transformed into the body frame and fused with the body IMU.
This is done using Multi-Sensor Fusion (MSF)~\cite{lynen2013robust}, which is a highly-configurable filter capable of taking multiple sensors and pose sources\footnote{\url{http://github.com/ethz-asl/ethzasl_msf}}.

In another configuration, where the only IMU on the system is hardware time-synchronized to a camera, no transformation or fusion needs to be done.
We only need to use the estimated Rovio IMU biases to propagate the odometry estimate using incoming IMU measurements.
This is done using a package called \texttt{odom\_predictor}, which queues incoming IMU measurements and re-applies them when new (delayed) estimates are available from Rovio.
It is also available open-source\footnote{\url{http://github.com/ethz-asl/odom_predictor}}.

\subsection{Localization}
This paper aims to address issues with creating dynamically-feasible global plans.
However, global planning requires \textit{global localization}.
Since all visual- and visual-inertial odometry frameworks drift, no matter how little, long-term operation or operation in previously-explored environments requires the ability to perform loop-closures.

To localize against a global map, we use maplab~\cite{schneider2018maplab}, an  open-source\footnote{\url{http://github.com/ethz-asl/maplab}} framework for creating, storing, optimizing, and localizing in visual-inertial maps.

To create global maps, we first generate a sparse pose-graph of landmarks in the observed scene.
This is done by using Rovioli~\cite{schneider2018maplab}, a front-end for Rovio that also does feature tracking and extraction (independently of Rovio's 25 tracked patch features).

This sparse graph can then be loop-closed and optimized using bundle adjustment in an offline process, giving optimized, globally-consistent poses for all keyframes.
To generate the global map, we can then replay all pointclouds from the initial flight and integrate them into a dense map using optimized poses.
Localization in the matching sparse map will then line up correctly with the optimized dense map.

If using ORB-SLAM~\cite{mur2015orb} as the SLAM system, a better approach is to build up a dense map using submaps, and only fuse them when the covariance between their relative poses is small enough, as presented in our previous work called \textit{c-blox}~\cite{millane2018c-blox}.
This allows us to get a globally-optimal dense map in a single step, without having to run a recorded dataset through an offline framework.
However, since ORB-SLAM does not support re-localization against a previous map, this limits us to global localization within a single mission.

\section{Dense Mapping}
\label{sec:dense}
\begin{figure}[tbp]
  \centering
  \includegraphics[width=0.8\columnwidth]{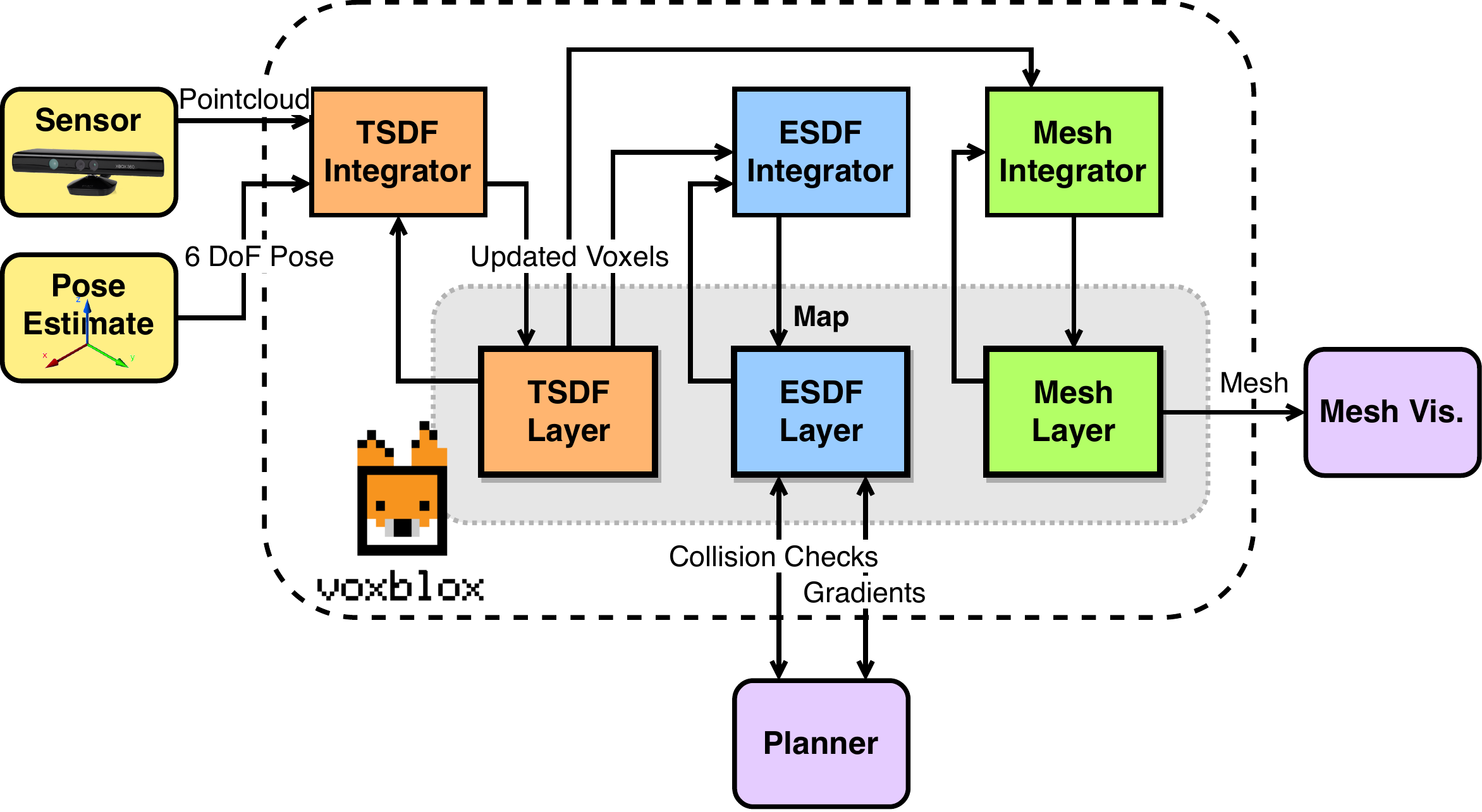}
  \caption{Voxblox system diagram, showing how the TSDF and ESDF layers are interconnected through integrators.}
  \label{fig:voxblox_system}
\end{figure}

Dense mapping is key to planning performance, as a plan can only be as good as the map.
We use a flexible mapping framework called \textit{voxblox}\footnote{\url{http://github.com/ethz-asl/voxblox}}, introduced in our previous work~\cite{oleynikova2017voxblox}.
The framework is centered around using Signed Distance Fields (SDFs), or voxel grids of distance values to surfaces.
We use two different types of SDFs: Truncated Signed Distance Fields (TSDFs), based on Curless and Levoy~\cite{curless1996volumetric} and KinectFusion~\cite{newcombe2011kinectfusion} for integrating point cloud data, in a method that gives a more accurate surface estimate than occupancy-based methods, but uses projective distances and truncates the value to a small band around surface crossings.
The second type of field is a Euclidean Signed Distance Field (ESDF), which store Euclidean, rather than projective distance to each obstacle, and are not truncated to a specific range.
These ESDFs are then the representation we use for planning, as they contain collision information for the entire map, and can also be used to quickly get obstacle gradients, which will be essential for some of the planning methods below.
A system diagram showing inputs, outputs, and data flow is shown in \reffig{fig:voxblox_system}.

\subsection{Euclidean Signed Distance Fields}
\begin{figure}[tbp]
  \centering
  \begin{subfigure}[b]{0.60\columnwidth}
  \includegraphics[width=1.0\columnwidth,trim=0 0 0 0 mm, clip=true]{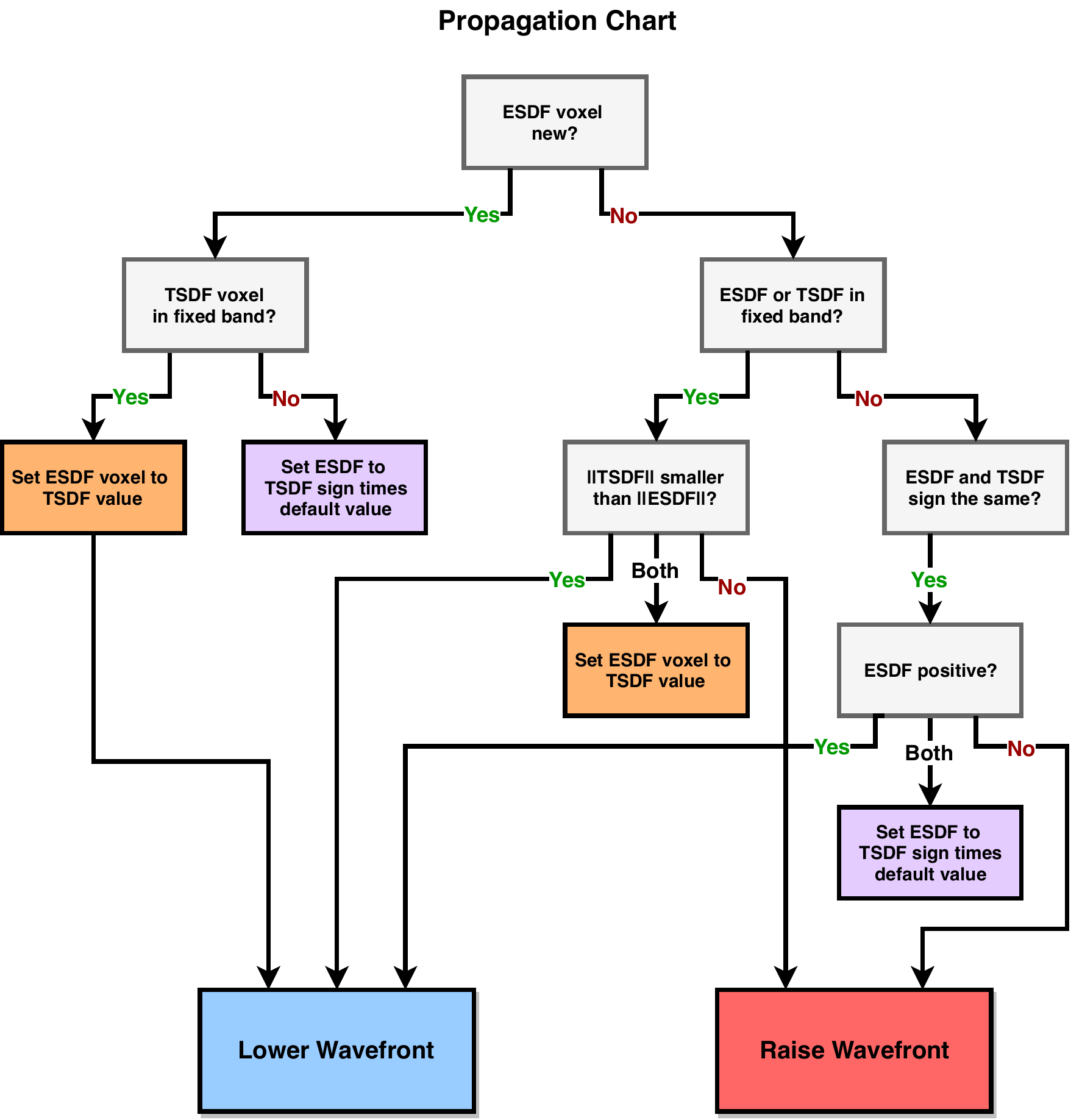}
  \caption{TSDF to ESDF Propagation}
  \label{fig:esdf_prop}
  \end{subfigure}
  \begin{subfigure}[b]{0.25\columnwidth}
  \includegraphics[width=1.0\columnwidth,trim=0 0 0 0 mm, clip=true]{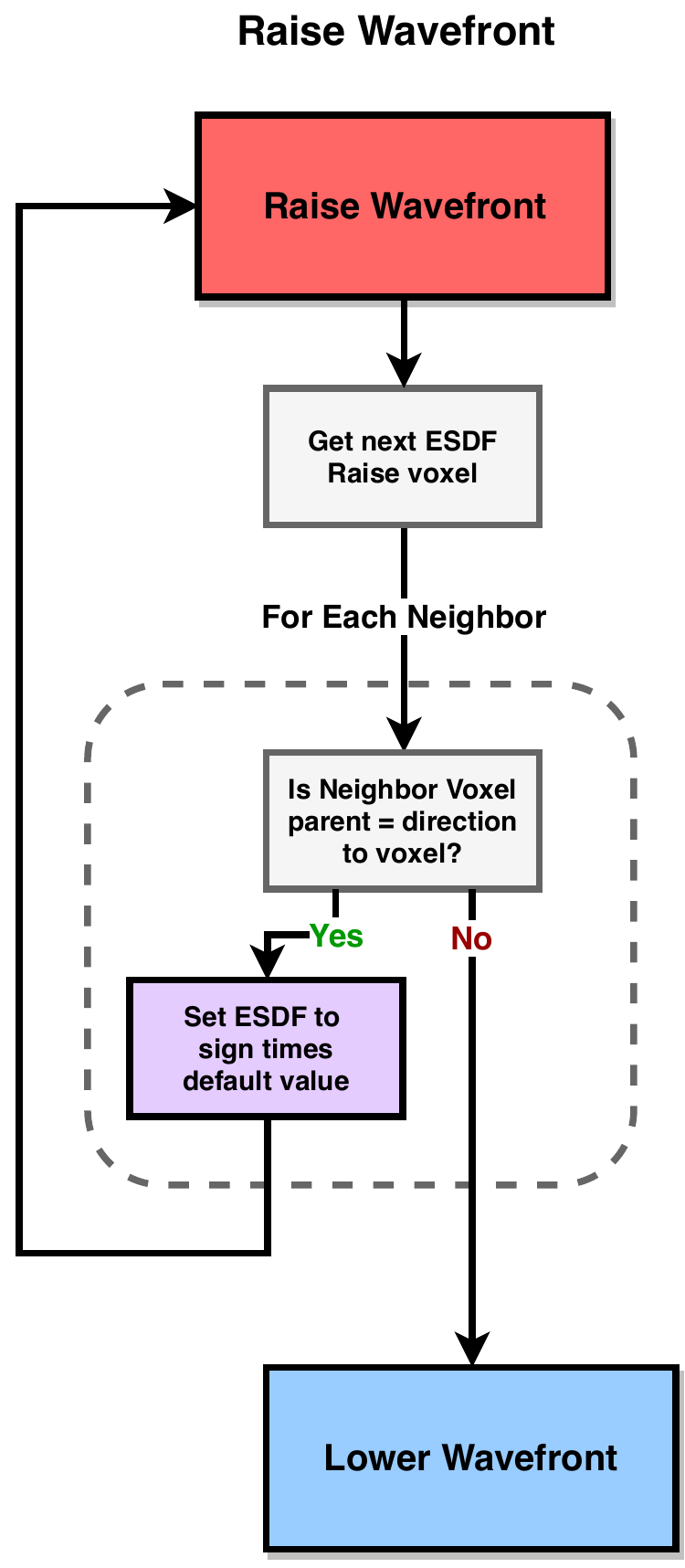}
  \caption{ESDF Raise}
  \label{fig:esdf_raise}
  \end{subfigure}
  \begin{subfigure}[b]{0.50\columnwidth}
  \includegraphics[width=1.0\columnwidth,trim=0 0 0 0 mm, clip=true]{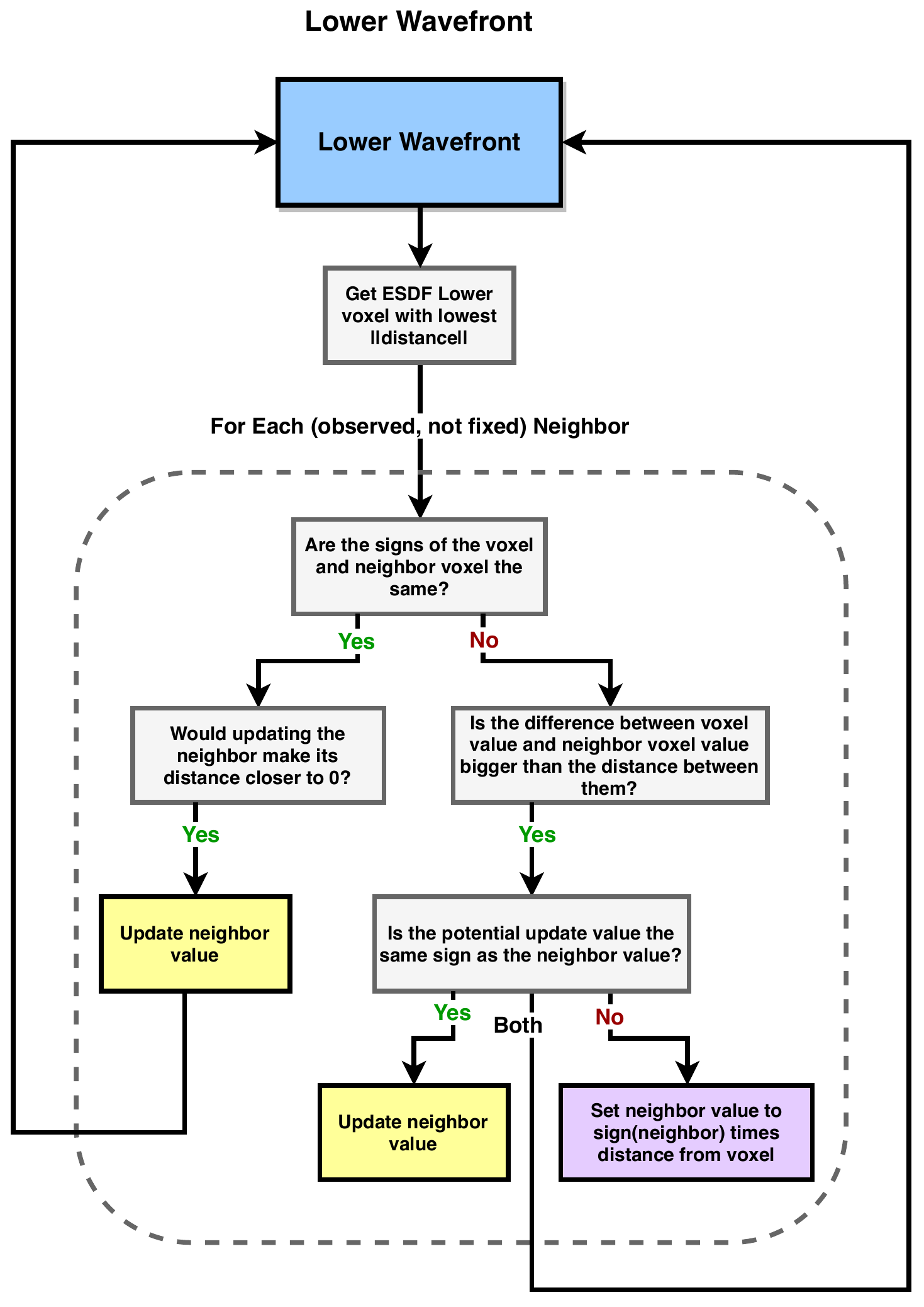}
  \caption{ESDF Lower}
  \label{fig:esdf_lower}
  \end{subfigure}
  \caption{Diagrams explaining how the ESDF is constructed in three stages.}
  \label{fig:esdf_diagrams}
\end{figure}

This section will discuss how an ESDF is computed from a TSDF.
A more thorough analysis, including upper bounds on error introduced by various assumptions and a comparison with occupancy-based methods is offered in our previous work~\cite{oleynikova2017voxblox}.

While it might seem unintuitive that there is a non-trivial process to convert from a TSDF to an ESDF, this is because the distances in both representations are computed differently.
TSDFs use projective distance, or distance along the ray cast from the sensor to the surface.
These distances are fairly accurate near surface crossings, but quickly accumulate large errors~\cite{oleynikova2016signed}.
In contrast, an ESDF needs true Euclidean distances, which can only be calculated in a global fashion.
Luckily, incremental algorithms exist for computing ESDFs from occupancy maps~\cite{lau2010improved}, and our work extends these methods to also work from TSDFs.

Generating the ESDF is done in three stages, detailed in \reffig{fig:esdf_diagrams}: propagation (\subref{fig:esdf_prop}), raise (\subref{fig:esdf_raise}), and lower (\subref{fig:esdf_lower}).
The first, and most different from occupancy-based methods such as \cite{lau2010improved}, is the TSDF to ESDF propagation.
Due to the inaccuracy of TSDF distance estimates, we define a radius called a ``fixed band'' around the surface, which must be at least one voxel size and at most equal to the truncation distance.
TSDF values that fall within this band are considered fixed, copied into the ESDF, and can not be altered in the ESDF update.
Next, updated voxels may simply retain their value, or be put into the ``lower'' wavefront (when their updated distance values are closer to the surface than before) or the ``raise'' wavefront (when their values become farther from the surface).
If performing a batch update (i.e., the entire layer at once), all voxels will go to the lower wavefront.

After propagating all updated values from the TSDF into the ESDF, we then process the raise wavefront.
This consists of simply invalidating all voxels in the wavefront and their children.
Since each voxel stores its ``parent'' (if the voxel is in the fixed band from the TSDF, it is its own parent), this is then an incremental brushfire operation.
All voxels cleared from the raise wavefront have their still-valid neighbors added to the lower wavefront, to guarantee that their values get updated.

The lower wavefront behaves similarly to the occupancy case: iterate over all voxels in the lower wavefront, if their neighbor's distance to the surface can be lowered through the current voxel, then update the neighbor and add it to the lower wavefront.
Special distinctions must be made when implicit zero-crossings exist: i.e., two voxels are neighbors with opposite signs, and neither is fixed.
This case is further explained in \reffig{fig:esdf_lower}.

\subsection{Unknown Space}
\label{sec:unknown}
A key problem with mapping for collision avoidance is deciding how unknown space is handled.
There are two options: treating all unknown space as free (optimistic), and treating all unknown space as occupied (pessimistic or conservative).

While many local collision avoidance works treat unknown space as free and have a high replan rate to avoid collisions~\cite{chen2016online}, this is inherently unsafe.
While it works well in uncluttered environments, where most unknown space \textit{is} free, this assumption gets progressively worse as obstacle density in the environment increases and sensor field of view decreases.

Since our work aims to deal with the worst possible case, which is very obstacle-dense environments and a narrow field-of-view sensor, we cannot adopt the optimistic strategy.
In fact, we always plan to stop in known free space, to guarantee safety in partially-unexplored environments.

However, it is a challenge to encode unknown space information in the ESDF, as the ESDF integration requires TSDF distances to build on, and those are either positive or negative.
There is the additional problem that the MAV has no knowledge of the state of its current position at start-up or take-off, as it has never observed the space it occupies at the start.
Furthermore, if the sensor has a very narrow field of view, the space it perceives in front of the sensor may not be wide enough to fit the entire robot body, essentially paralyzing the robot to never move.
Finally, it is not clear how to correctly treat voxels bordering unknown space in ESDF computation.

\begin{figure}[tbp]
  \centering
  \includegraphics[width=0.35\columnwidth]{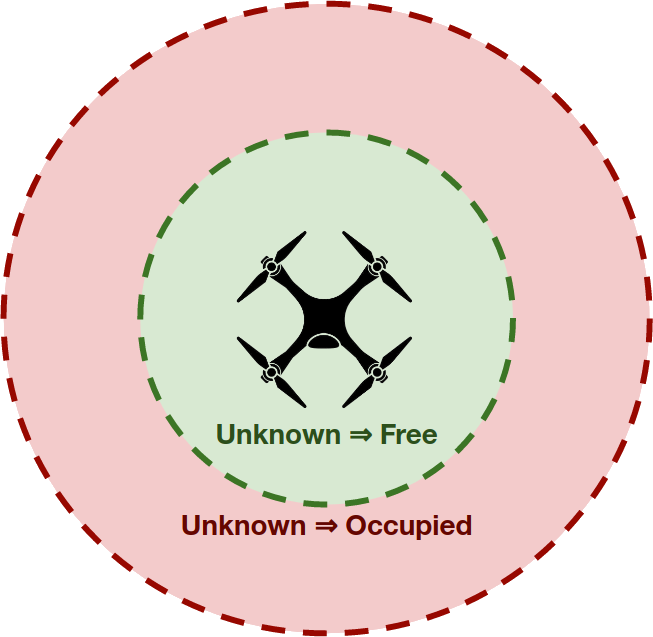}
  \caption{Diagram showing the two radiuses around the current robot pose: a small clear sphere, which should be only slightly larger than the robot, and an occupied sphere, which should be roughly the size of the planning radius.}
  \label{fig:clear_sphere}
\end{figure}

We propose a simple strategy to resolve these issues, originally proposed in our previous work~\cite{oleynikova2018safe}.
The idea is to have two overlapping ``spheres'' centered around the current robot pose that are applied in the ESDF, shown in \reffig{fig:clear_sphere}.
The inner sphere is small and only slightly larger than the robot radius and is called the ``clear sphere'', which sets unknown space within it to free in the ESDF.
The outer sphere affects all points not within the clear sphere and sets all unknown voxels to occupied.
Any voxels that receive distances from this operation are marked as hallucinated in the map, so that as soon as real distance measurements are available from the TSDF, their values are overwritten.

\subsection{Comparison to Other ESDF Construction Methods}
Following our original fixed-volume ESDF construction method from occupancy data~\cite{oleynikova2016continuous-time} for MAV planning in 2016, there has been a significant amount of work on creating ESDFs more efficiently as their usefulness in local replanning scenarios has been recognized.
Most notably, \textit{ewok}~\cite{usenko2017real}, which incrementally computes the ESDF from occupancy in a fixed-size sliding window around the current pose of the MAV, and FIESTA~\cite{han2019fiesta}, which suggests a time-efficient way of performing the global computation in an incrementally-growing data structure from occupancy.

While both proposed methods are faster to compute than \textit{voxblox}, building a TSDF rather than an occupancy map first has some advantages.
The TSDF is able to more accurately model the underlying surface geometry, as it has \textit{sub-voxel} resolution in estimating the zero-crossings of the surface: since each voxel encodes its distance to the surface rather than whether it is occupied or not, methods like marching cubes~\cite{lorensen1987marching} allow us estimate a more exact surface boundary.
What this essentially means it that it is simpler to create more human-readable maps from TSDFs than from occupancy maps: a smooth, sub-voxel resolution mesh is much easier to interpret for a person than large occupancy blocks, especially for an untrained operator.
How important this is varies from application to application, and which intermediate representation is best used to construct an ESDF depends on a variety of project requirements.
However, for our proposed search and rescue scenario, the most critical output of the system is a map that a firefighter or rescue worker can easily interpret and understand.
To this end, we believe that using \textit{voxblox} and building a TSDF first, and then an ESDF out of it is a more appropriate path to take.

\section{Sparse Topology}
\label{sec:sparse_topology}

\begin{figure}[tbp]
  \centering
  \begin{subfigure}[b]{0.25\columnwidth}
    \includegraphics[width=1.0\columnwidth,trim=80 0 140 0 mm, clip=true]{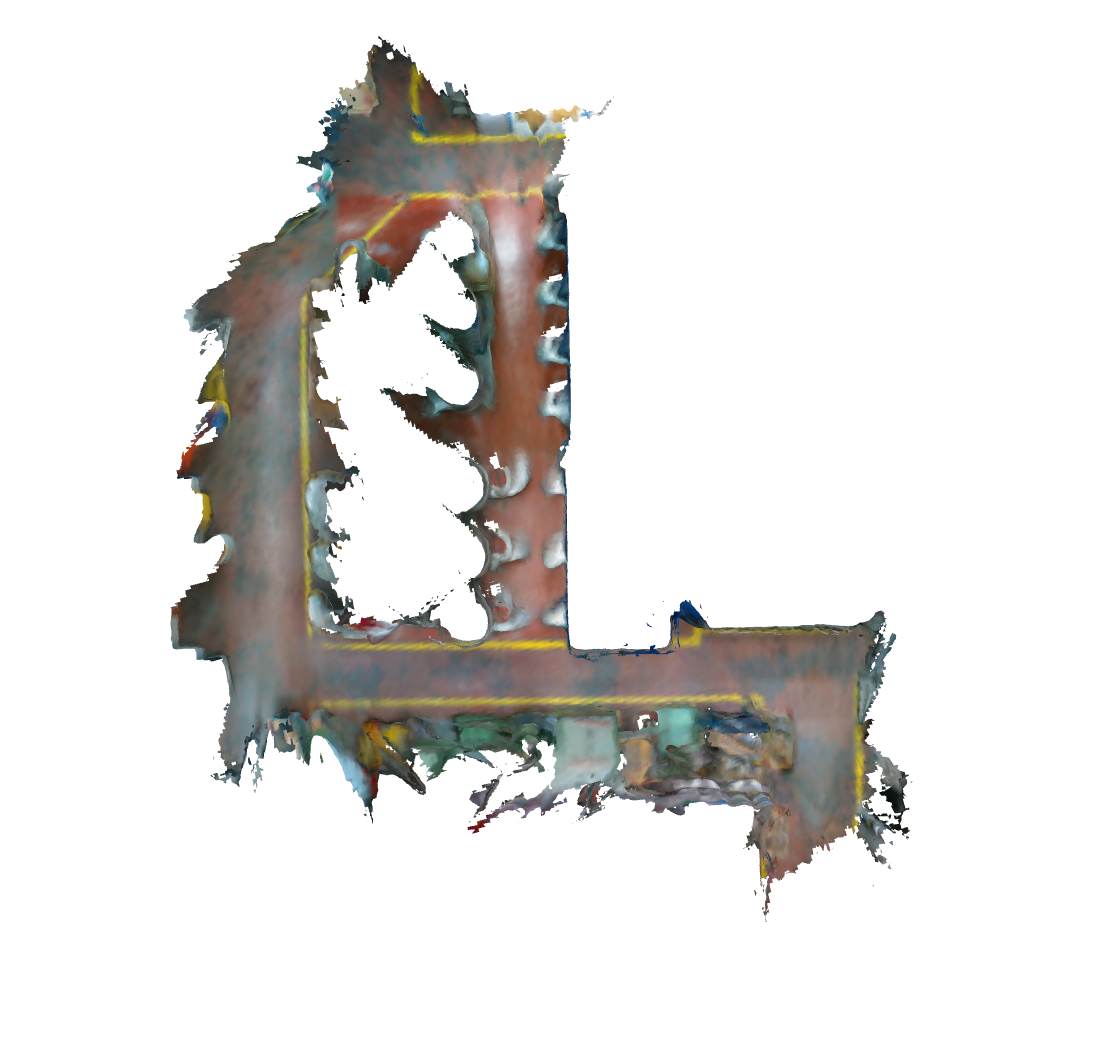}
    \caption{Mesh}
    \label{fig:topo_mesh}
  \end{subfigure}
  \begin{subfigure}[b]{0.25\columnwidth}
    \includegraphics[width=1.0\columnwidth,trim=80 0 140 0 mm, clip=true]{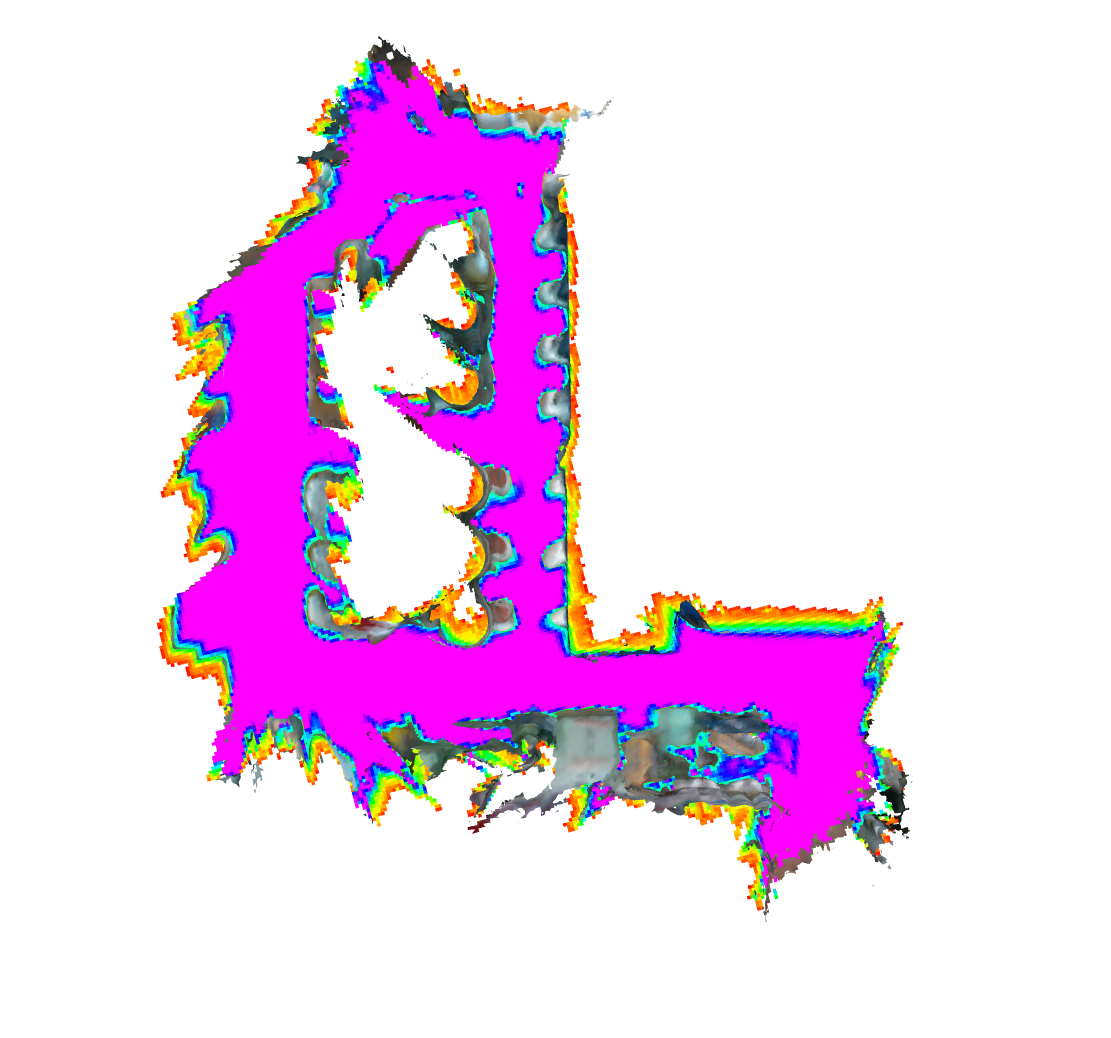}
    \caption{TSDF}
    \label{fig:topo_tsdf}
  \end{subfigure}
  \begin{subfigure}[b]{0.25\columnwidth}
    \includegraphics[width=1.0\columnwidth,trim=80 0 140 0 mm, clip=true]{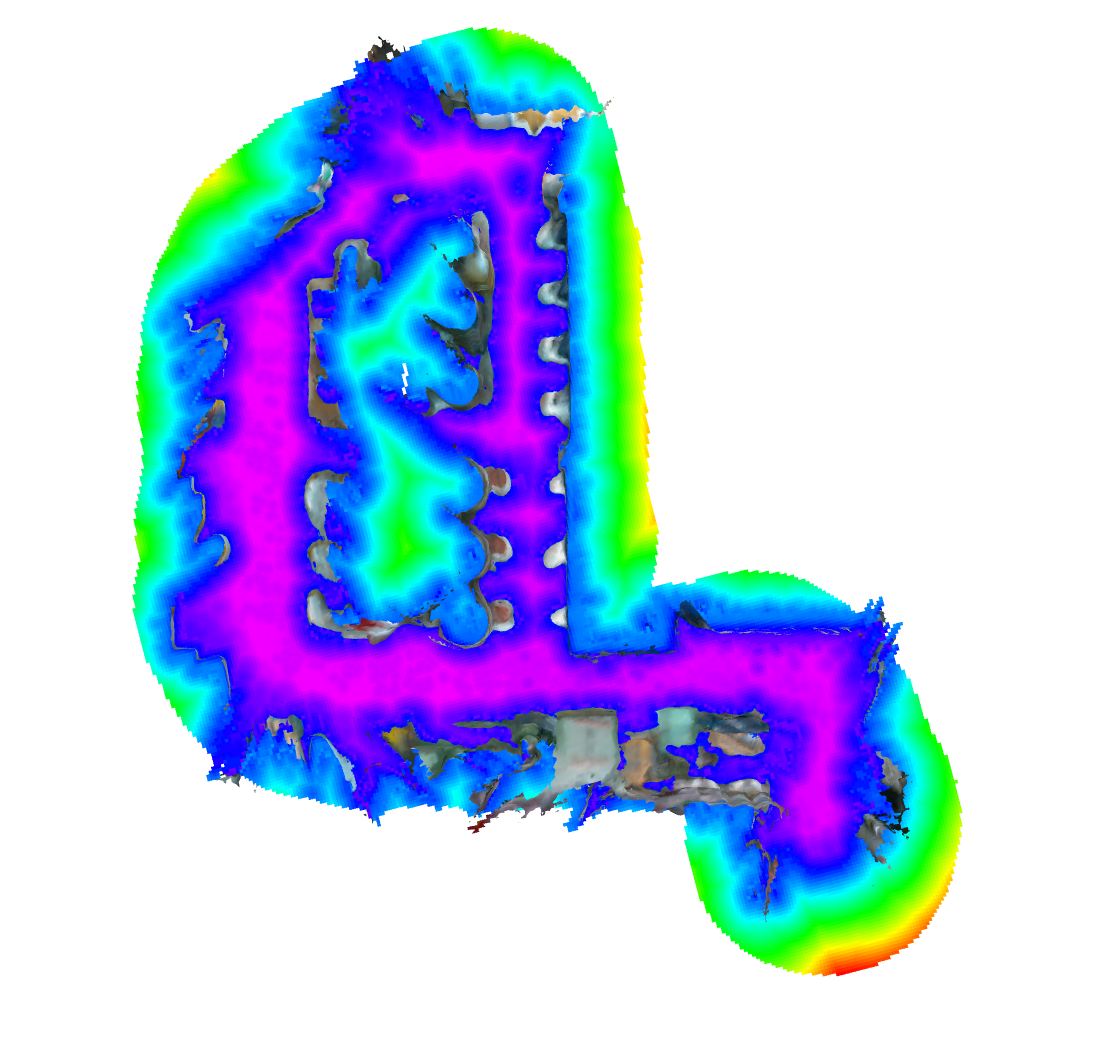}
    \caption{ESDF}
    \label{fig:topo_esdf}
  \end{subfigure}
  \begin{subfigure}[b]{0.25\columnwidth}
    \includegraphics[width=1.0\columnwidth,trim=80 0 140 0 mm, clip=true]{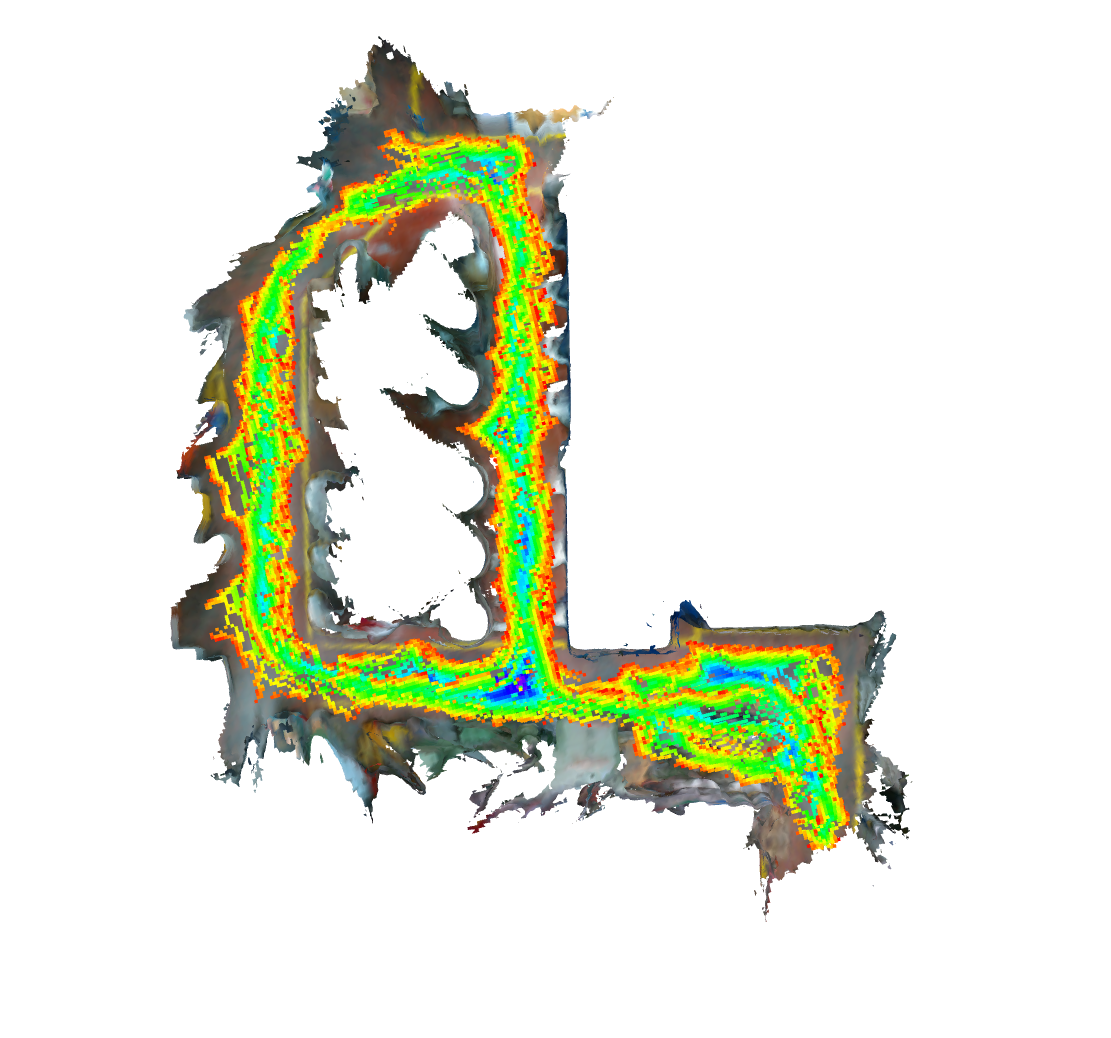}
    \caption{GVD}
    \label{fig:topo_skeleton}
  \end{subfigure}
  \begin{subfigure}[b]{0.25\columnwidth}
    \includegraphics[width=1.0\columnwidth,trim=80 0 140 0 mm, clip=true]{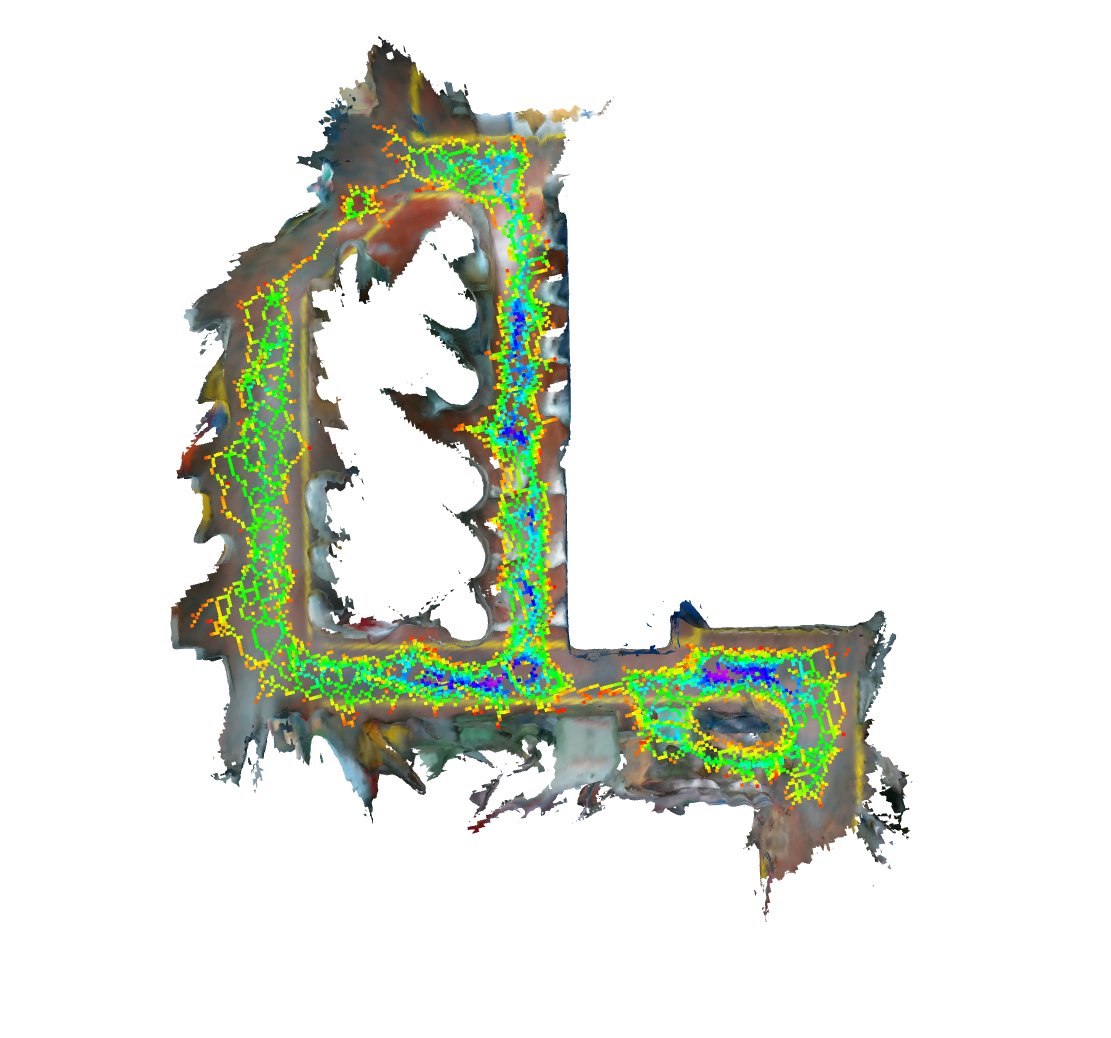}
    \caption{Thin GVD}
    \label{fig:topo_skeleton_thin}
  \end{subfigure}
  \begin{subfigure}[b]{0.25\columnwidth}
    \includegraphics[width=1.0\columnwidth,trim=80 0 140 0 mm, clip=true]{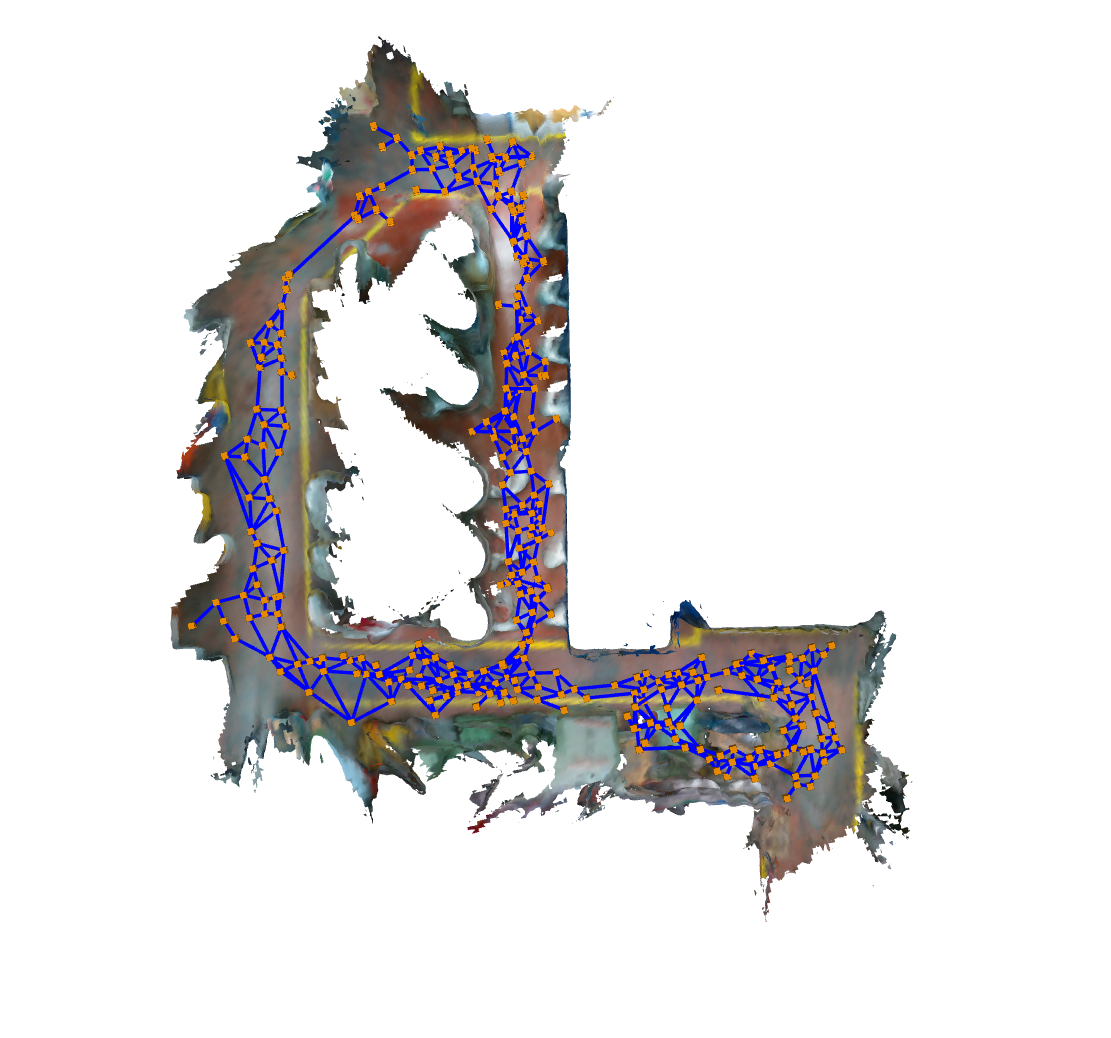}
    \caption{Sparse Graph}
    \label{fig:topo_sparse_graph}
  \end{subfigure}
  \caption{Different stages of sparse topology skeleton computation, shown on the machine hall realsense dataset. Colors represent distances from obstacles.}
  \label{fig:topo}
\end{figure}

In this section, we extend our work on generating sparse topological skeletons from~\cite{oleynikova2018sparse}.
We describe a complete method to extract a sparse graph of the traversable free space in an ESDF from only ESDF map data.
This sparse graph is then used for very fast global planning later sections of this work. 

Our main contributions over our previous work in this section include:
\begin{easylist}[itemize]
  \ListProperties(Space=-0.25cm,Space*=-0.25cm)
  & Switching to a simpler definition of the Generalized Voronoi Diagram (GVD), using 26-connectivity.
  & Extending the method to work for both full and quasi-Euclidean distance.
  & Speeding up sparse graph construction through use of flood-fill operations on the edges.
  & Proposing a new sparse graph simplification and sub-graph re-connection methods, which produce more usable graphs significantly faster.
\end{easylist}

\begin{figure}[tbp]
  \centering
  \includegraphics[width=0.9\columnwidth,trim=0 0 0 0 mm, clip=true]{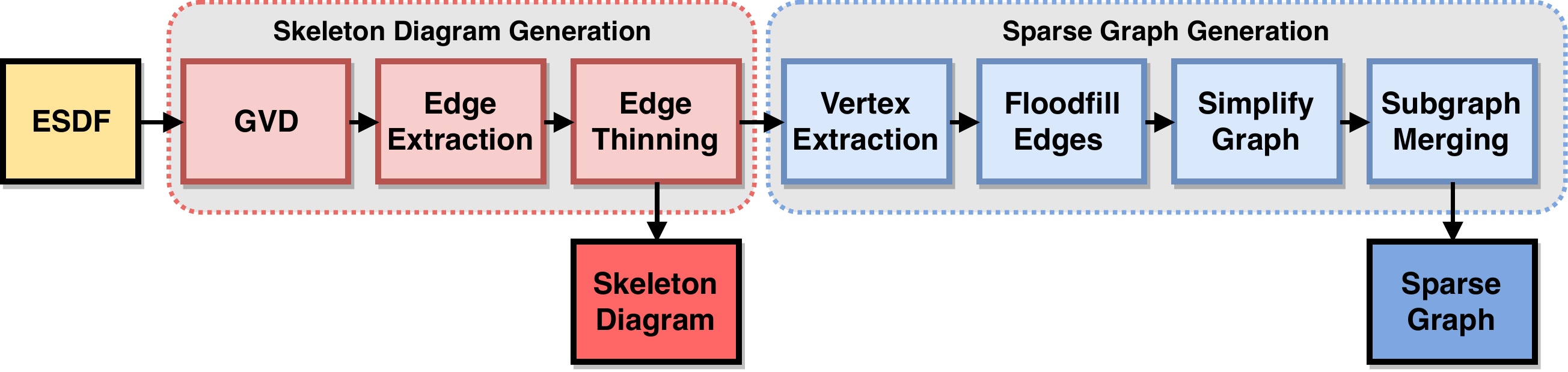}
  \caption{Steps in generating a sparse topological graph, also referred to as a skeleton graph, by first creating a voxelized ``skeleton diagram'' and later fitting a graph that is independent of the voxel size to best fit the structure in a more compact representation.}
  \label{fig:topo_diagram}
\end{figure}

\reffig{fig:topo_diagram} shows the stages in the process, illustrated at key points with \reffig{fig:topo}.

First, the generalized voronoi diagram (GVD) is constructed.
The diagram is a voxelized repersentation of the topology of the space.
In 2D, the GVD contains lines that represent the areas of maximum separation between two or more surfaces, but in 3D, the structure of the shape is much more complex: with vertices, edges, and faces.
We extract the 3D GVD by iterating all voxels in the ESDF to attempt to find ``ridges'' or ``basis points''.
A point or voxel is considered a basis point if its neighbors have parent vectors (that is, vectors that point toward the nearest surface from a voxel) that are at least some angle apart.
The angle differs depending on whether quasi-Euclidean or full Euclidean distance is employed, as quasi-Euclidean has a lower resolution in parent directions (recall that quasi-Euclidean distance approximates the unit circle as an octagon, allowing us to simplify ESDF computation by not distinguishing between small changes in parent directions).
For full Euclidean distance, the separation angle is \SI{45}{\degree}, and for quasi-Euclidean it is \SI{90}{\degree}.

We use a different definition  of what belongs on the GVD from our previous work, making the definition simpler and more physically meaningful.
Before we used 6-connectivity when evaluating belonging on the GVD, but now we use full 26-connectivity for all parts of the process.
A point is considered a GVD face if it has 9 or more neighbors that are basis points, an edge if it has 12 or more, and a vertex at 16 or more.
For the purposes of the remaining method, we do not consider faces, as the goal is to build as sparse of a diagram as possible.
An example of the edges and vertices of a GVD built through this process from a mesh in \reffig{fig:topo_mesh} is shown in \reffig{fig:topo_skeleton}.

Next, to create a sparse diagram (which is simply a voxel layer containing number of basis point neighbors and whether the point is an edge or vertex or not), we apply a series of thinning operations described in more detail in \cite{oleynikova2018sparse}.
While an ideal GVD would already be ``thin", in practice noisy input data (as from on-board sensors) instead of perfectly smooth CAD data creates many aliasing and doubling effects in the diagram, and is a well-studied problem in computer graphics~\cite{tagliasacchi20163d}, and we adapt one of the standard approaches called the simplified medial axis (SMA)~\cite{foskey2003efficient}.
\reffig{fig:topo_skeleton} and \subref{fig:topo_skeleton_thin} show the difference before and after thinning: after the operation is applied, all that is left is a one voxel-thick skeleton diagram.

To be more useful for planning, we want to further sparsify this diagram into a sparse graph, removing the notion of discrete voxel sizes.
That is, we only keep vertices and connect them with straight lines without considering the voxel grid from which they originated, allowing us to create a graph that is as invariant as possible across different map resolutions (see \cite{oleynikova2018sparse} for quantitative comparisons of this property).

We propose a different method of generating the sparse graph from the skeleton diagram from our previous work, which does not follow the underlying diagram exactly (as our previous work did) but greatly simplifies it.
The downside to this is that edges no longer follow the maximum-clearance edges, but may now pass through intraversable space. 
However, they will always be \textit{very near} traversable space, therefore as long as a flexible smoothing method is employed afterwards, a feasible path will be found. 

We take all vertices in the skeleton diagram, assign them unique vertex IDs, and perform a flood-fill in all directions that contain edges, labelling the edges with the two nearest vertex IDs.
This is a speed improvement and simplification over our previous ``edge-following'' algorithm, and does not suffer from edge cases where edge connections are missed.

To simplify this graph, we attempt to remove all vertices that are not adding information to the graph -- essentially, vertices that are on straight lines or nearly-straight lines between other vertices.
The filter consists of removing vertices that have exactly two edges, and whose removal will not displace the edges more than 2 voxels.

As a final step, we attempt to find any way to reconnect disconnected subgraphs.
We label each self-contained subgraph in the sparse graph with another flood-fill operation, and assign subgraph IDs.
We then iteratively search for connections from all subgraphs to all other subgraphs: if a connection is found, one of the subgraphs is relabelled.
To see if two sub-graphs can be connected, we first search along the skeleton diagram using A*: if a connection exists, we insert a new edge between the two closest vertices.
If no connection exists in the diagram (which very occasionally happens due to discretization error), we search in the traversable space of the ESDF, again using A*.
If a path exists there, then we attempt to verify that ESDF path is valid and close to straight-line.

Both of these methods together are much faster at producing usable sparse planning graphs than our previous approach, which we validate and quantify on a number of datasets in \refsec{sec:skeleton_eval}.
The new improvements introduce up to an order of magnitude in speed-up, bringing the total computation time of the complete sparse toplogical graph down to between 1 and 10 seconds for complex maps, allowing this method to be used for a variety of global planning applications.

\section{Global Planning}
In this section we will discuss different global planning strategies and their advantages and disadvantages.
It is assumed that a complete global map is available for these methods, which in our example applications of surveying earthquake damaged buildings or performing repeated industrial inspection, are created from an initial flight through a space using the methods described in \refsec{sec:dense}.
We plan only in position space, assuming the MAV is a sphere for collision-checking purposes (as this makes it rotationally-invariant), and the output of all the global planners is a list of position waypoints from start to goal.
While the final dynamically-feasible trajectory that the MAV will execute will have both positions, yaw, and their derivatives, indexed by time, the task of converting these waypoints to such a trajectory are left to the local planners described in \refsec{sec:local_replanning} or the path smoothers described in \refsec{sec:path_smoothing}.
All methods described here are available open-source\footnote{\url{http://github.com/ethz-asl/mav_voxblox_planning}}.

\subsection{Sampling-based Methods}
The first class of methods we will consider are sampling-based methods.
They are very well suited for large 3D problems, as they do not suffer from the same scaling problem as search-based methods, such as Dijkstra or A*, in large 3D volumes.
One of the most commonly-used classes of methods is Rapidly-exploring Random Trees (RRT)~\cite{lavalle2000rapidly}, where random points are sampled in the planning space (in our case, just 3D position space) and iteratively connected to a tree.
Once the goal point is sampled, there exists a path from the start point (the root of the tree) to the goal.

One key advantage of this class of methods is that all that is required is a way to determine if a randomly-sampled state is valid or not, and a way to determine whether connections between states are valid. 
We propose two different methods to do this: one in the ESDF, which is pessimistic (assuming unknown space is occupied) and on the TSDF, which is optimisic (assuming unknown space is free).
As long as they are coupled with a conservative/pessimistic local planner, either method can be used.

The look-up in the ESDF requires only a single point per pose (as the ESDF already stores the distance to the nearest obstacle, and therefore as long as that distance is larger than the robot radius, the point is feasible), while the TSDF look-up requires looking up an entire sphere and every voxel within.
Additionally, for determining if motions between two states are valid, we use a ray-cast operation to not miss any potentially occupied voxels.

Here we briefly review a few of the methods that we benchmark in the evaluations.

RRT-Connect is a very fast variant of the original RRT algorithm, which does a bi-directional search: growing a tree from both the start and the goal points~\cite{kuffner2000rrt}.
One the downside, it does not optimize the paths (the algorithm terminates once a path is found), so they are often much longer than necessary in complex environments.
For the purposes of our benchmarks, we treat this as the ``first solution'' time and solution length for RRT-based algorithms.

RRT* combines the best of both A* (which is an optimal planning algorithm) and random sampling to create a probabilistically-optimal planning solution~\cite{karaman2011sampling} .
This method samples new points and rewires the graph for shorter solutions, up until a time-limit is reached.
There are other variants, such as Informed RRT*~\cite{gammell2014informed}, which iteratively shrink the sampling space after an initial solution is found, similar to how the admissible heuristic is used in A*.

In general, this is the class of methods we prefer to use, as they give short, nearly-optimal paths, and it is possible to decide how to trade-off computation time versus optimality depending on application.

For planning in a changing map, RRT-based methods are a very powerful tool, as they do not store any information between planning queries.
However, for global planning in a static map, this discards and replicates a large amount of sampling effort.
Therefore, Probabilistic Roadmaps (PRMs) are suitable for many applications where the map remains fixed.

These approaches are similar to the topological graphs described below, in that they usually consist of two stages: building the roadmap, and then searching the roadmap for a solution. 
However, they suffer the same drawbacks as all probabilistically-optimal methods: it is not clear how much sampling is ``enough'' sampling, so it can only be decided by heuristics, and small openings or narrow corridors pose huge problems for PRM-based methods (as the chances of samples landing within them are small).

\subsection{Topological Graphs}
Our proposed method is a search through the sparse topological graph generated in \refsec{sec:sparse_topology}.
Unlike PRM-based methods, ours is deterministic and has a fixed computation time.
Additionally, since the graph is based on the structure of the ESDF, which already encodes the geometry of the scene, it does not suffer from sampling issues in narrow corridor openings.
The downside to this fixed execution and search time is that while the graph will contain all topologically-distinct homotopies of the space, it is not guaranteed that the path length in the graph is the same as the shortest path length through the space.
We perform graph shortening (described below) to attempt to overcome this issue, but it is possible that an incorrect homotope will be selected (though this can also be a problem in PRMs, depending on how the point distributions are sampled).

The method works as a two-stage search, starting from the sparse graph from \refsec{sec:sparse_topology}.
We first find the nearest sparse graph vertex to the start and goal by using a pre-computed k-D tree of vertices.
Then we find a path through the graph using A*.
Due to the small size of the graph, this is a very fast operation, so it is possible to solve the problem for multiple start and goal vertices from the k-D tree to attempt to find a better solution.
Finally, we plan from the start pose to the start vertex using A* in the ESDF, and likewise for the goal.
Since these distances are always short, this is also not an expensive process.

\subsubsection{Graph Shortening}
\label{sec:path_shortening}
While planning through the sparse topological graph is extremely fast, the waypoints it produces aim to maximize clearance, not necessarily minimize absolute path length (path length on the \textit{graph} is minimized) through all traversable space.
For instance, even a straight-line path from A to B would zig-zag along the maximum clearance lines in the graph.

\begin{figure}[tbp]
  \centering
  \includegraphics[width=1.0\columnwidth,trim=0 0 0 0 mm, clip=true]{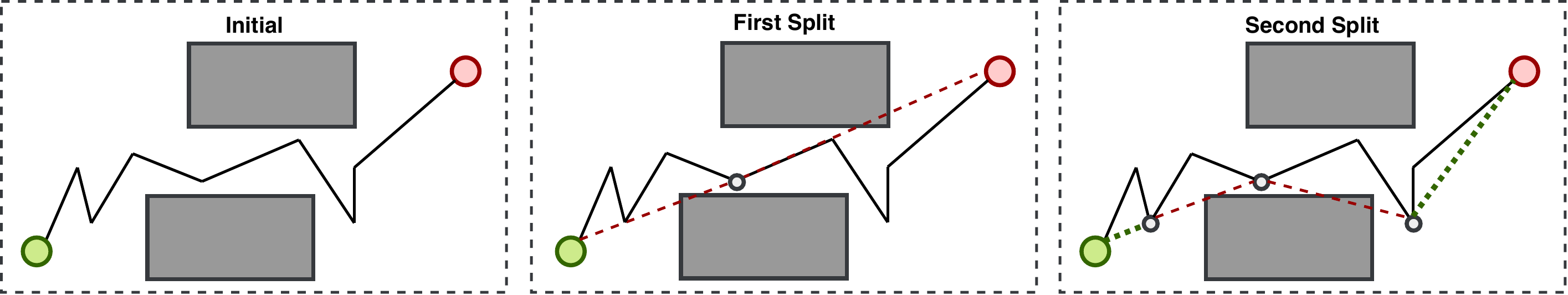}
  \caption{First three steps of binary-search graph shortening. The path is iteratively split in half, until sub-paths are able to be shortened or no more splits are possible. Green shows successful shortened paths, while red shows invalid shorten attempts.}
  \label{fig:graph_shortening}
\end{figure}

To overcome this, we use iterative path shortening in the ESDF.
We attempt to short-cut between pairs of waypoints on the initial graph path in a binary search manner, checking for traversability in the ESDF map, shown in \reffig{fig:graph_shortening}.

We first try to shortcut directly from start to goal; if the straight-line path is not traversable, we then split the waypoint list into two halves: front to middle and middle to back.
Each half is then iteratively checked, whether the intermediate vertices can simply be removed; if not, it is further split into two halves.

We perform this full splitting procedure multiple times to ensure that no further shortening is possible.
This is similar to what the OMPL library~\cite{sucan2012open} does with the RRT-planned paths, with the important distinction that our method is deterministic, while theirs randomly tries to connect pairs of waypoints.
This means that ours does not need heuristics to know when it is terminated: once no more changes are made, the waypoint list is as shortened as possible.
The randomized approach requires options such as maximum steps and maximum empty steps (steps that do not shorten) before terminating, which means that it may still be possible to shorten the graph at termination.


\section{Path Smoothing}
\label{sec:path_smoothing}
Path smoothing deals with taking a set of waypoints and converting them to a smooth, dynamically-feasible path.
We present three methods we compare for these purposes: velocity ramp, polynomial, and our approach named Loco.
We enforce dynamic constraints in the form of maximum velocity and acceleration limits. 

\subsection{Velocity Ramp}
The simplest method is velocity ramp.
A straight-line path is drawn between consecutive pairs of waypoints, and maximum acceleration is applied until the velocity limit is reached, at which point the acceleration is zero.
The same principle is applied on deccelerating toward the next point.

The total time between two waypoints is described as:
\begin{equation}
 t =  \frac{v_{max}}{a_{max}}  + \frac{\norm{\vec{x}_{goal} - \vec{x}_{start}}}{v_{max}}
\end{equation}
where $t$ is the time in seconds, $v_{max}$ and $a_{max}$ are the velocity and acceleration constraints, respectively, and $\vec{x}_{start}$ and $\vec{x}_{goal}$ are the 3 DoF positions of the start and goal.

\subsection{Polynomial}
High-degree polynomial splines are a common representation for MAV trajectories, as they are easy to compute, can be smooth and continuous up to high derivatives, and are shown to be dynamically feasible as long as velocity and acceleration constraints are met~\cite{mellinger2011minimum}, \cite{richter2013polynomial}.

We implement a path smoothing method from Richter \etal\cite{richter2013polynomial}, where a polynomial spline is fit to the waypoints and then iteratively split at collisions.

First, we discuss the optimization problem.
We formulate it to minimize a high derivative such as jerk or snap, as shown to be desirable by Mellinger \etal~\cite{mellinger2011minimum}.

We will consider a polynomial spline in $K$ dimensions, with $S$ segments, and each segment of order $N$.
Each segment has $K$ dimensions, each of which is described by an $N$th order polynomial:
\begin{equation}
f_k(t) = a_0 + a_1 t + a_2 t^2 + a_3 t^3 \dots a_N t^N
\end{equation}
with the polynomial coefficients:
\begin{equation}
\vec{p}_k = \begin{bmatrix} a_0 & a_1 & a_2 & \dots & a_N \end{bmatrix}^\top.
\end{equation}

Rather than optimizing over the polynomial coefficients directly, which has numerical issues at high $N$s, we instead optimize over the end-derivatives of segments within the spline~\cite{richter2013polynomial}.
We distinguish between fixed derivatives $\vec{d}_F$ (such as end-constraints) and free derivatives $\vec{d}_P$ (such as intermediate spline connections):
\begin{equation}
\vec{p} = \mat{A}^{-1} \mat{M} 
\begin{bmatrix}
\vec{d}_F \\
\vec{d}_P
\end{bmatrix}.
\end{equation}
Where $\mat{A}$ is a mapping matrix from polynomial coefficients to end-derivatives, and $\mat{M}$ is a reordering matrix to separate $\vec{d}_F$ and $\vec{d}_P$.

We aim to minimize the derivative cost, $J_d$, which represents a certain derivative (often jerk or snap) of the position \cite{mellinger2011minimum}, with $\mat{R}$ as the augmented cost matrix.
\begin{eqnarray}
J_d &=& \vec{d}_F^\top \mat{R}_{FF} \vec{d}_F + \vec{d}_F^\top \mat{R}_{FP} \vec{d}_P + \nonumber\\
&& \vec{d}_P ^\top \mat{R}_{PF} \vec{d}_F + \vec{d}_P^\top \mat{R}_{PP} \vec{d}_P \label{eq:j_d}
\end{eqnarray}

Finding the $\vec{d}_P^\star$ that minimized $J_d$ is possible to do in closed-form~\cite{richter2013polynomial}:
\begin{equation}
\vec{d}_P^\star = -\mat{R}_{PP}^{-1} \mat{R}_{FP}^\top \vec{d}_F
\end{equation}

This method allows us to fit a smooth polynomial spline to a series of waypoints, by using the positions of the waypoints as vertex constraints.

However, since waypoint trajectories are planned such that the \textit{straight-line} path (visibility graph) between them is collision-free, the smoothed path often runs into collision. 
To remedy this, any time a collision is detected, this method adds a new waypoint on the visibility graph closest to its projection onto the straight-line path, and the optimization is re-run.

While this is fast and easy to implement, this method suffers from occasionally not being able to escape collisions, and in difficult cases, creates many extra waypoints.
The optimization problem does not scale well numerically when there are many waypoints, especially close together in time, as the segment times get very short (and must be raised to high powers).
Furthermore, adding these additional waypoints often perturbs the trajectory in unexpected ways, causing the robot to take large detours.

\subsection{Local Continuous Optimization (Loco)}
\label{sec:loco}
To overcome these issues, we propose our own LOcal Continuous Optimization method,  Loco~\cite{oleynikova2016continuous-time}.
Rather than iteratively collision-checking and splitting the smoothed polynomial trajectories, we introduce the collisions as soft costs in the optimization, following the general structure that Ratliff \etal~\cite{ratliff2009chomp} proposed in their CHOMP method.

Introducing this soft cost leads to the following optimization problem, where $w$ terms are constant weights:
\begin{equation}
\vec{d}_P^* = \argmin_{\vec{d}_P}\; w_d J_d + w_c J_c\label{eq:optimization}
\end{equation}

$J_d$ remains as in \refeq{eq:j_d} above, and we introduce a new term, $J_c$ represents a soft collision cost:

\begin{equation}
J_c = \sum_{t=0}^{t_m} c (\vec{f}(t)) \norm{\vec{v}(t)} \Delta t \label{eq:j_c}
\end{equation}
which approximates the line integral of costs along the path, where $c(\vec{x})$ is the collision cost from the map, $\vec{f}(t)$ is the position along the trajectory at time $t$, and $\vec{v}(t)$ is the velocity at time $t$.

For the collision cost in the map, we use a smooth gradually decreasing function proposed in CHOMP˜\cite{ratliff2009chomp}, where $\epsilon$ is a tuning value for how far outside the robot radius we care about collisions, $\vec{x}$ is a position in the map, and $d(\vec{x})$ is the ESDF distance at that point:
\begin{equation}
c(\vec{x}) = 
\begin{cases} 
-d(\vec{x})+\frac{1}{2}\epsilon & \text{if } d(\vec{x}) < 0 \\
\frac{1}{2\epsilon} (d(\vec{x})-\epsilon)^2 & \text{if }  0\leq d(\vec{x}) \leq \epsilon \\
0 & \text{otherwise}
\end{cases}
\end{equation}

In practice our robot radius is rarely $0$, so we subtract the robot radius $r$ from $d(\vec{x})$. 

Now that we have a method of locally optimizing trajectories to be collision-free, evaluated at length in our previous work~\cite{oleynikova2016continuous-time}, the question is how to best make it fit through a series of waypoints.

We can use the starting method described above, where each waypoint simply becomes a control point/vertex in the spline.
This means that for every waypoint, we have another segment in the spline.
This has various downsides, most notably that for long, complex trajectories, the problem gets significantly slower computationally, and can have numerical scaling issues.

We found in practice that the optimization works best with a small (3-5) number of segments, therefore we explored methods to fit a visibility graph to these segments.
The first solution was to generate an initial polynomial solution passing through all waypoints, and then re-sample it down to $S$ segments, by selecting $S-1$ evenly-spaced times to sample the trajectory at.
These then become the new waypoints.
We will refer to this strategy as ``polynomial resampling'' in the results in \refsec{sec:loco_results}.

The second method is to instead sample directly on the visibility graph.
Rather than sampling evenly-spaced $t$s on a polynomial trajectory, we instead fit a velocity ramp straight-line trajectory to a visibility graph, and sample the $t$s on this trajectory.
This method is referred to as ``visibility resampling'' in the results.

As shown in \refsec{sec:loco_results}, both of these methods create better, higher-quality paths faster than simple waypoint fitting.

\section{Local Replanning}
\label{sec:local_replanning}
The system components we have described so far allow us to plan efficiently in a known map.
However, in order to build the map to begin with, the MAV needs to autonomously navigate in an initially unknown environment, while remaining safe.
Similarly, to deal with changes in the environment between when the global map was constructed and in repeated missions, we want to use the local planner to re-plan any path that may have become in collision.
In our sample applications, this covers examples where more rubble has collapsed since the initial survey of an earthquake-damaged building, or equipment has moved from the last visit in an industrial inpsection scenario.
The robot must remain safe in both of these cases, even if it is using a global plan to navigate.

\begin{figure}[tbp]
  \centering
  \includegraphics[width=1.0\columnwidth]{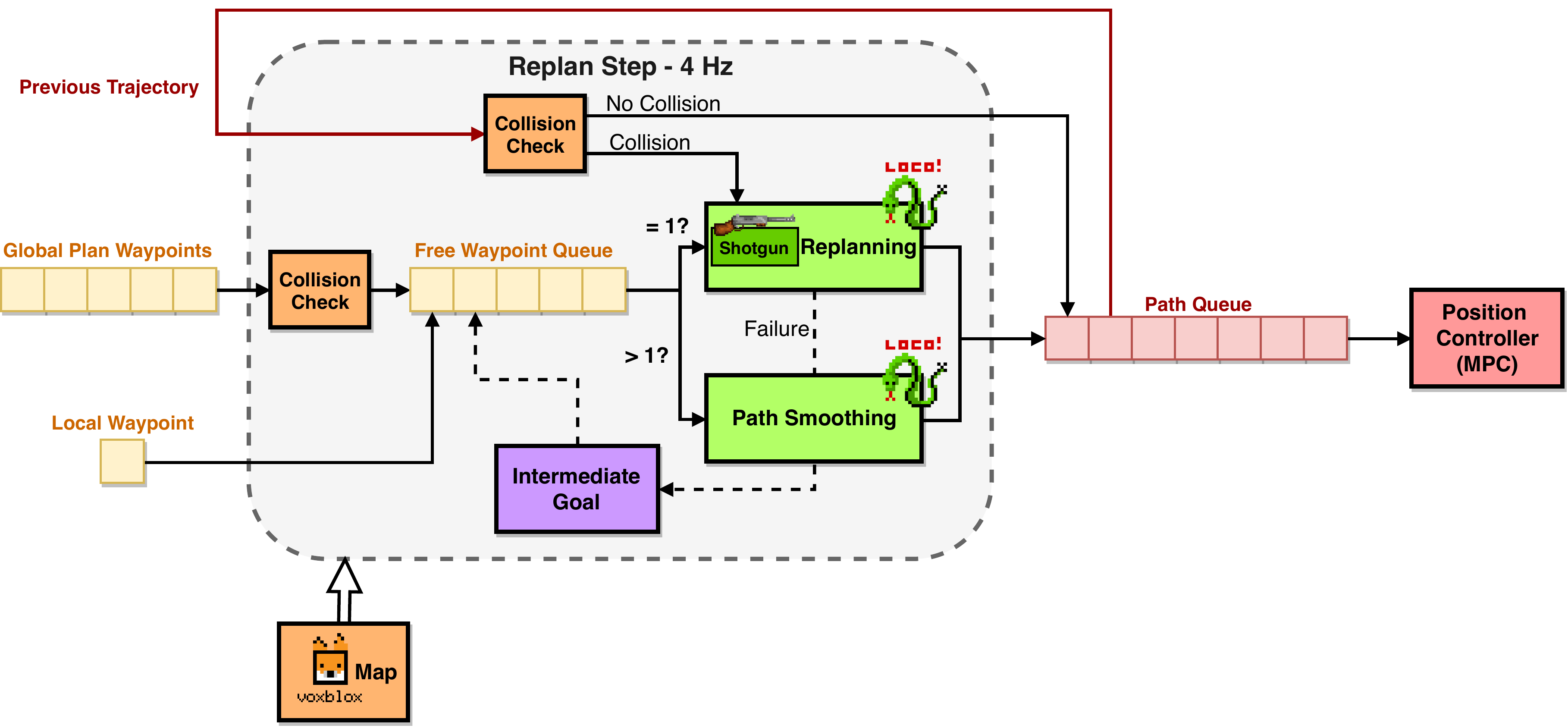}
  \caption{Details of the local planner. This architecture is designed to track both single waypoints and a global waypoint path, and constantly check the executed trajectory for new collisions. This allows our system to be safe even in the presence of new dynamic obstacles, drifting coordinate frames, or if the global map used is different from the current state of the environment.}
  \label{fig:local_planner_system}
\end{figure}

Therefore, we propose a local planner that is able to handle both a single general goal that may be in unknown or occupied space (for initial map building missions), and a more complex series of waypoints from a global planner, and ensure safety of any trajectory sent to the controller.
While the underlying optimization we use for generating the trajectories is Loco and already described in \refsec{sec:loco} above, here we describe how to handle the different local replanning cases described above in analogous ways, especially in partially unknown or changing environments.
A diagram of our system is shown in \reffig{fig:local_planner_system}.

\subsection{Comparison to Related Work}
It is worth mentioning that after our introduction of continuous-time local trajectory optimization methods in~\cite{oleynikova2016continuous-time}, this has become a popular approach to solving local replanning problems.
The system presented by Lin \etal~\cite{lin2017autonomous} uses the same general form of optimization problem, while seeding the optimization with an initial waypoint plan guess from a search-based algorithm, which treats unknown space as free, allowing them to quickly find a path often all the way to the goal.
However, depending on how cluttered the environment is, the distance and field of view that the sensor can perceive, and the speed with which the MAV is traveling, this may not be a safe assumption.
Though their system replans continuously, it is possible to create cases where either the plan will traverse unseen space before the next replanning cycle or that the space will never be seen with the current trajectory.
For sparsely-cluttered applications, these are not important safety concerns, but are crucial for our earthquake-damaged building scenario.

Usenko \etal~\cite{usenko2017real} speed up the optimization significantly by switching to B-splines rather than Hermetian splines and better exploiting the structure of the optimization problem, and introduce soft costs for the goal to deal with situations where the goal is inside an obstacle or inaccessible.
The algorithm is seeded with an initial path from RRT, following Richter \etal \cite{richter2013polynomial}, again assuming unknown space is free.
However, since the map they use is only a local sliding window, they would not be able to plan out of situations where a new obstacle requires a larger detour than the sliding window covers.
Likewise, their approach assumes unknown space is free both in the global planning step and in the local replanning, which can create safety issues with narrow field-of-view sensors or complex environments.

Morell \etal~\cite{morrell2018comparison} also propose a similar method, with the additional options to constrain the optimization within computed free-space corridors in the space or respect other hard constraints in the optimization.
They propose multiple variants, ranging from conservatively planning only through convex regions of free-space to treating unknown space as free.
However, their planning times are still in the order of magnitude of seconds, which makes the proposed methods well-suited for path smoothing, but not appropriate for online replanning, especially in the presence of new obstacles.

In contrast all these works, this section focuses on the system-level considerations of how a local planner, using this local trajectory optimization strategy, can be used in conjunction with both single goals and global planners. 
We focus specifically on tracking waypoints which may not be in free space, attempting to follow global plans which may no longer be valid as the environment has changed since the global map was created, and guarantee safety even in unknown or slowly changing environments.

\subsection{Local Planner System}
We show an overview of the local planner system, which is able to handle both single goal waypoints or a series of waypoints from a global planner, in \reffig{fig:local_planner_system}.
We show that all user inputs, whether an entire waypoint trajectory from a global planner or a single waypoint to track, are treated similarly in a waypoint queue.
If there are multiple waypoints in free space in the queue, we attempt to smooth a path through them directly using the path smoothing strategy described in \refsec{sec:loco}.
Otherwise, we attempt track the next waypoint in the queue using the local replanning directly.
While the underlying optimization is the same, the difficult component is selecting the goal point of the initial condition of the optimization.
Even if we have a soft cost (rather than a hard constraint) on the goal point, it is still necessary that the initial goal is in or very near free space.
We propose a shotgun approach for selecting the initial trajectory end-point for the polynomial optimization, described in \refsec{sec:shotgun} below, to intelligently select a free-space end-point when the first waypoint is in collision or unknown space.

Another issue is that the planner will not always reach a waypoint even with a valid goal, for a variety of reasons -- too narrow corridors, lack of connectivity in regions, and especially narrow field of view in vision-based sensors, which may not clear enough observed free-space for the MAV to pass.
In cluttered scenarios, local optimization alone may not lead to observing enough free space to reach a global goal.
To overcome these further issues, we introduce intermediate goal points which are inserted into the waypoint queue when the local replanning fails.
This idea was originally introduced in our previous work~\cite{oleynikova2018safe} and reviewed in \refsec{sec:intermediate_goal} below.

\subsection{Shotgun Approach for Seeding Initial Solution}
\label{sec:shotgun}
Shotgun goal selection deals with the issue of what happens if the currently-tracked waypoint is in unknown or occupied space.
As mentioned before, our local planning strategy is based on local trajectory optimization, which will fail if the goal point is not a valid state for the MAV to be in.
However, in cases such as exploration or with the random or exploration-seeking goal strategies described in the sections below, there are frequently cases that the MAV is attempting to seek a goal that is in collision.
To seed the initial solution of the polynomial optimization with a valid end-state (and optionally, valid intermediate states), we propose a ``shotgun'' strategy that is in spirit analogous to shooting methods in mathematics.

In our previous work~\cite{oleynikova2018safe}, we explored finding an initial end-point along a straight-line path from the start to the desired goal, or following the collision gradient back from the goal.
However, this added the unnecessary constraint that there must be valid free-space on either the straight-line path between the goal and the start point, or if using gradient-following, that there are no local minima in the gradient.
This led to a number of failures in more complex environments, including cases where disconnected pockets of free space were found as the initial end-point, creating unnecessary failures in the local planning.

Instead of searching backwards from the original goal, we propose a strategy to find an initial trajectory end-point by searching from the current position of the MAV (or the trajectory start point in case of continuous replanning).
The idea draws inspiration from shooting methods in mathematics combined with ideas from particle filtering.
Our ``particles'' represent potential paths through the voxelized map, combining goal-seeking behavior, exploration through a random walk, and trying to maximize clearance to obstacles.

This formulation is designed with several ideas in mind: first, allow users of the system to trade-off between multiple potential behaviors that the system may have.
For example, in scenarios where there are both cluttered areas and large open spaces, it may be desirable to maximize path clearance to prefer paths through more open areas.
In scenarios where it is anticipated to have many dead-ends and horseshoe problems, it would be desirable to trade in favor of exploring the space more.

Finally, the core idea and advantage of this approach is to exploit the observed free space as much as possible.
The previous approach of searching backwards from the goal often missed opportunities to get closer to the goal simply by approaching the boundaries of free space in the correct direction.
This property is quantitatively evaluated in \refsec{sec:local_planning_benchmarks} and \reffig{fig:local_benchmark_length}.

The method can be described as follows: every time we attempt to plan to a waypoint, we generate a number of shotgun particles, each of which is allowed to run for a fixed number of iterations.
At each iteration, a particle randomly decides between 3 options, with probabilities assigned by the user:
\begin{easylist}
  \ListProperties(Space=-0.25cm,Space*=-0.25cm)
 & take the direction closest to the goal,
 & take the direction maximizing obstacle distance,
 & or take a random direction (other than where the particle came from).
\end{easylist}

This process terminates when either a particle reaches the goal, or all particles have finished.
If no particle was able to reach the goal, the particle that was able to minimize its straight-line distance to the goal is chosen as the initial end-point for the polynomial optimization.

By trading off between the probabilities of these 3 outcomes, the shotgun particles may seek the goal,  maximize clearance to obstacles, or explore.
To cover the trivial edge-case where the goal lies in free space and is reachable through the current map, the first shotgun particle generated will always perform $100\%$ goal-seeking behavior.
Since any particle to reach the goal terminates the procedure, this means that the trivial case is solved in minimum time.

Optionally, the path that the ``winning'' shotgun particle took to the goal contains valuable information about the initial solution to the planning problem.
This information can be used to seed the complete inital solution to the polynomial optimization: the path of the shotgun particle is subsampled and shortened using the approach described in \refsec{sec:path_shortening}, and then optionally fed as the initial waypoints of the initial solution  to the optimization problem in \refsec{sec:loco}.
This method is described as ``Shotgun'' in the evaluations in \refsec{sec:local_planning_benchmarks}, while not taking advantage of this is referred to as ``Shotgun No Path''.

\subsection{Intermediate Goal Points}
\label{sec:intermediate_goal}

Especially when seeking to navigate through unexplored cluttered environments, local path-planning alone may not be sufficient.
This is especially true in cases of narrow field-of-view sensors, where the MAV may not observe enough free space in the correct direction to be able to perceive a path toward the goal.
As shown in our previous work~\cite{oleynikova2018safe}, adding a strategy to select intermediate goal points when local path planning fails substantially increases chances of success, especially in very cluttered scenes.
Our previous work introduced two main strategies to find intermediate goal points: random and local exploration.

In the random strategy, whenever the optimization would fail to find a viable solution toward the goal, we would select a random intermediate goal within a fixed radius of the MAV's current pose.
This random goal does not need to be in free space, as the replanning strategy can deal with waypoints in collision or unknown space safely (see section above).

The local exploration strategy combines goal-seeking with maximizing an exploration gain.
First, we select a number of candidate points in \textit{free space}.
Note that the position may not be feasible for the entire MAV to fit, the single point at the center needs to simply be free (please refer to our previous work for quantitative evaluations of this property \cite{oleynikova2018safe}).
We then select the best point to use as the intermediate goal by evaluating all points on a combination of two metrics: exploration gain and goal-seeking.

Using knowledge of the model of the MAV's depth sensor (assuming it is a normal pinhole camera whose field of view can be modeled with a frustum), we can evaluate the exploration gain of a randomly-selected pose as follows, where $\vec{x}$ is the position of the MAV in space and $\gamma$ is its yaw angle:
\begin{equation}
l(\vec{x}, \gamma) = \# \{v | v \in \textrm{frustum}(\vec{x}, \gamma) \cap v \in \textrm{unknown}(v)\}
\end{equation}
Which essentially counts the unknown voxels $v$ in the view of the camera frustum.

While this will select points that have the possibility of exploring the most unknown space, it is also essential to make progress toward the global goal ($\vec{g}$) (or the next ``real'' waypoint).
The goal-seeking gain $d_{g}$ is a function of the sampled position $\vec{x}_n$, the trajectory start point $\vec{x}_s$, and maximum sampling radius $r$. $R$ is the total reward of a sampled point $\vec{x}_n$, which we seek to maximize among the samples.
\begin{eqnarray}
d_{g} & = & \norm{\vec{g} - \vec{x}_s} + r \\
R(\vec{x}_n, \gamma, \vec{g}) & = & w_{e} l(\vec{x}, \vec{\gamma}) + w_{g} \frac{d_g - \norm{\vec{g}-\vec{x}_n}}{d_g}
\end{eqnarray}
$w_{e}$ and $w_{g}$ are hand-tuned weights trading off between exploration gain and goal-seeking behavior.

However, whether these strategies make sense to use depends heavily on the application.
One main downside is that they make the behavior of the robot unpredictable: the location of the next most promising point for exploration is difficult for a safety pilot to predict, and the robot will often go into parts of the environment that may not be desirable to explore.
These downsides are shared by all sampling-based exploration algorithms, such as Next Best View planners~\cite{bircher2016receding2}: while performing excellently in simulation, they are often scary to do real experiments with.
For this reason, while we also evaluate these methods in the results section, in many applications employing these intermediate goal point strategies is not a great practical choice.
This is why we developed the Shotgun Approach in \refsec{sec:shotgun}: to capture some of the benefits of these methods while maintaining predictable behavior from the MAV.

\section{Evaluation and Experiments}
We aim to validate our complete system, spanning from quantitative global planning on multiple real-world scenarios and maps, to evaluating a variety of path smoothers in comparison to one another, to thousands of local planning simulation scenarios and finally fully integrated system tests in closed loop on-board the platform.

\subsection{Evaluation Datasets}
\label{sec:datasets}
\begin{figure}[tbp]
  \centering
  \begin{subfigure}[b]{0.45\columnwidth}
    \includegraphics[width=1.0\columnwidth,trim=0 0 0 0 mm, clip=true]{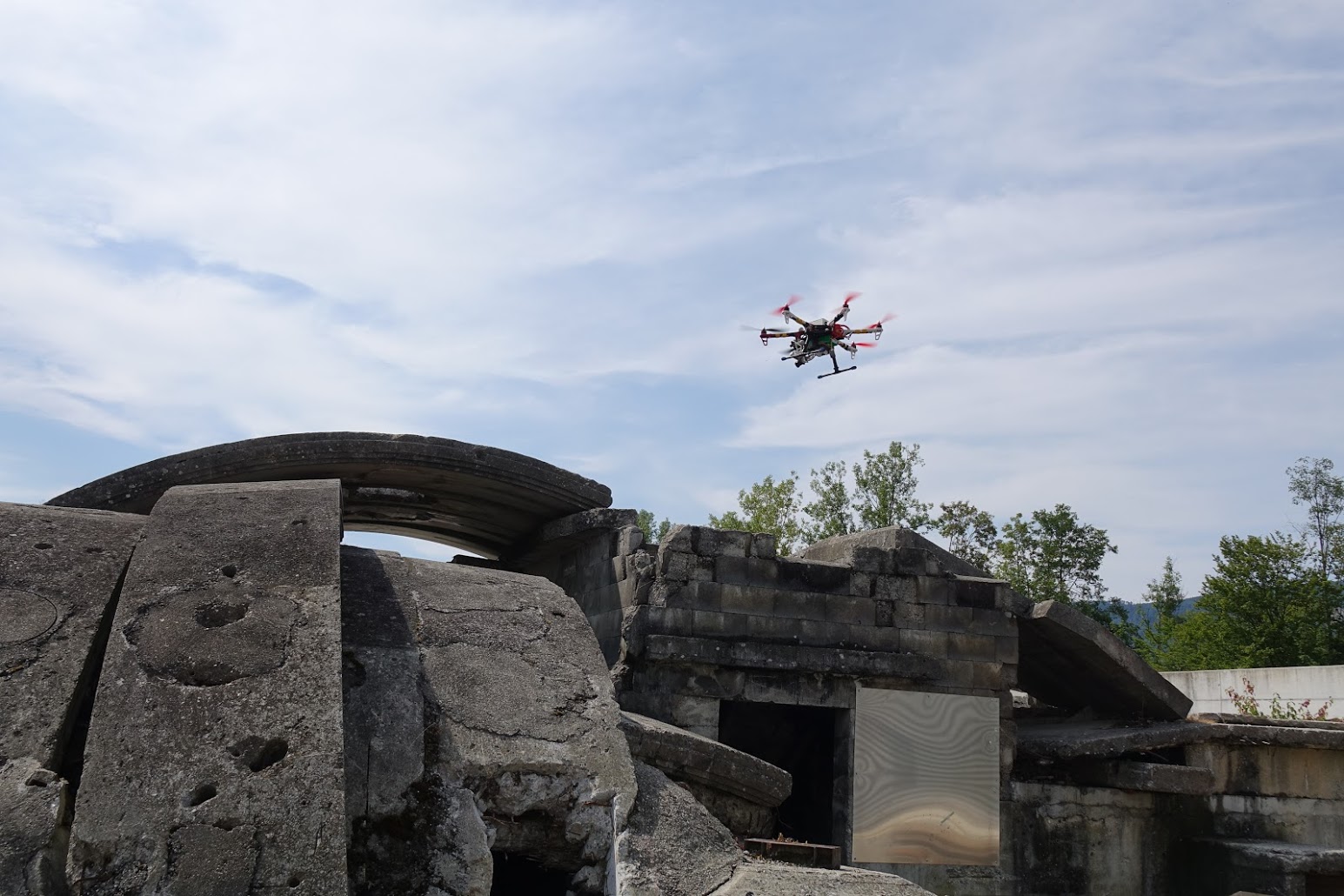}
    \caption{Shed, outdoors}
    \label{fig:jay_arche_outdoors}
  \end{subfigure}  
  \begin{subfigure}[b]{0.45\columnwidth}
  \includegraphics[width=1.0\columnwidth,trim=0 0 0 0 mm, clip=true]{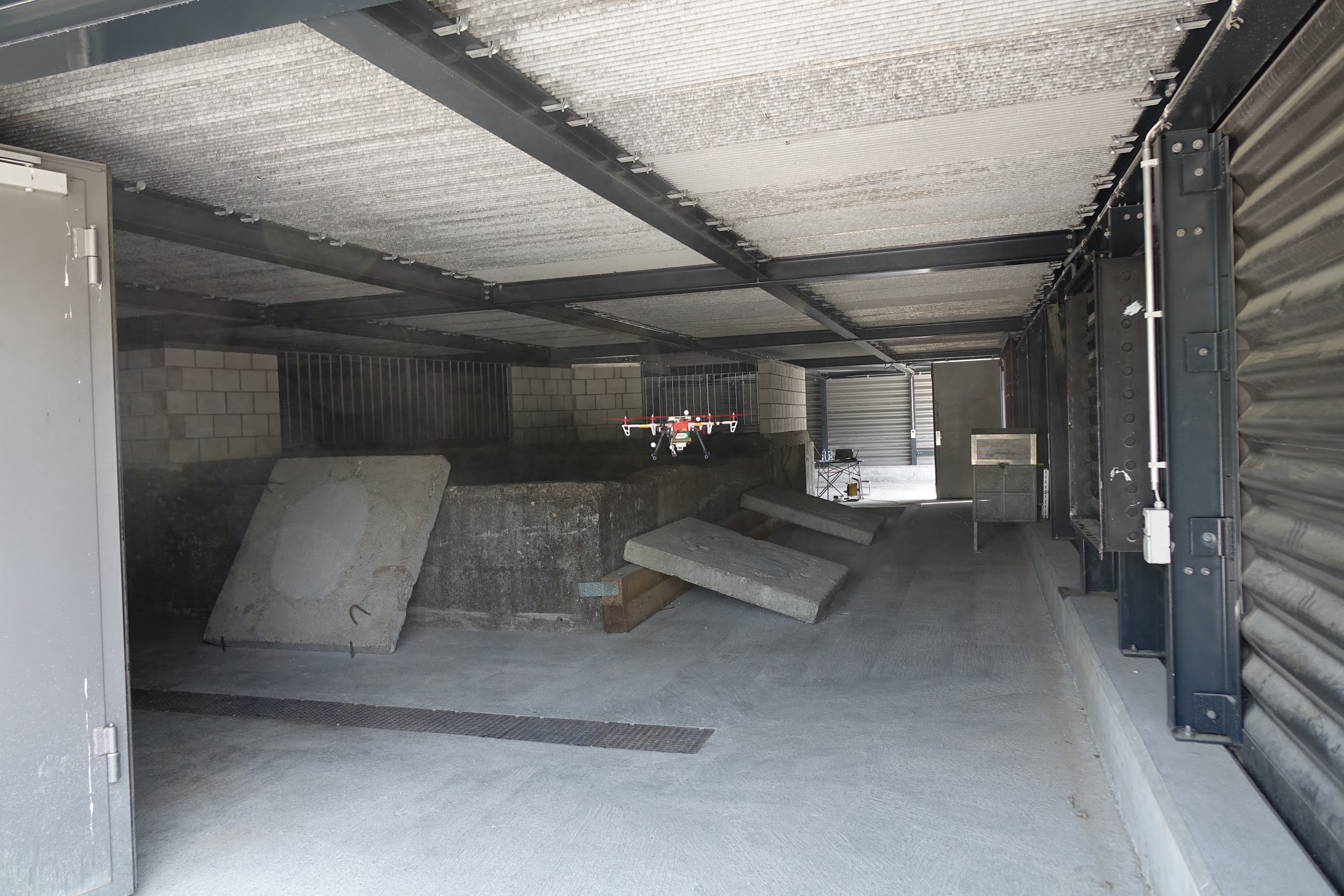}
  \caption{Shed, indoors}
  \label{fig:jay_arche_indoors}
  \end{subfigure}
  \begin{subfigure}[b]{0.45\columnwidth}
  \includegraphics[width=1.0\columnwidth,trim=0 0 0 0 mm, clip=true]{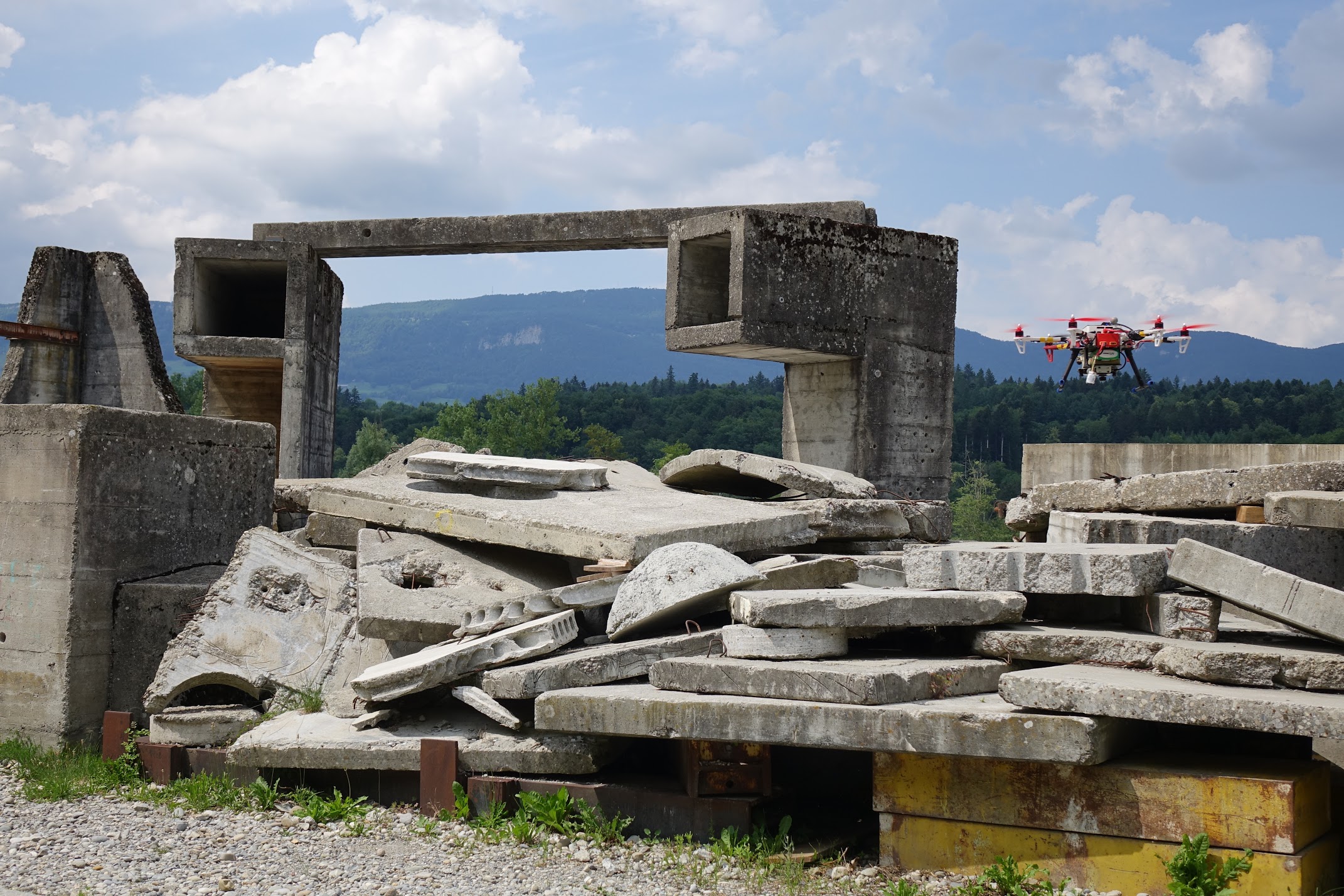}
  \caption{Rubble}
  \label{fig:jay_rubble}
  \end{subfigure}  
  \begin{subfigure}[b]{0.45\columnwidth}
  \includegraphics[width=1.0\columnwidth,trim=0 0 0 0 mm, clip=true]{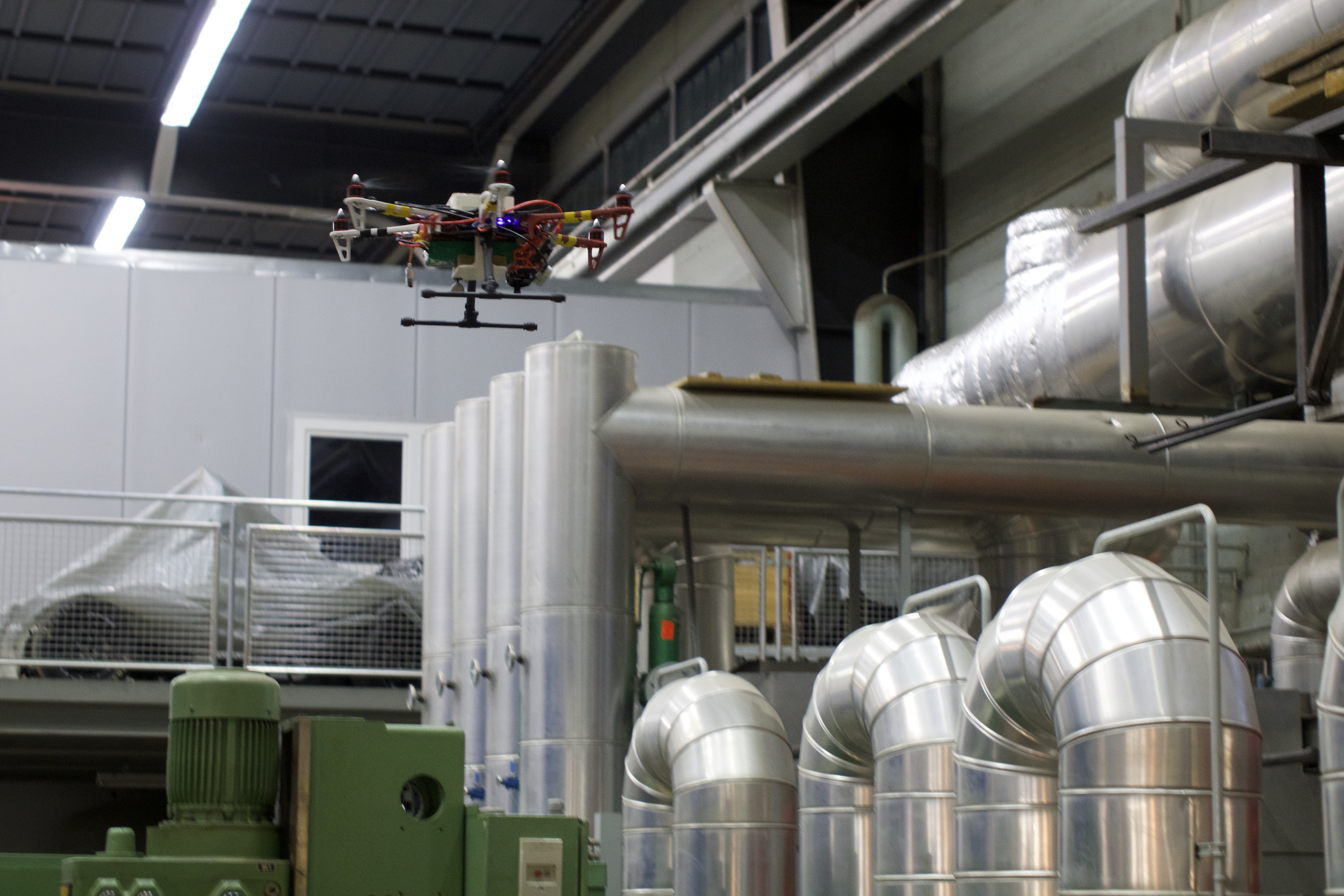}
  \caption{Machine Hall}
  \label{fig:jay_machine_hall}
  \end{subfigure}
  \caption{Photos of the three scenarios from the benchmark dataset. Top (left and right): outdoor and indoor parts of the shed scene. Bottom left: rubble scene. Bottom right: machine hall scene.}
  \label{fig:jay_datasets}
\end{figure}

We focus our evaluations on real datasets, collected in typical scenarios for search and rescue and industrial inspection.
We performed three inspection flights, two at a military search-and-rescue training ground at Wangen an der Aare, and one in the ETH Z{\"u}rich Machine Hall.
Photos of the three areas, named ``Shed'', ``Rubble'', and ``Machine Hall'' respectively, are shown in \reffig{fig:jay_datasets} and described in \reftab{table:datasets}.

The ``Shed'' and ``Rubble'' datasets come from our proposed earthquake-damaged building scenario: the shed is an indoor space with rubble strewn around, and the outside area featured in the dataset also contains collapsed concrete blocks.
Especially the indoor area of the shed features tight passages and turns that are difficult for the MAV to navigate.
The rubble area is an outdoor earthquake-damaged area, with some collapsed and some still standing concrete structures and potentially tight passages between the two standing structures.
The machine hall dataset addresses the industrial inspection scenario: it is an industrial space, again tight and cluttered, that may realistically need repeated inspection from an MAV as not all spaces are reachable by a human, and it contains large machinery that may need regular inspection to check for failures.

All datasets were collected with the MAV and sensor set-up described in \refsec{sec:hardware}, using both stereo and RGB-D (Intel Realsense D415) sensors.

We make six global maps available: two per dataset, one using the stereo cameras and one using the RGB-D sensor (RS).
All the maps are available for download~\footnote{\url{http://github.com/ethz-asl/mav_voxblox_planning}}.

The provided maps were generated with 10 cm voxels, 1 meter clear and 4 meter occupied spheres (described in \refsec{sec:unknown}), 8 meter maximum ray distance for TSDF construction, and 4 meter maximum ESDF computation distance.

\begin{table*}[tbp]
  \begin{tabular}{llllp{5cm}}
    \toprule
    \textbf{Name} & \textbf{Location} & \textbf{Duration} & \textbf{Volume} & \textbf{Contents} \\
    \midrule[1.5pt]
    Shed & Wangen a. A, BE, CH & 217 sec & 38 x 35 x 12 m & Mixed indoor and outdoor dataset with narrow openings \\
    Rubble & Wangen a. A, BE, CH & 159 sec & 28 x 27 x 12 m & Outdoor dataset, over earth-quake damaged buildings \\
    Machine Hall & ETH Z\"urich, ZH, CH & 251 sec & 24 x 30 x 8 m & Indoor area, with large industrial machinery \\
    \bottomrule
  \end{tabular}
  \caption{Dataset statistics and descriptions.}
  \label{table:datasets}
\end{table*}

\begin{figure}[tbp]
  \centering
  \begin{subfigure}[b]{0.45\columnwidth}
    \includegraphics[width=1.0\columnwidth,trim=0 0 320 0 mm, clip=true]{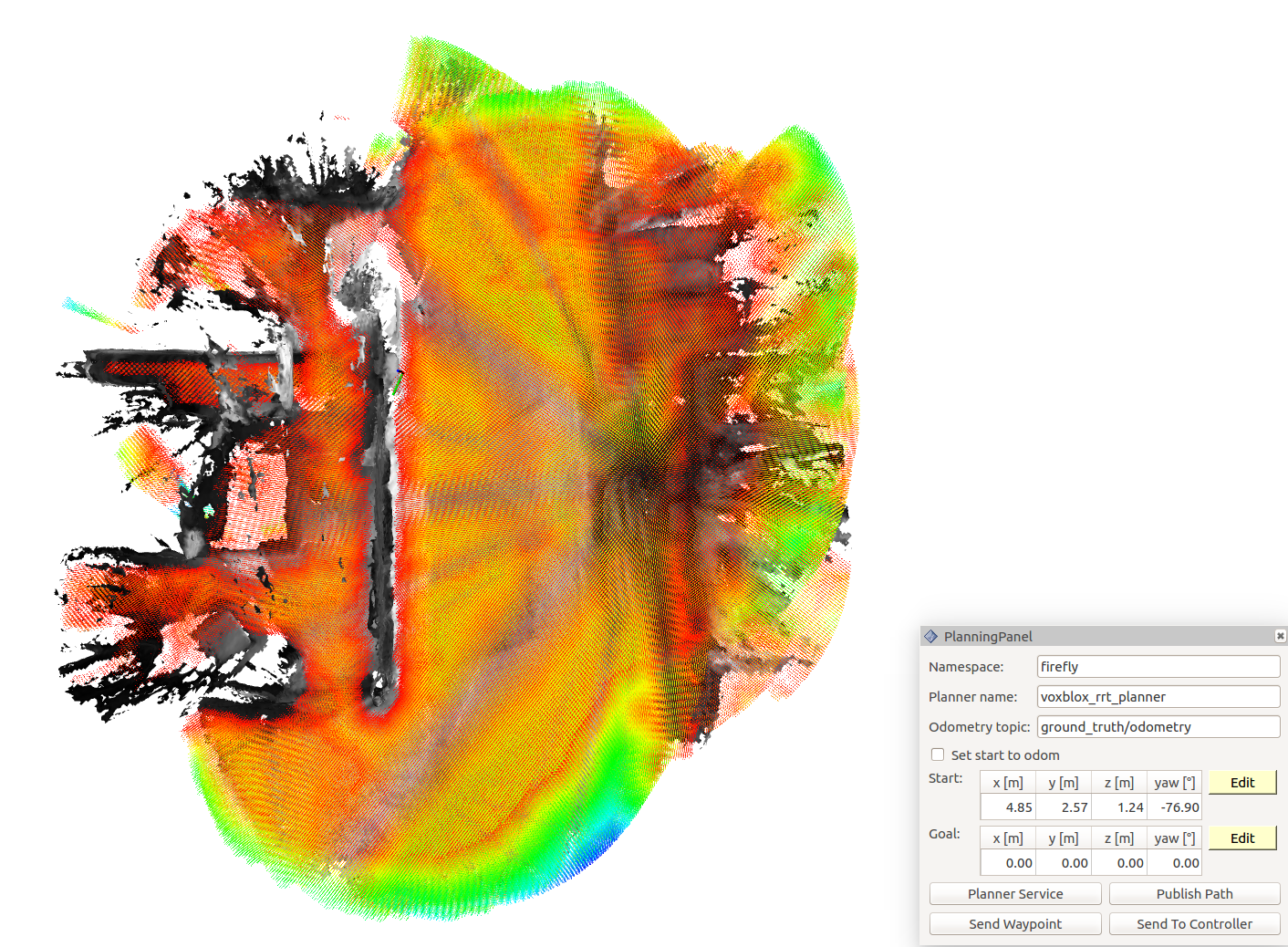}
    \caption{Shed (Stereo)}
  \end{subfigure}    
  \begin{subfigure}[b]{0.45\columnwidth}
    \includegraphics[width=1.0\columnwidth,trim=0 0 320 0 mm, clip=true]{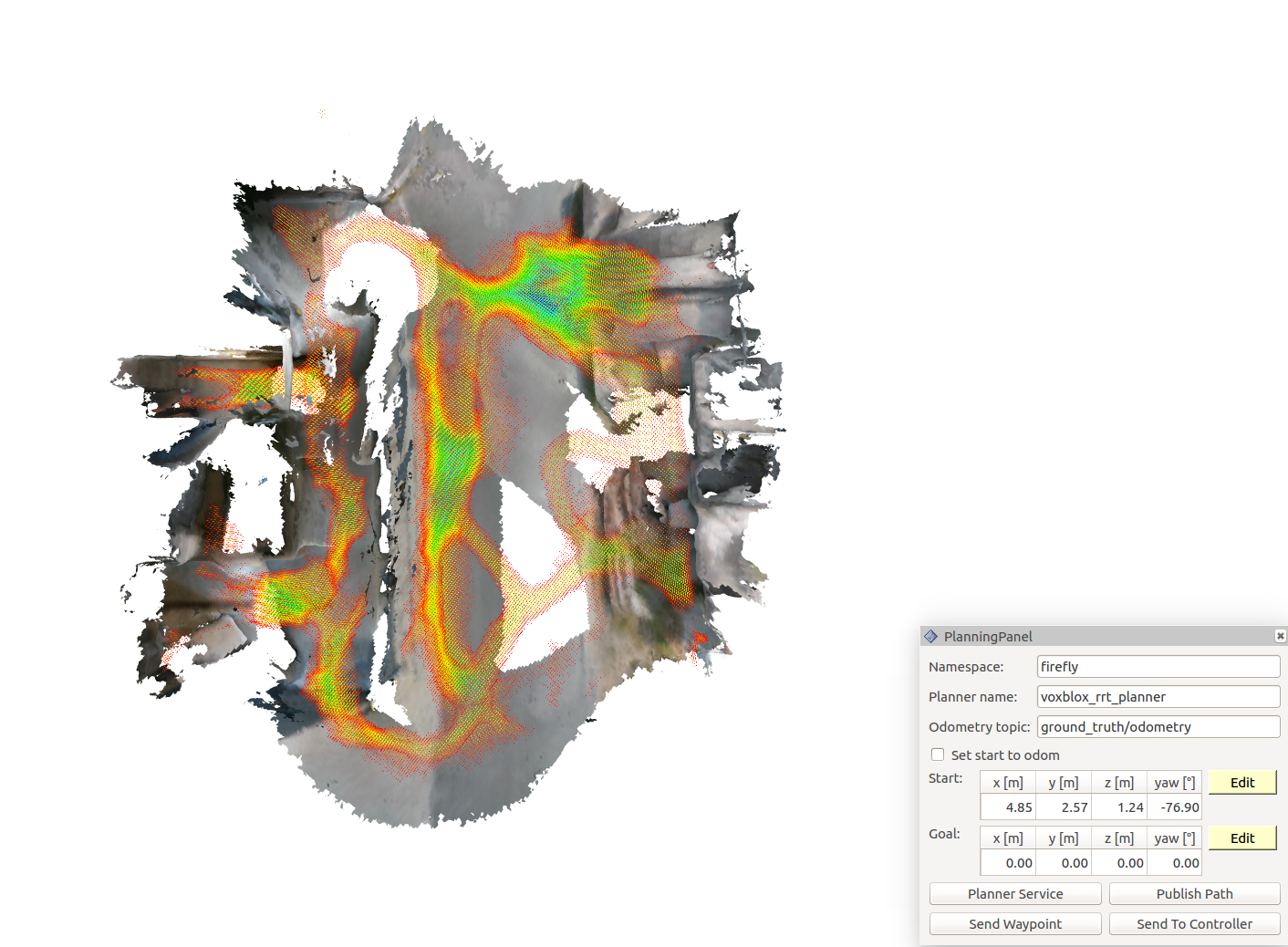}
    \caption{Shed (RS)}
  \end{subfigure}    
  \caption{Traversability differences between stereo datasets and realsense datasets. Traversable space (given an 0.5 meter robot radius) is shown as colored points. As can be seen, much more space is considered traversable with stereo data.}
  \label{fig:traversability}
\end{figure}

We provide both a stereo and an RealSense (RGB-D) map of all environments due to the narrow field of view (FoV) of the RealSense camera, but more accurate depth measurements (and color information).
\reffig{fig:traversability} shows the difference in traversability between the stereo and the RGB-D version of the shed dataset, assuming an 0.5 meter robot radius: again, this is due to the difference in field of view between the stereo camera and the RealSense camera: our stereo visual-inertial sensor has an $86\deg$ field-of-view for each camera, and the RealSense D415 has a $65\deg$ horizontal field of view.
This means that in regular flight, when the MAV is facing in the direction of travel, the RealSense will observe far fewer obstacles on the sides, especially if the obstacles are far.

\begin{figure}[tbp]
  \centering
  
\begin{subfigure}[b]{0.45\columnwidth}
  \includegraphics[width=1.0\columnwidth,trim=0 80 0 80 mm, clip=true]{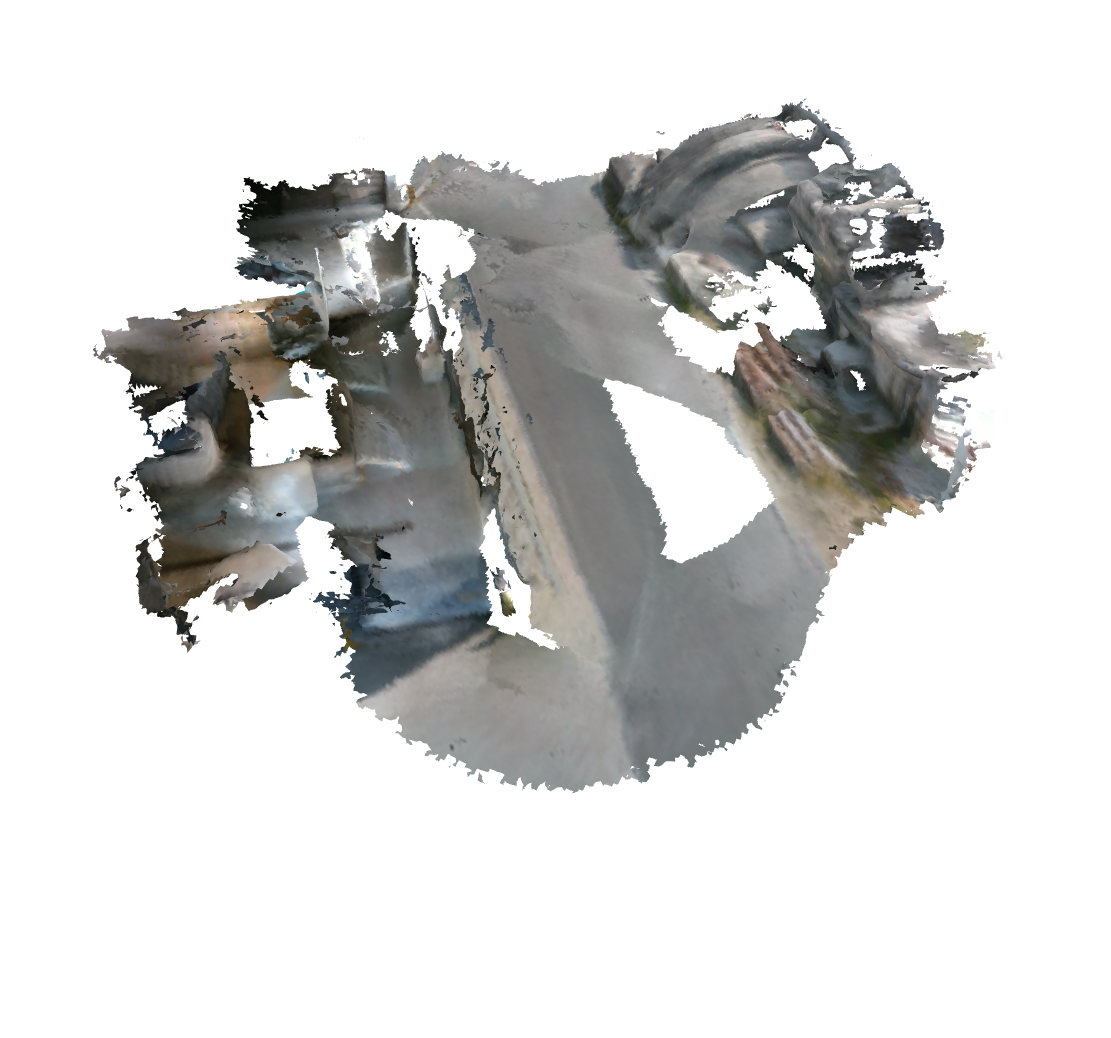}
  \caption{Shed (RS)}
  \label{fig:shed_rs}
\end{subfigure}    
\begin{subfigure}[b]{0.45\columnwidth}
  \includegraphics[width=1.0\columnwidth,trim=0 80 0 80 mm, clip=true]{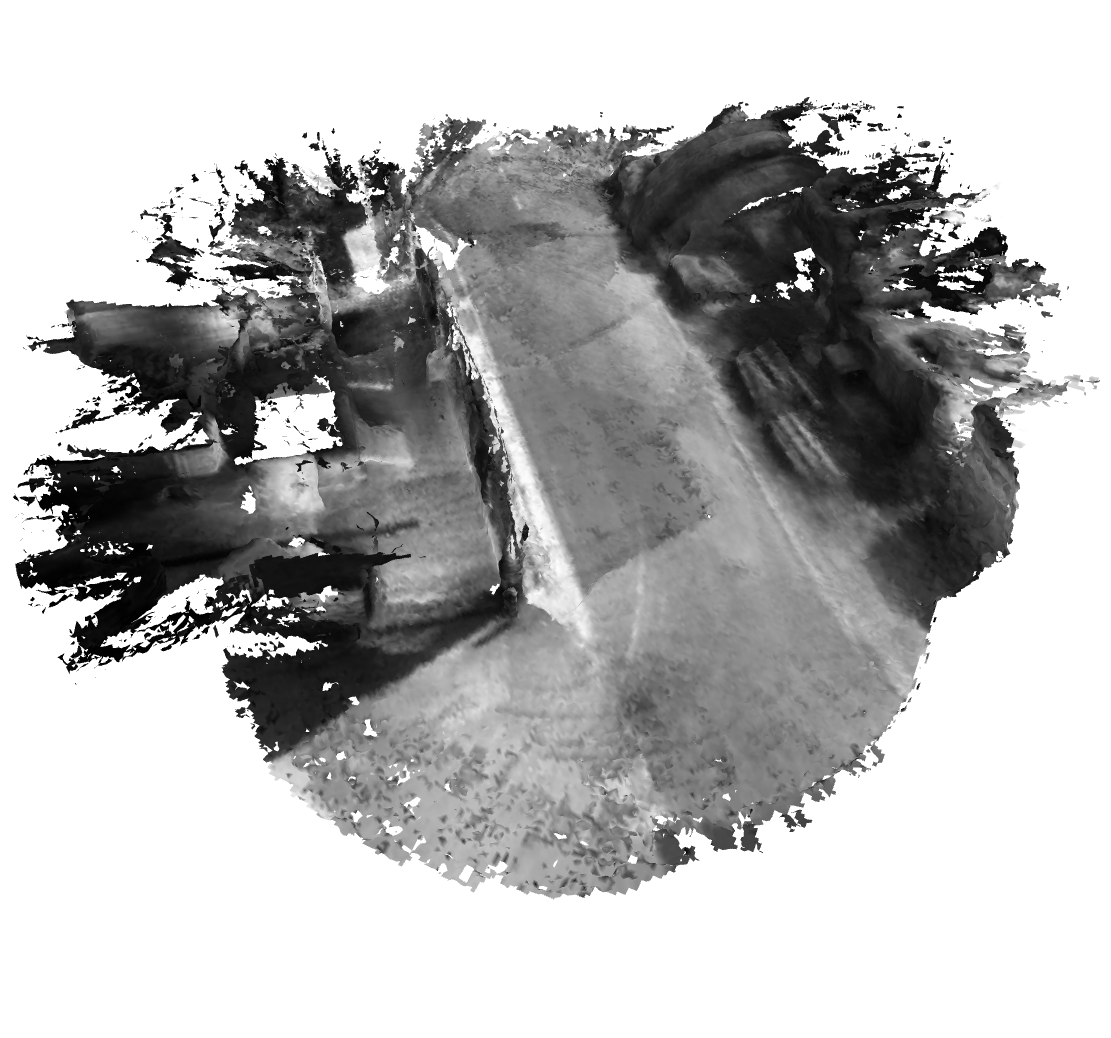}
  \caption{Shed (Stereo)}
  \label{fig:shed_stereo}
\end{subfigure}    

\begin{subfigure}[b]{0.45\columnwidth}
  \includegraphics[width=1.0\columnwidth,trim=0 80 0 200 mm, clip=true]{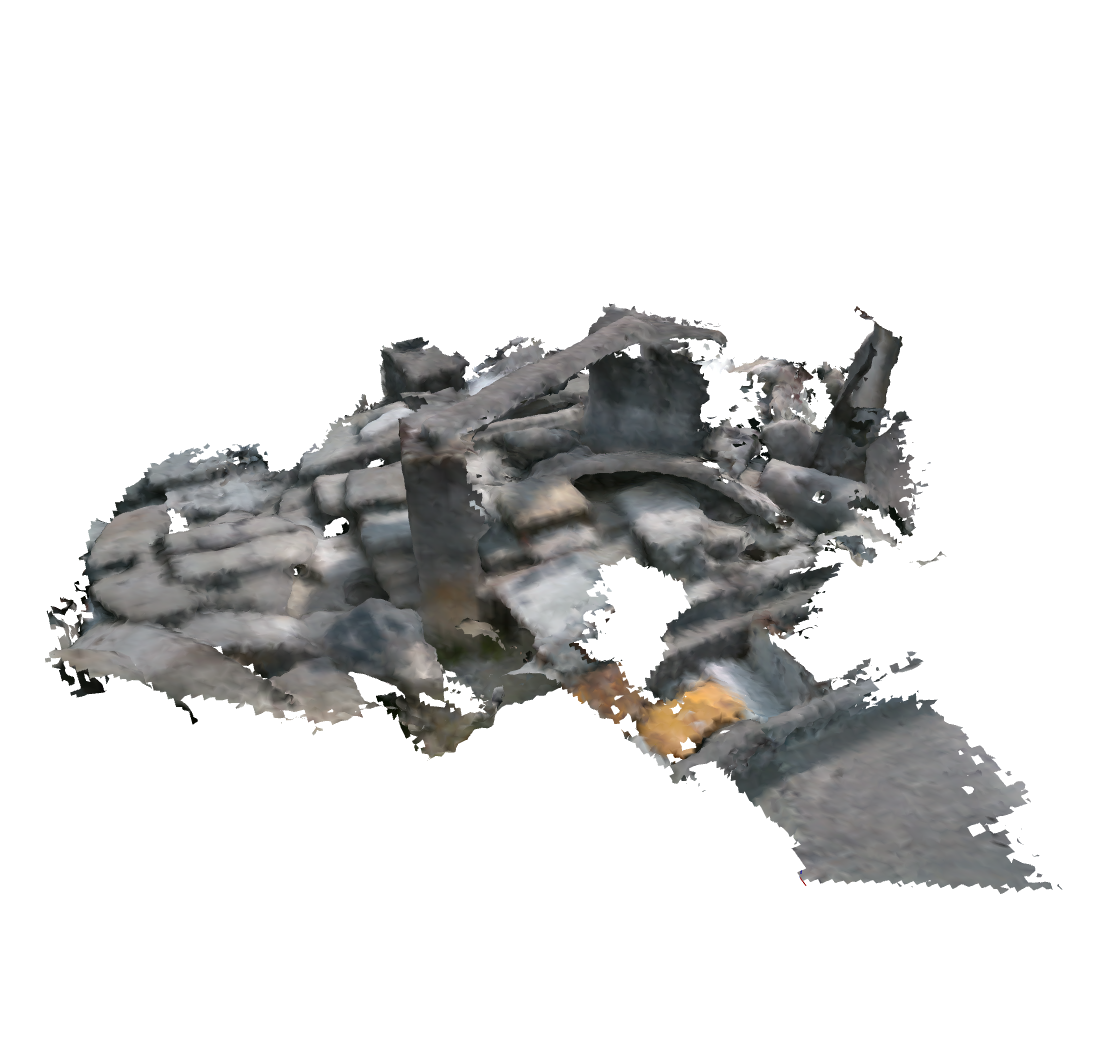}
  \caption{Rubble (RS)}
  \label{fig:rubble_rs}
\end{subfigure}    
\begin{subfigure}[b]{0.45\columnwidth}
  \includegraphics[width=1.0\columnwidth,trim=0 80 0 200 mm, clip=true]{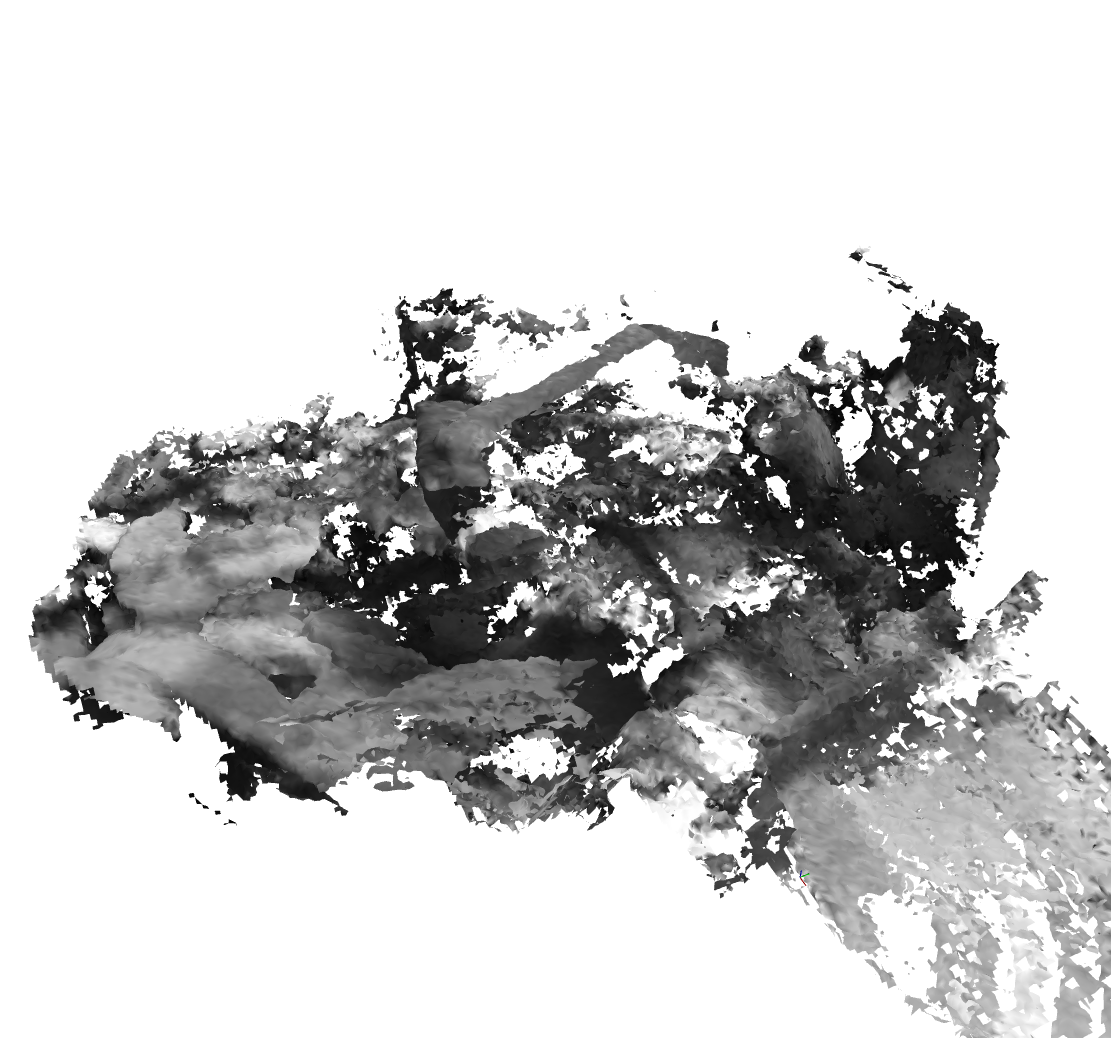}
  \caption{Rubble (Stereo)}
  \label{fig:rubble_stereo}
\end{subfigure}    

\begin{subfigure}[b]{0.45\columnwidth}
  \includegraphics[width=1.0\columnwidth,trim=0 0 0 80 mm, clip=true]{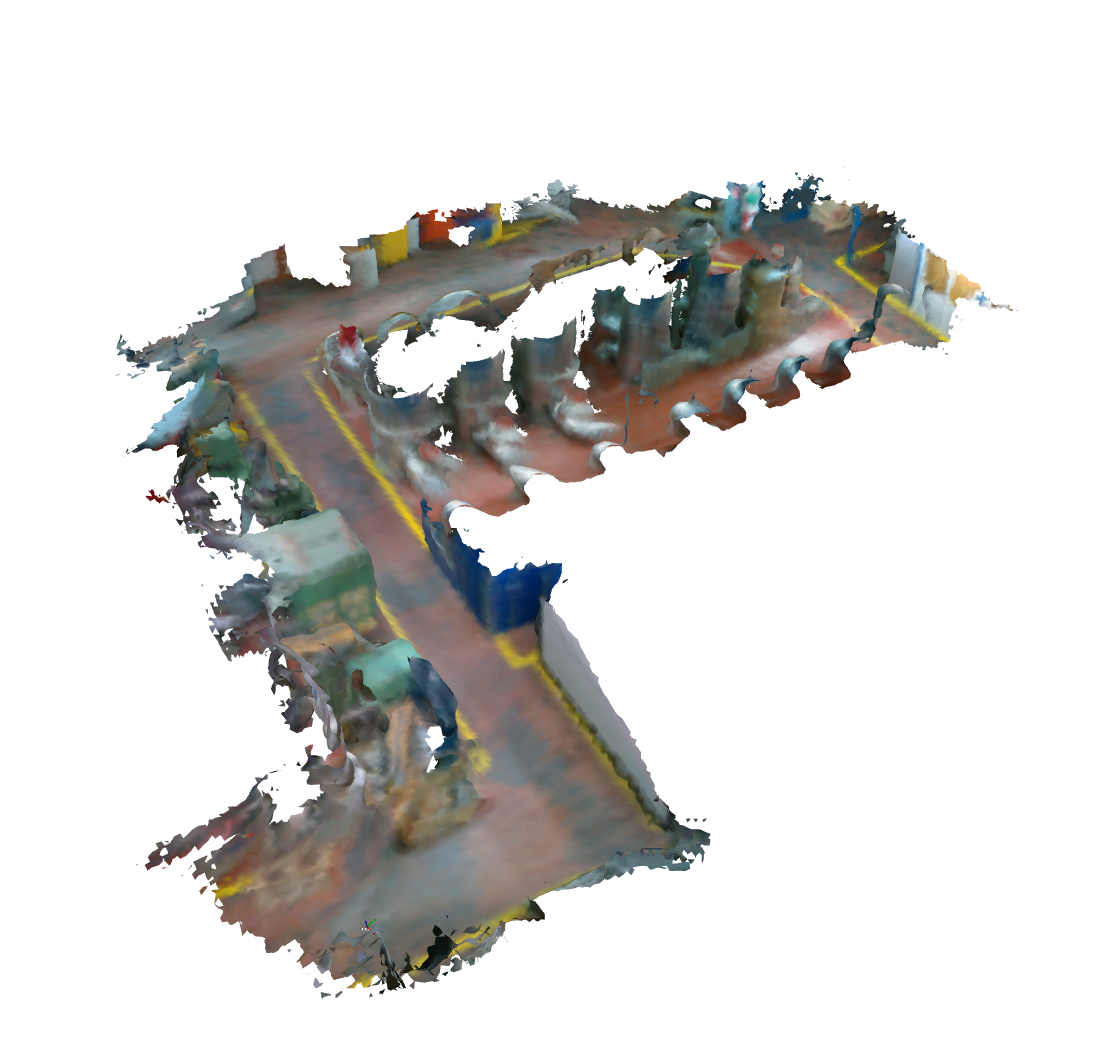}
  \caption{Machine Hall (RS)}
  \label{fig:machine_hall_rs}
\end{subfigure}    
\begin{subfigure}[b]{0.45\columnwidth}
  \includegraphics[width=1.0\columnwidth,trim=0 0 0 80 mm, clip=true]{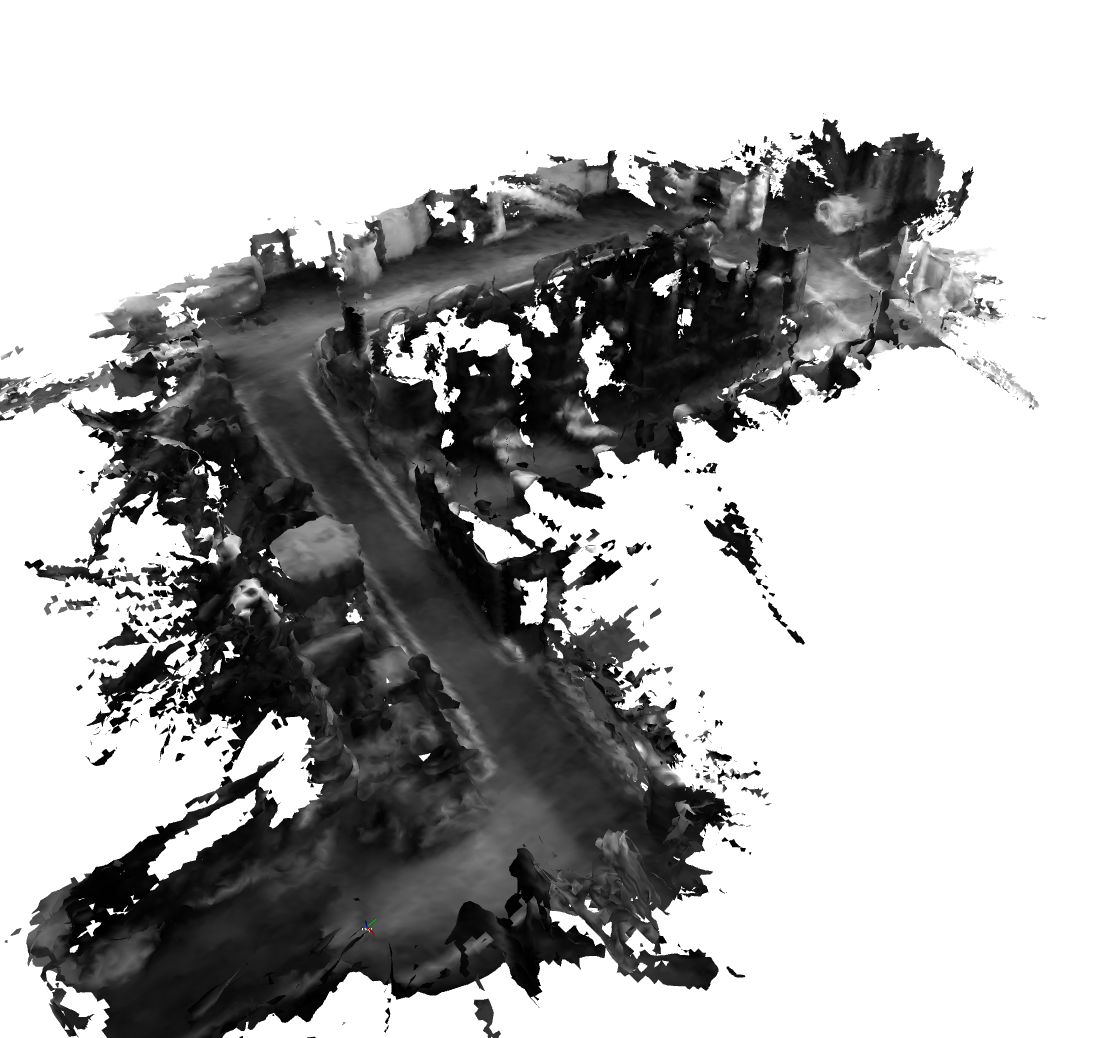}
  \caption{Machine Hall (Stereo)}
  \label{fig:machine_hall_stereo}
\end{subfigure}    
 
  \caption{Colored meshes of the realsense (RS) and stereo versions of the three maps used for benchmarking. Our stereo system is based on grayscale cameras, so no color data is available.}
  \label{fig:benchmark_screenshots}
\end{figure}

This is very apparent in the shed dataset, where especially in the outdoor space, the rubble is often outside of the field of view of the RealSense.
This effect is even visible in the meshes in \reffig{fig:shed_rs} versus (\subref{fig:shed_stereo}), where the RealSense never observes large parts of the pavement outside.

For other datasets, such as machine hall, this is less important because the structure is much closer and therefore the narrow FoV makes little difference.

\subsection{Sparse Topology Generation}
\label{sec:skeleton_eval}
\begin{figure}[tbp]
  \centering
  \includegraphics[width=0.6\columnwidth]{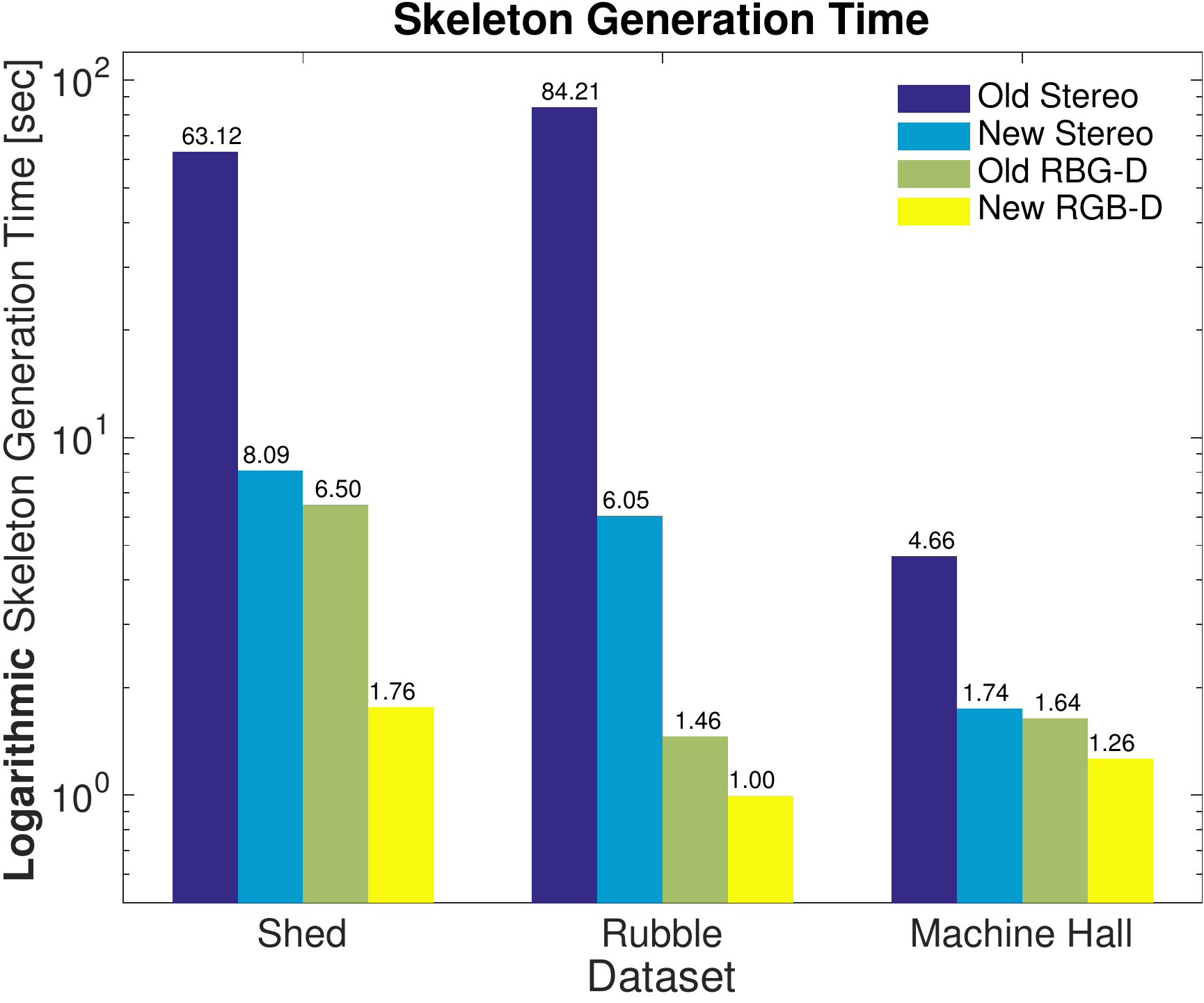}
  \caption{Sparse topological (skeleton) graph generation timings for the test datasets, compared between the ``old" approach presented in our previous work~\cite{oleynikova2018sparse} and the ``new" approach containing the improvements and extensions in this journal. Please note the logarithmic vertical scale. The stereo datasets generally take longer because there is more traversable space, and therefore more volume to process for any algorithm. In the stereo computations of the "old" approach, the sparse graph generation dominates the computation time.}
  \label{fig:skeleton_generation}
\end{figure}

To show that our sparse topology graph (or skeleton) is a feasible global planning strategy, we analyze the amount of time it takes to generate the sparse graph from an ESDF for each dataset, and compare the results to our previous proposed method from 2018~\cite{oleynikova2018sparse}.
The results are shown in \reffig{fig:skeleton_generation}, where the vertical scale is logarithmic.

The first thing to note is that the proposed changed in this journal decrease the runtime by up to an order of magnitude in the case of the shed and rubble stereo datasets.
And especially in the case of the large runtimes of those two datasets, the time is dominated by sparse graph generation and refinement, which are greatly simplified and sped up in our new approach.
In both old and new approaches, there is a large difference in the generation time between stereo and realsense for the machine hall and rubble datasets, due to how much more space is traversable in the stereo datasets (see \reffig{fig:traversability} for a qualitative representation of the difference).
However, most datasets are generated in 2 seconds or less, and the worst-case is 10 seconds using the new approach presented in \refsec{sec:sparse_topology}.
With the new sped-up method, we consider this very feasible for a pre-processing step for global planning, as the actual planning times are orders of magnitude faster than other methods.

\subsection{Loco}
\label{sec:loco_results}
\begin{figure}[tbp]
  \centering
  \includegraphics[width=0.6\columnwidth]{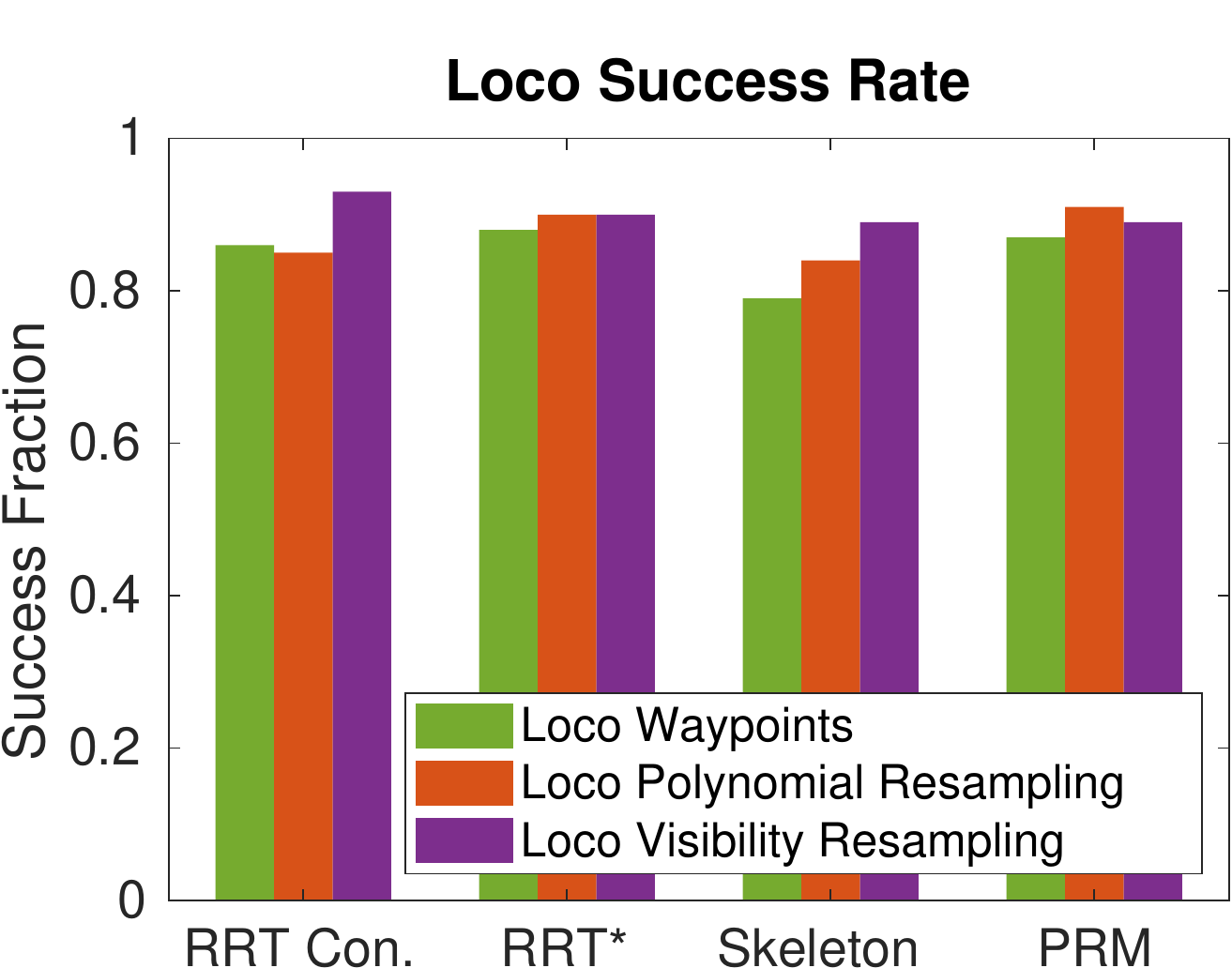}
  \caption{A comparison of different waypoint fitting methods for Loco, and their success rate on the shed stereo dataset given different global planning methods. In general, visibility resampling has the highest success rate.}
  \label{fig:loco_results}
\end{figure}

We analyze the effect of different waypoint fitting methods for Loco, as described in \refsec{sec:loco}.
We compare three methods: waypoint fitting (where each pair of waypoints has a segment between them), polynomial resampling (where an initial polynomial trajectory is fit with one segment through each waypoint, then resampled to a fixed number of waypoints), and visbility graph resampling (where intermediate waypoints are sampled directly off the visibility graph).
The results are shown in \reffig{fig:loco_results}, evaluated on the stereo shed dataset.

The visibility graph resampling has the highest success rate, and we use that variant as `Loco' for the remaining evaluations.

\subsection{Global Planning Benchmarks}
To demonstrate the differences between different global planning methods, and how choice of path smoother affects the final result, we run 100 trials on all provided datasets.
Each trial starts and ends at a random location, a minimum of 2 meters apart.
The robot radius is 0.5 meters for all planners.

We use multiple global planning methods, summarized below:
\begin{description}
  \item[None] Straight-line path between start and goal, no planning, meant to give an estimate of how many of the test cases have trivial solutions.
  \item[RRT Conn.] RRT-Connect, which grows a bi-directional tree from and toward the goal. Very fast, ran with an upper time bound of 1.0 sec (though terminates when first solution is found).
  \item[RRT*] Probabilistically-optimal random planner, ran for 2.0 seconds.
  \item[Skeleton] Our sparse topological planner, using path shortening on the output path.
  \item[PRM] Probabilistic roadmap, aimed to compare versus our skeleton-based method. The pre-planning stage is run for 2.0 seconds (to mirror the average dataset processing time of the sparse topology), and each planning query is given an additional 0.1 seconds.
\end{description}

Likewise, we test a variety of path smoothing methods:
\begin{description}
  \item[No Smoothing] Not an actual path smoothing method, just an indicator showing whether the global planner succeeded or not.
  \item[Velocity Ramp] Velocity ramp method, always applying maximum or no acceleration. Follows straight-line paths between waypoints.
  \item[Polynomial] The polynomial splitting approach of~\cite{richter2013polynomial}, described in sections above. 
  \item[Loco] Our local continuous trajectory optimizaton algorithm, run with visibility waypoint re-sampling, as determined from the previous sections to be the best.
\end{description}

\begin{figure}[tbp]
  \centering  
  \includegraphics[width=0.6\columnwidth,trim=0 0 0 0 mm, clip=true]{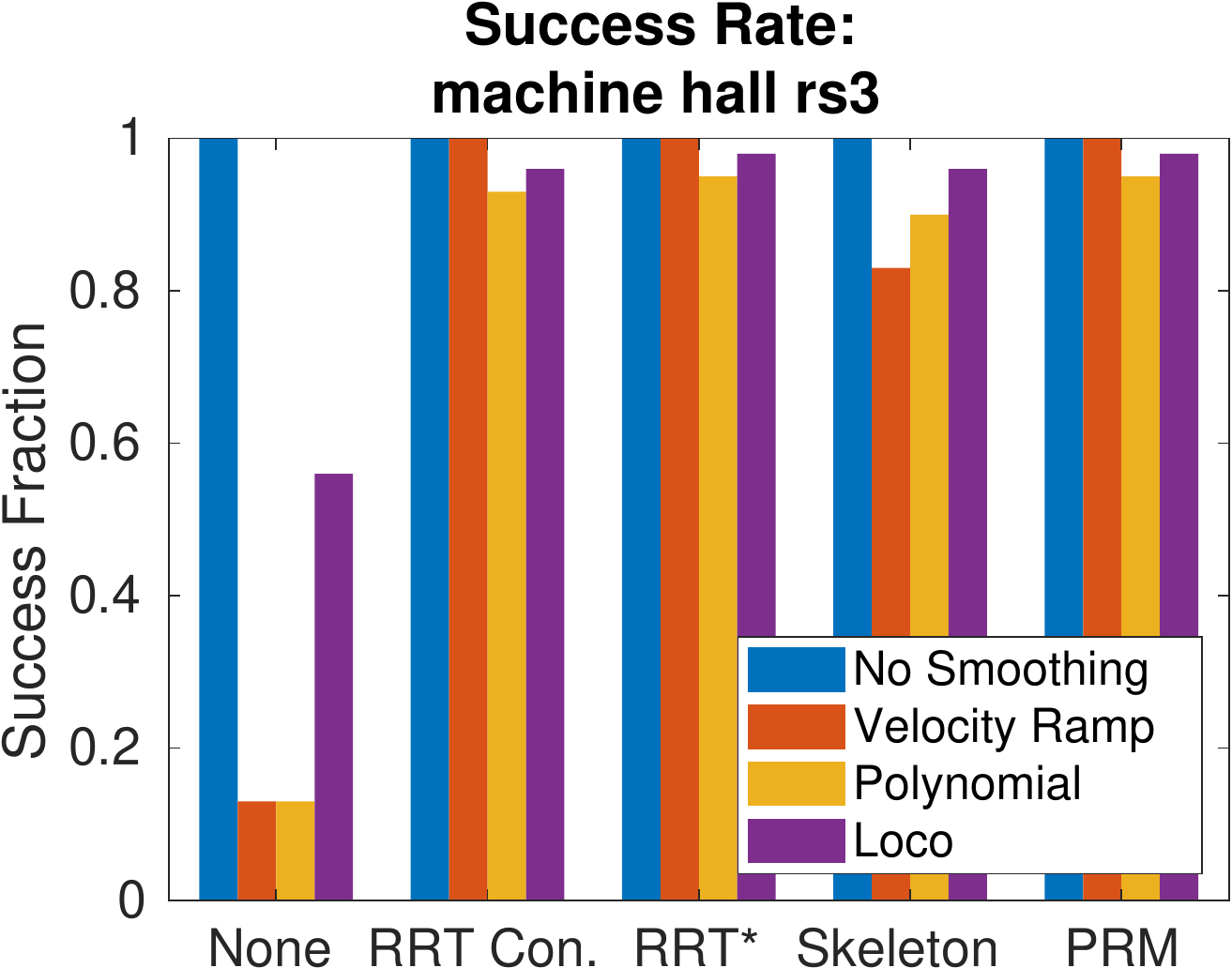}  
  \caption{Success rate of various global planner and path smoothing methods on the Machine Hall Realsense dataset, showing our method (Loco) is able to give smooth dynamically-feasible paths for more tests cases than competing methods.}
  \label{fig:success_machine_hall_rs}
\end{figure}

\reffig{fig:success_machine_hall_rs} shows a comparison of all described methods on the Machine Hall Realsense dataset.
There are a number of take-aways from these results.
When not using a global planner (i.e., attempting to draw a straight-line path between start and goal), only 13\% of the test cases have trivial solutions, but Loco is able to solve 56\% of problems with no global plan.
In general, the success rate of Loco is also slightly higher, as it is able to better utilize the information in the map.

All the global planners are able to solve all of the planning problems.
One key point to note is that the velocity ramp method does not work very well with the topological skeleton planner: this is because our graph simplification method contains some edges that do not lie perfectly on the straight-line.
However, the Loco planner has a comparable success rate with the skeleton planner as other planners, again because it can follow gradient information in the map and slightly perturb the waypoints to produce a collision-free path.

\begin{figure}[tbp]
  \centering
  \includegraphics[width=0.6\columnwidth]{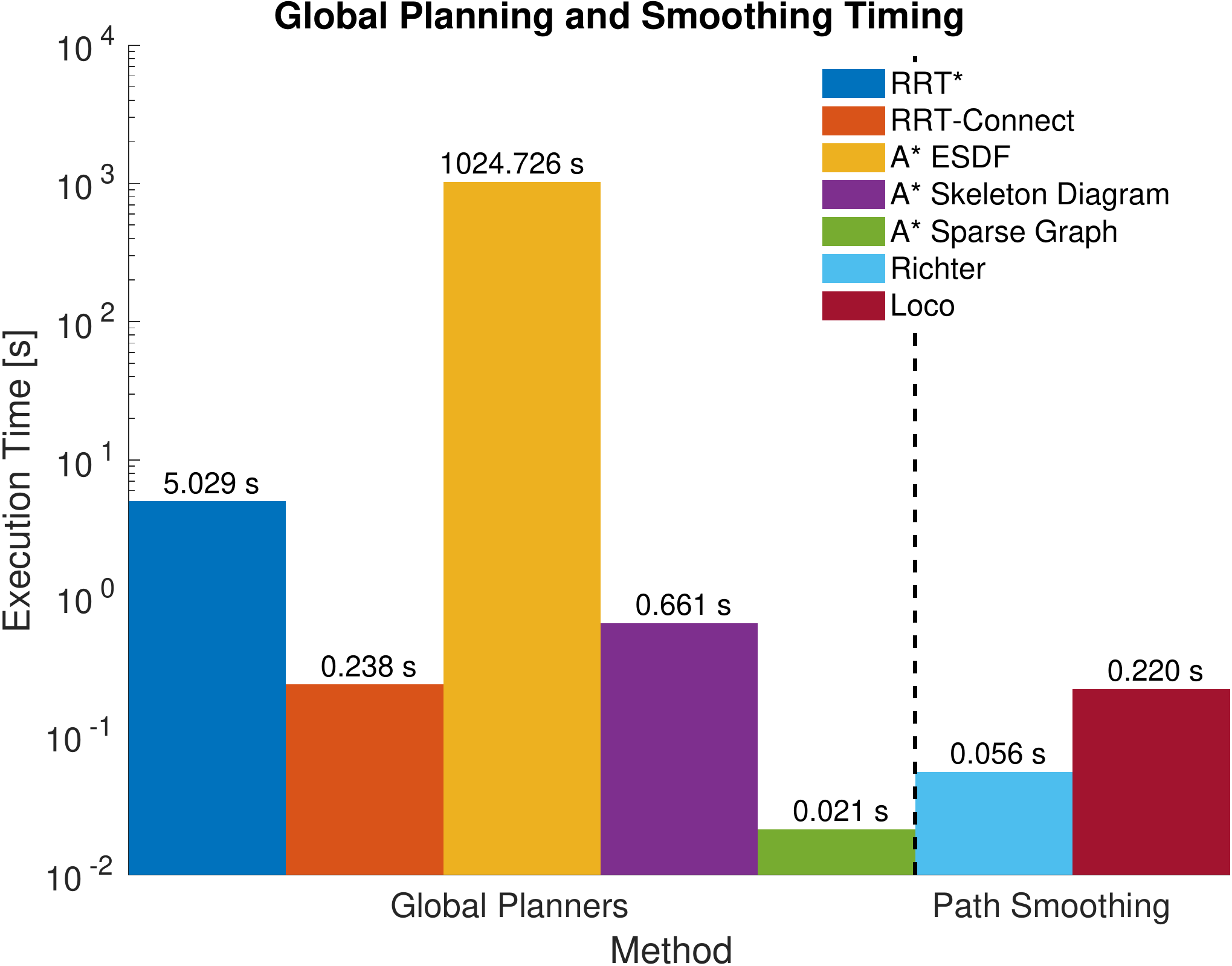}
  \caption{Timings for various global and local planning methods. Note the log scale. As can be seen, skeleton planning is at least one order of magnitude faster than other global planning methods, and while loco timing is slightly slower than polynomial, it is still within bounds for fast global planning applications.}
  \label{fig:rrt_loco_timing}
\end{figure}

The timings for a single typical trajectory on the Shed stereo dataset are shown in \reffig{fig:rrt_loco_timing} (note the log scale).
The topological skeleton planning method is 10x faster than even RRT Connect (and produces much shorter path lengths).
Both the polynomial and loco smoothing methods are acceptable for fast global planning, though the polynomial method is approximately 2x faster. 

\subsection{Local Planning Benchmarks}
\label{sec:local_planning_benchmarks}
In order to evaluate the complete local planning approach, we repeat a series of simulated evaluations similar to those in \cite{oleynikova2018safe}.
We focus specifically on extremely cluttered environments, such as the random Poisson forests   introduced by Karaman \etal~\cite{karaman2012high}.

We generate a map that is $15 m \times 15 m \times 5 m$ with a random distribution of cylindrical obstacles.
The density of the obstacles varies between $0.1$ objects per square meter to $0.5$ objects per square meter, which corresponds approximately to $10\%$ of the volume being occupied to $50\%$. 
To simplify the testing procedure, a $2$ meter long section on the left and right side of the map is always left obstacle-free, allowing us to set the start and goal positions in the same location in each trial.

\begin{figure}[tbp]
  \centering  
  \includegraphics[width=0.4\columnwidth,trim=0 0 200 200 mm, clip=true]{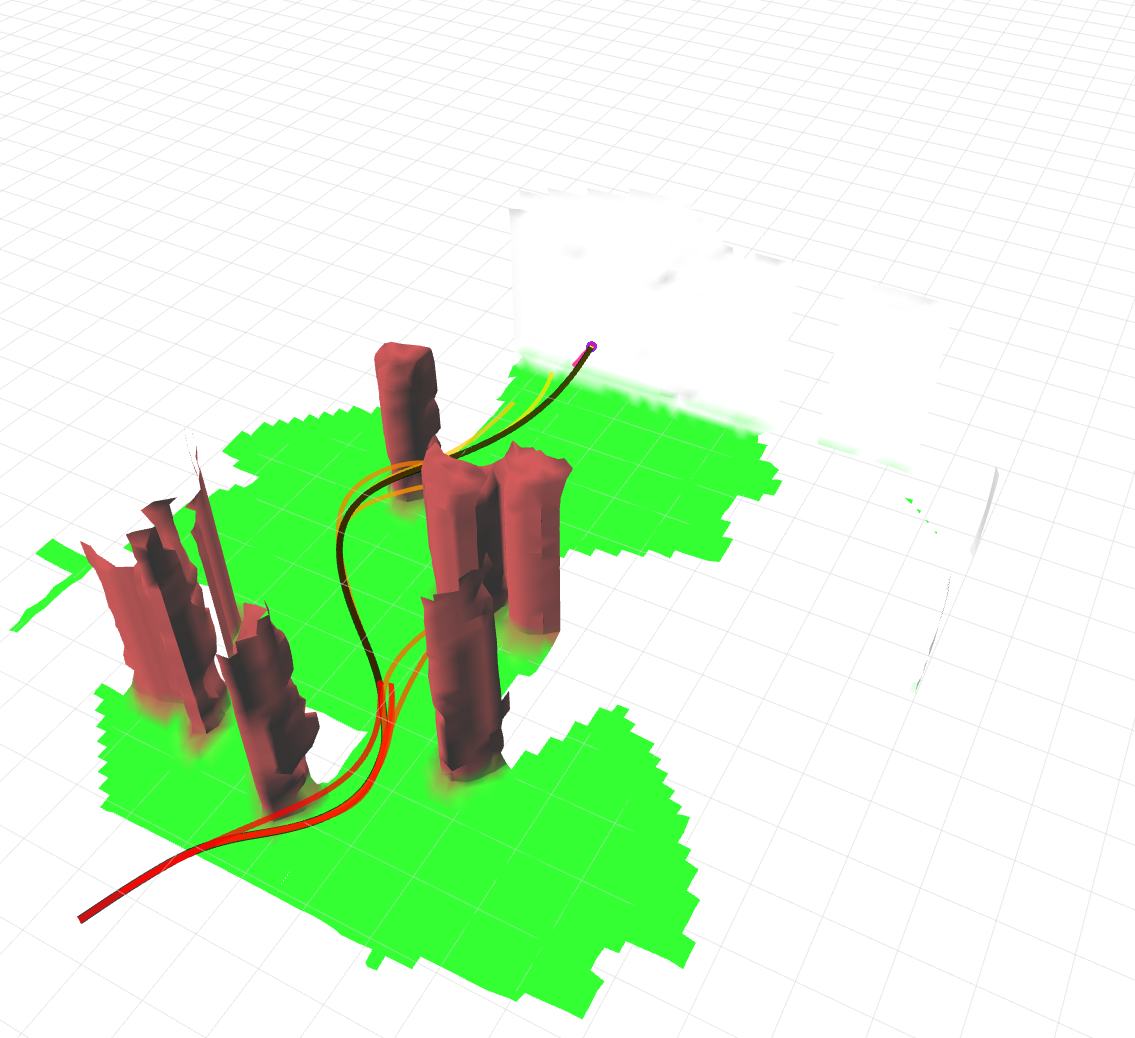}  
  \caption{An example of the environment used by the local benchmark, with an obstacle density of $0.2 \textrm{ objects}/m$, which roughly corresponds to $20\%$ of the volume being occupied. The incremental plans of the MAV are shown in color, from red to yellow over time, and the final executed path is shown in black.}
  \label{fig:local_benchmark_screen}
\end{figure}

Each trial consists of a maximum of 60 simulated steps, each corresponding to a simulated second of flight.
At the beginning of each iteration, the simulated MAV observes its environment once, with a forward-facing sensor with a $320 \times 240$ pixel resolution and a $90\deg$ horizontal field of view.
This data is then incorporated into a voxblox map, and we run one step of the local planner described in \refsec{sec:local_replanning}.
A sample test environment, as explored by the MAV, and a path between the start an goal is shown in \reffig{fig:local_benchmark_screen}.

\begin{figure}[tbp]
  \centering 
  \begin{subfigure}[b]{0.49\columnwidth}
    \includegraphics[width=1.0\columnwidth,trim=0 0 0 0 mm, clip=true]{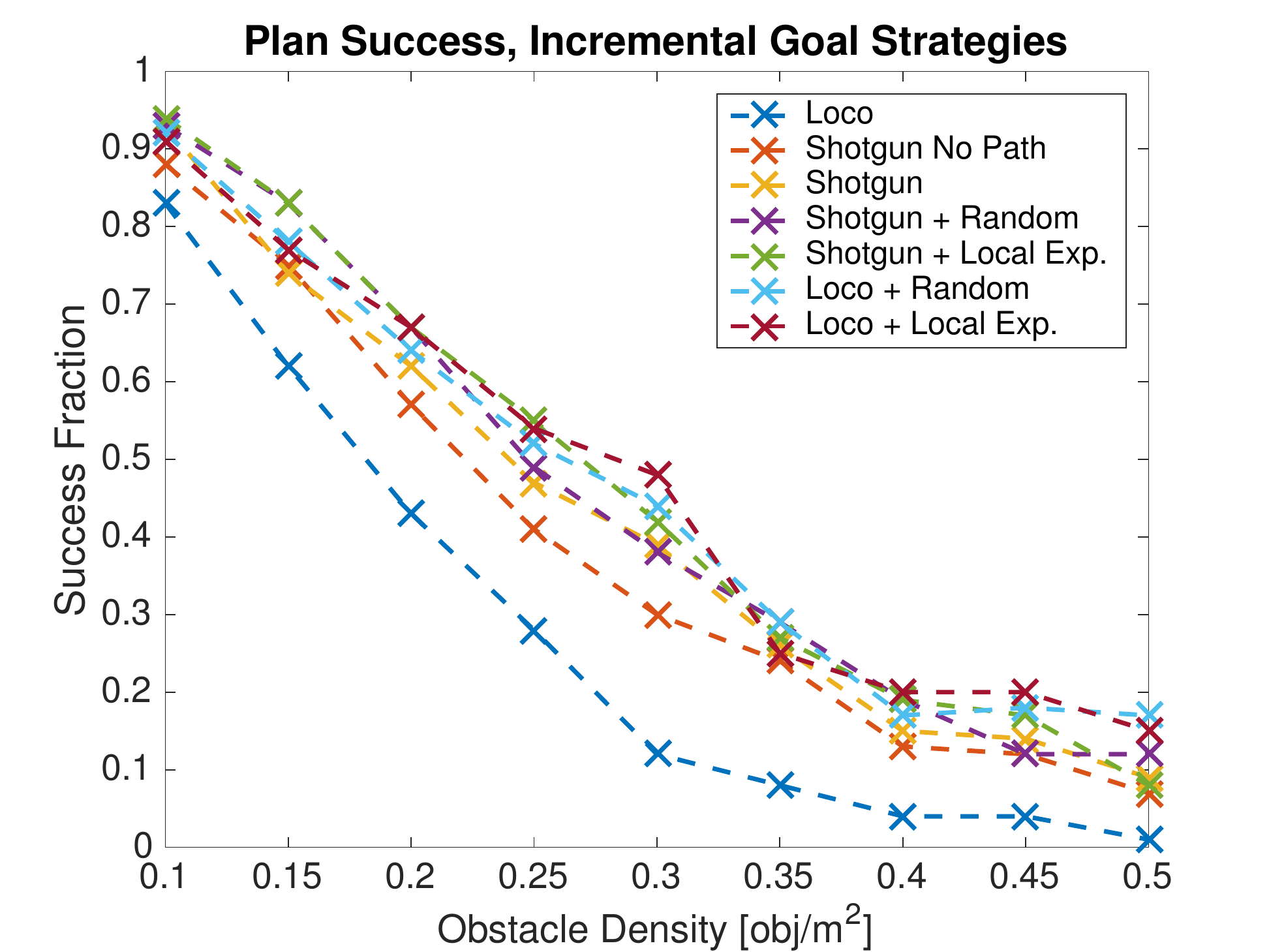}
    \caption{}
    \label{fig:local_benchmark_success}
  \end{subfigure}    
  \begin{subfigure}[b]{0.49\columnwidth}
    \includegraphics[width=1.0\columnwidth,trim=0 0 0 0 mm, clip=true]{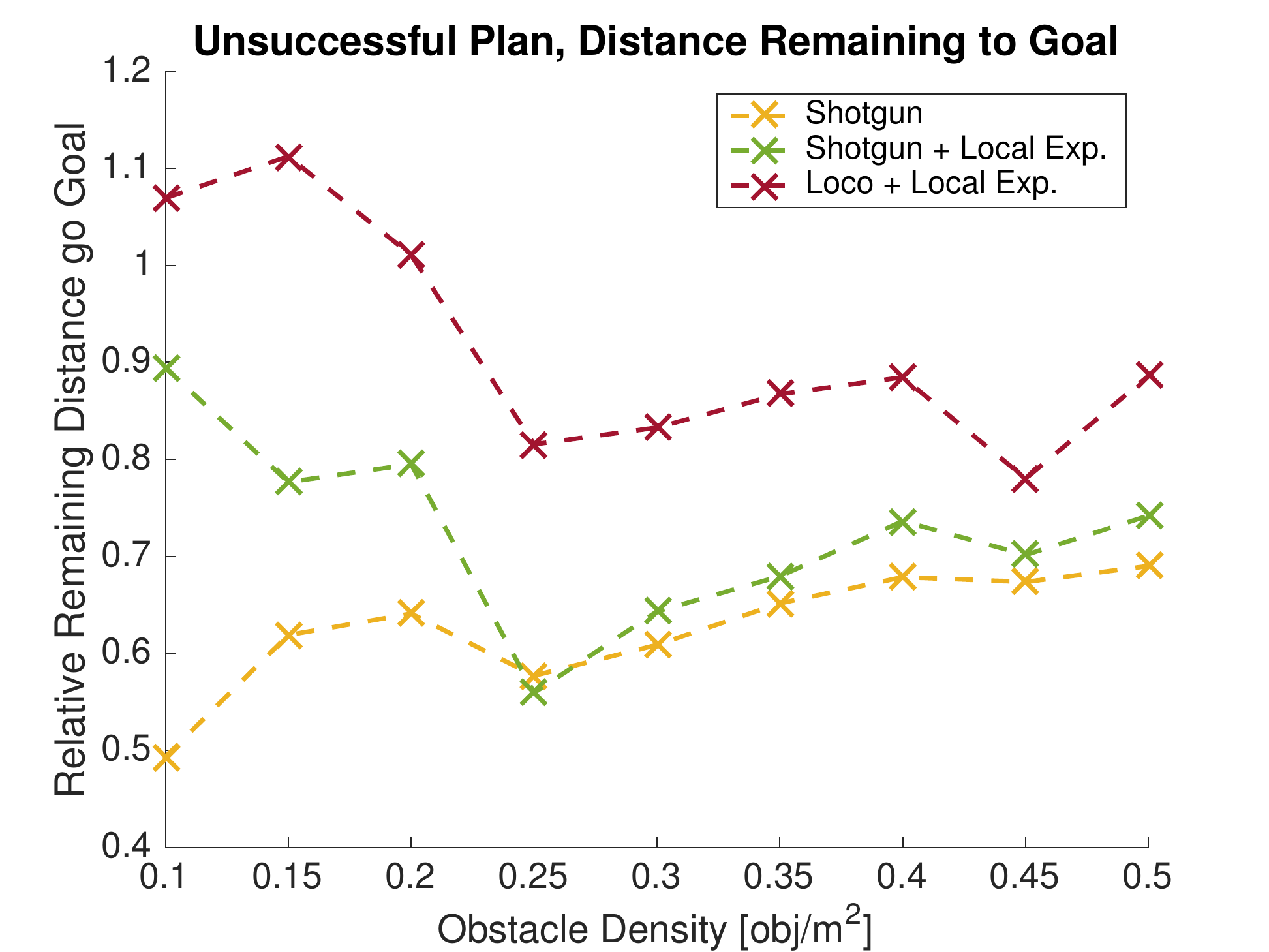}
    \caption{}
    \label{fig:local_benchmark_length}
  \end{subfigure}    
  \caption{(\subref{fig:local_benchmark_success}) shows success rate of multiple variants of our local planner. Loco is the original Loco method presented in \refsec{sec:loco}, Shotgun No Path \refsec{sec:shotgun} uses the shotgun goal selection but without using its path to hotstart the optimization, Shotgun uses the hotstarts, and the other methods use the intermediate goal strategies described in \refsec{sec:intermediate_goal}. As can be seen, using the Shotgun selection significantly increases the success rate of the naive planner, but its advantages in success rate are cancelled out by intermediate goal strategies. In contrast, \subref{fig:local_benchmark_length} shows for the cases that all three methods \textit{failed}, how far the robot was from the goal (lower is better). In these cases, shotgun methods outperform the loco methods and local exploration - as the shotgun methods are able to better exploit the seen free space.}
  \label{fig:local_benchmark_results}
\end{figure}

We ran the evaluations for a total of 1000 trials per method, amounting to 100 trials per density.
The aggregated results are shown in \reffig{fig:local_benchmark_results}.
The ``Loco'' method presented is the raw Loco method from \refsec{sec:loco}, without any initial end-point selection or intermediate goal strategies, which as can be seen performs the worst.
Using the Shotgun initial end-point selection described in \refsec{sec:shotgun} substantially improves success rate of the raw Loco method, though if the Shotgun path is not used for hotstarting the optimization (in the Shotgun No Path method), the performance is slightly worse than when using the path.

What is interesting to note, however, is that the last four methods, combining Shotgun and Loco with the intermediate goal strategies described in \refsec{sec:intermediate_goal}, all perform similarly in terms of success rate in \reffig{fig:local_benchmark_success}.
In our previous work~\cite{oleynikova2018safe}, we saw the same effect between the random and local exploration strategies, but the local exploration had significantly shorter path lengths.
However, the similar performance between the Shotgun strategies and the Loco strategies when augmented with intermediate goal selection suggests that they perform largely the same role as the Shotgun initial end-point selection.

However, the shotgun method has one additional advantage in complex cases, which may not have a feasible path to the goal.
Since it exploits the seen free space by design, it is able to get closer to the goal than other strategies, even in cases when it cannot reach it.
This is shown in \reffig{fig:local_benchmark_length}, where we compare three methods in the cases where planning was \textit{not successful}, and they were not able to reach the goal.
We compare the remaining relative path lengths to the goal: that is, compared to the total straight-line distance to the goal, how much straight-line distance to the goal remains at the time of planning failure.
This demonstrates the property of predictability that the Local Exploration methods trade-off: in some cases, ``Loco + Local Exploration'' actually ends the planning farther from the goal than it started, while the Shotgun-based methods always reduce the distance significantly.
Shotgun alone, while not the best in terms of overall success, is the best at getting close to the goal when it is unable to plan farther.
Note that for fairness of comparison, \reffig{fig:local_benchmark_length} is shown only on cases where \textit{all three methods} failed.

This suggests that in scenarios where a randomized goal selection strategy such as random or local exploration is suitable (such as exploring empty environments without people, or where the entire space is safe to enter), the Shotgun strategy is not necessary.
On the other hand, in cases where an operator would like to have more high-level control over the motion of the drone and less unpredictability, the Shotgun strategy has most of the same advantages while having the robot behave predictably.
Additionally, in cases where it may not be possible to reach the target, shotgun will get closer than other approaches, potentially allowing inspection of areas that may not otherwise be reachable or visible.
This is especially relevant in the earthquake-damaged building scenario, where it may be desirable to get inaccessible areas within sensing range to find potential buried victims.

\subsection{Platform Experiments}
While most of the evaluations in the previous sections are on real datasets collected with the MAV, allowing us to quantitatively test the performance of global planning and local smoothing, in this section we aim to show full closed-loop on-board experiments, showcasing the complete system.

\begin{figure}[tbp]
  \centering  
  \includegraphics[width=0.56\columnwidth,trim=0 0 0 0 mm, clip=true]{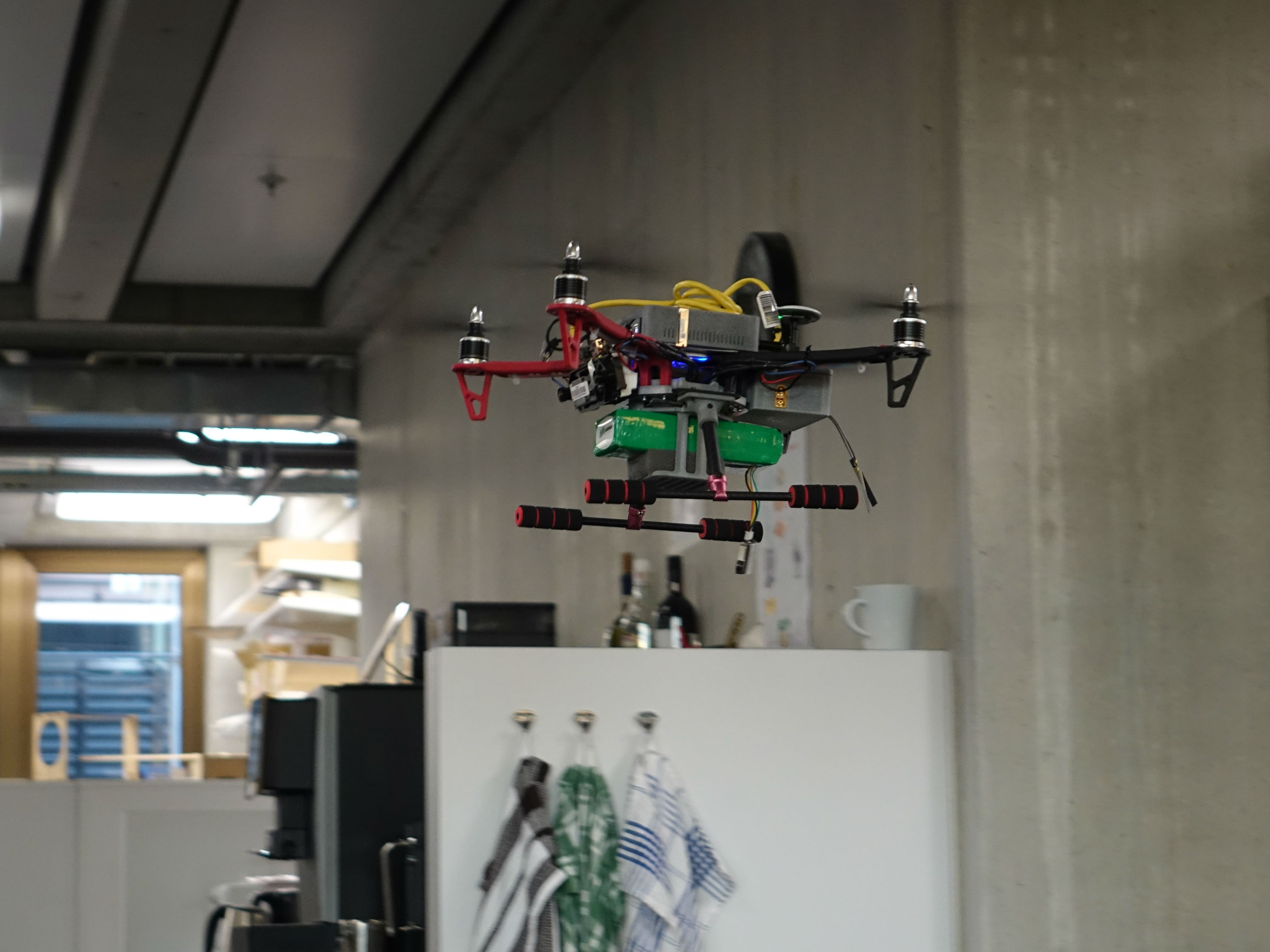}  
  \caption{The platform used for the closed-loop experiments, a smaller quadcopter based on a DJI F450 frame with a stereo visual-inertial sensor~\cite{nikolic2014synchronized} mounted to the front for both state estimation and depth measurements. Here it is shown flying through a cluttered office space in Experiment 1.}
  \label{fig:experiments_platform}
\end{figure}

The tests were performed on a platform similar to the hexacopter described in \refsec{sec:hardware} and shown in \reffig{fig:jay_system}.
For these experiments, we used a quadcopter based on the DJI F450 frame (as we have lower payload requirements due to the minimal sensor setup), a Pixhawk flight controller, an Intel NUC for on-board processing, and a stereo visual-inertial sensor~\cite{nikolic2014synchronized} for both state estimation and depth data.
The system is shown flying in a lab environment in \reffig{fig:experiments_platform}. 
Instructions, part files, and a description and links to the complete software stack are available online\footnote{\url{github.com/ethz-asl/mav_tools_public}}.

\begin{figure}[tbp]
  \centering
  \begin{subfigure}[b]{0.24\columnwidth}
    \includegraphics[width=1.0\columnwidth,trim=0 0 0 0 mm, clip=true]{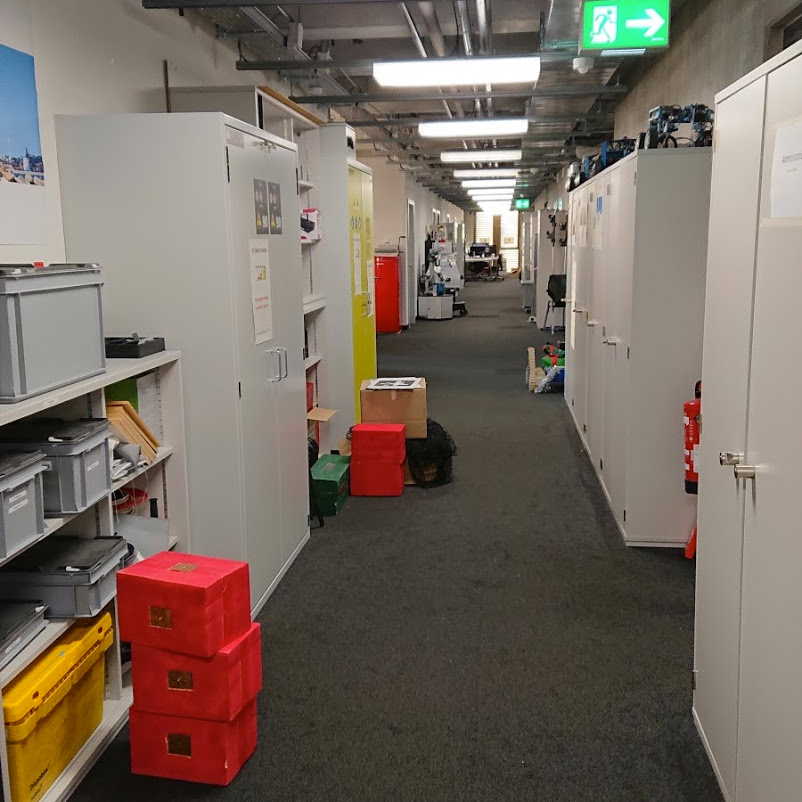}
    \caption{Office}
    \label{fig:office_before}
  \end{subfigure}
  \begin{subfigure}[b]{0.24\columnwidth}
    \includegraphics[width=1.0\columnwidth,trim=0 0 0 0 mm, clip=true]{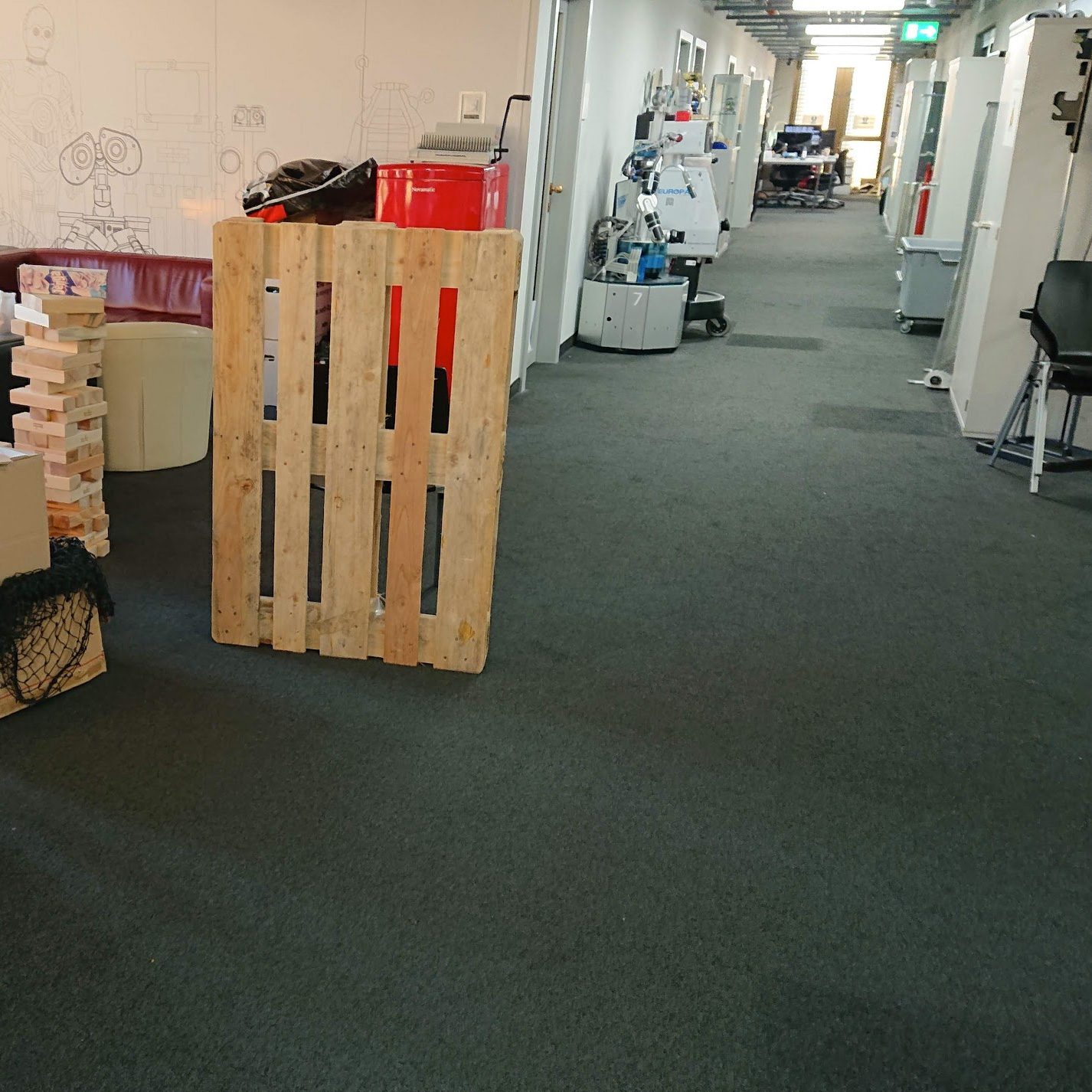}
    \caption{Office + Obstacle}
    \label{fig:office_after}
  \end{subfigure}  
  \begin{subfigure}[b]{0.24\columnwidth}
    \includegraphics[width=1.0\columnwidth,trim=0 0 0 0 mm, clip=true]{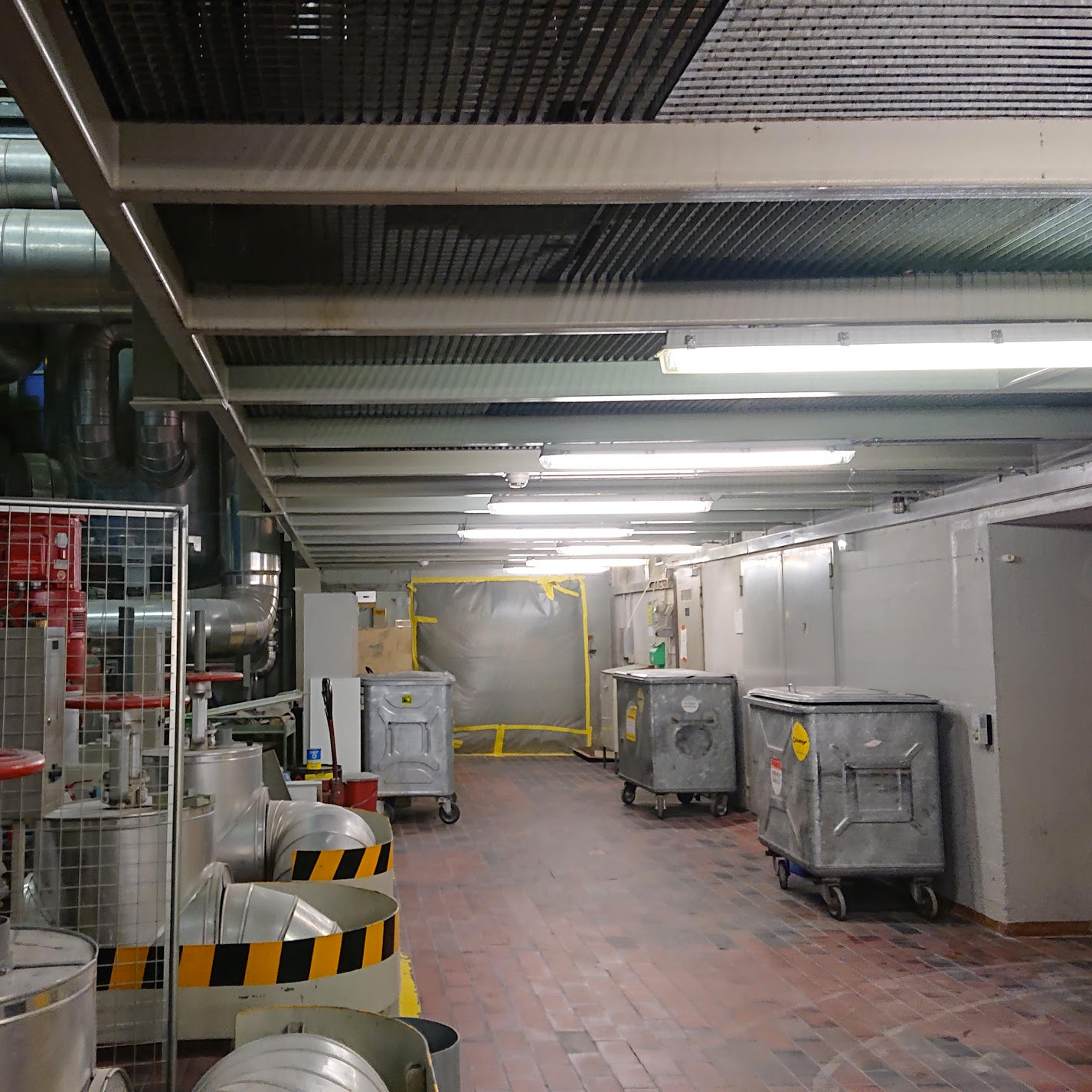}
    \caption{Machine Hall}
    \label{fig:machine_hall_before}
  \end{subfigure}
  \begin{subfigure}[b]{0.24\columnwidth}
    \includegraphics[width=1.0\columnwidth,trim=0 0 0 0 mm, clip=true]{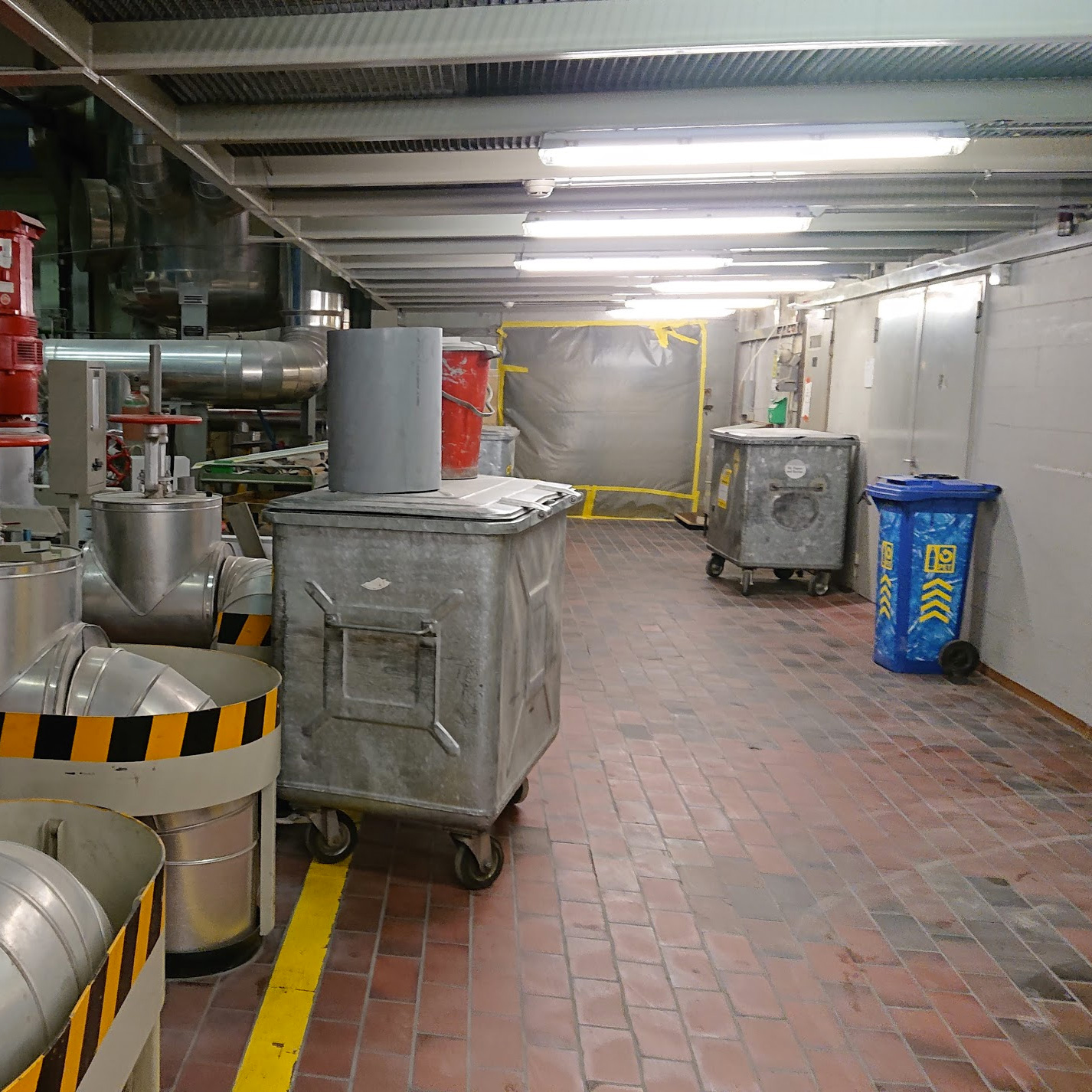}
    \caption{Machine Hall + Obs.}
    \label{fig:machine_hall_after}
  \end{subfigure}
  \caption{Photos of scenes that the local planner experiments were performed in, with and without the added obstacle. The obstacles were used in trials where the global planner uses a previously-built map, to verify that the local planner is able to cope with changes to the global plan in the presence of obstacles.}
  \label{fig:experiment_photos}
\end{figure}

The experimental validation consisted of 6 experiments, focusing on show-casing the local planner and its interaction with a global planner.
Experiments were done in two scenes; one in a cluttered robotics office, shown in \reffig{fig:office_before} and a machine hall, similar section to what was showcased in the Machine Hall evaluation dataset in \refsec{sec:datasets} and shown in \reffig{fig:machine_hall_before} and aimed to mimic the industrial inspection scenario.

In every experimental set-up, the MAV starts out with an empty local map.
Optionally, it may either build its own global map (i.e., use the same map for both the local and the global planner), or use a previously-built global map.
In the latter case, we always add some obstacles that are not present in the original global map to show that the local planner always verifies any global trajectories against the current state of its local map.
The obstacles we add to both scenes are shown in \reffig{fig:office_after} and \reffig{fig:machine_hall_after}, and are placed directly in the line the planner normally took to the goal.
If we use the same map for local and global planning, since we do not do any online localization, the odometry may have drifted during flight. Being able to return safely even in a drifting map again showcases the safety of our local planner framework.

\begin{figure}[tbp]
  \centering
  \begin{subfigure}[b]{0.31\columnwidth}
    \includegraphics[width=1.0\columnwidth,trim=0 50 70 240 mm, clip=true]{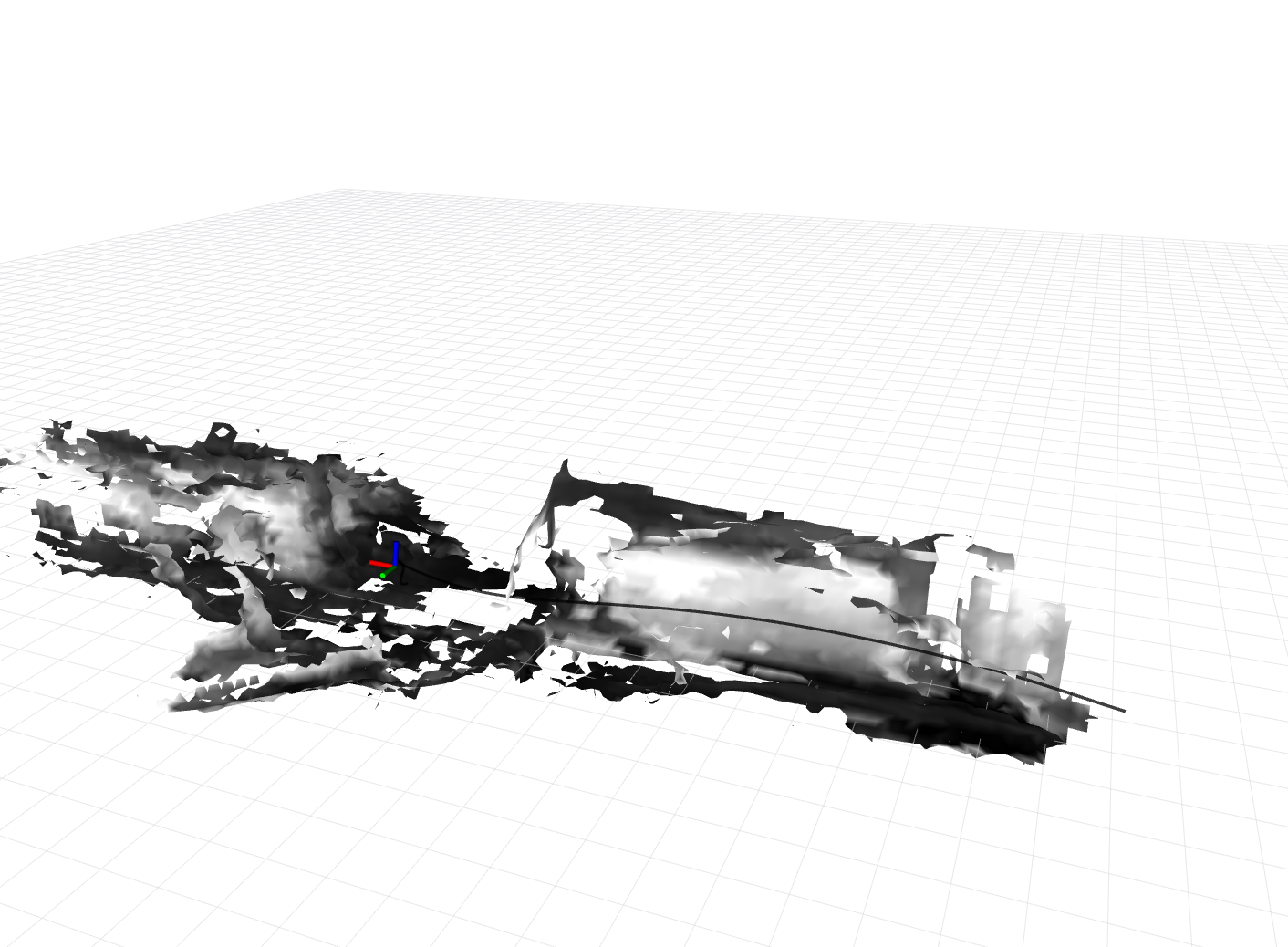}
    \caption{Exp. 1}
    \label{fig:exp1}
  \end{subfigure}
  \begin{subfigure}[b]{0.31\columnwidth}
    \includegraphics[width=1.0\columnwidth,trim=0 50 70 240 mm, clip=true]{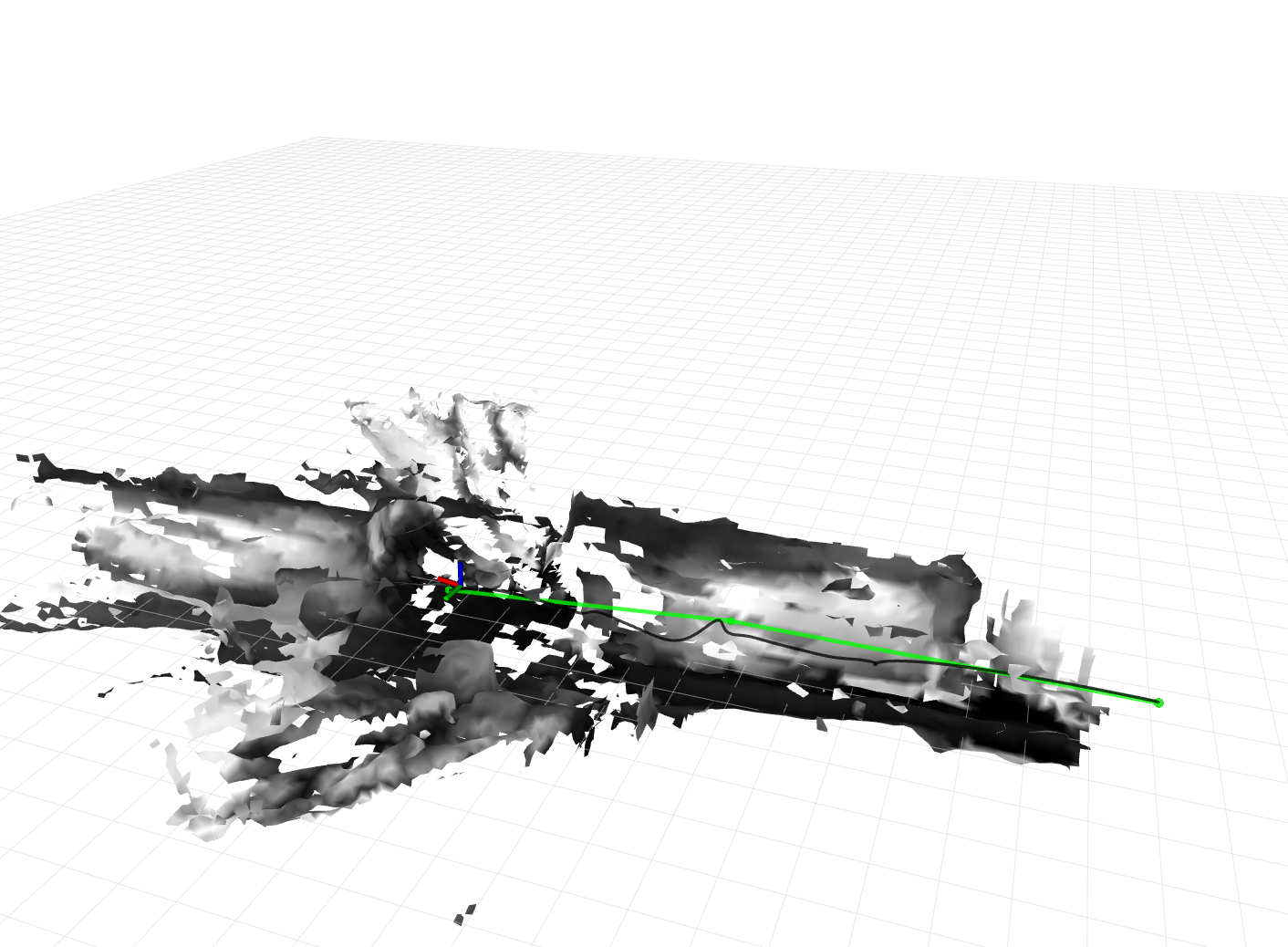}
    \caption{Exp. 2}
    \label{fig:exp2}
  \end{subfigure}  
  \begin{subfigure}[b]{0.31\columnwidth}
    \includegraphics[width=1.0\columnwidth,trim=0 50 70 240 mm, clip=true]{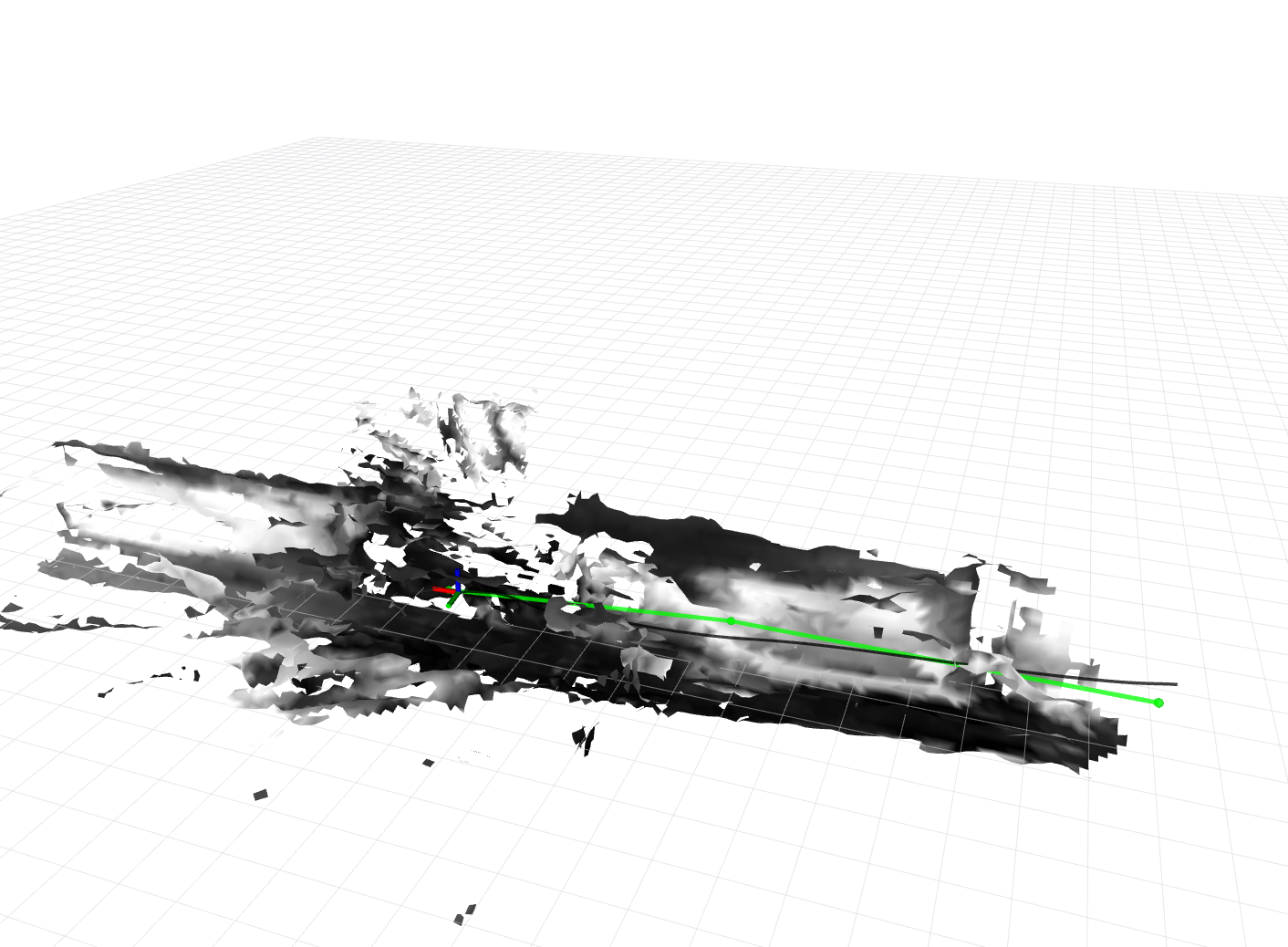}
    \caption{Exp. 3}
    \label{fig:exp3}
  \end{subfigure}
  \begin{subfigure}[b]{0.31\columnwidth}
    \includegraphics[width=1.0\columnwidth,trim=0 00 240 120 mm, clip=true]{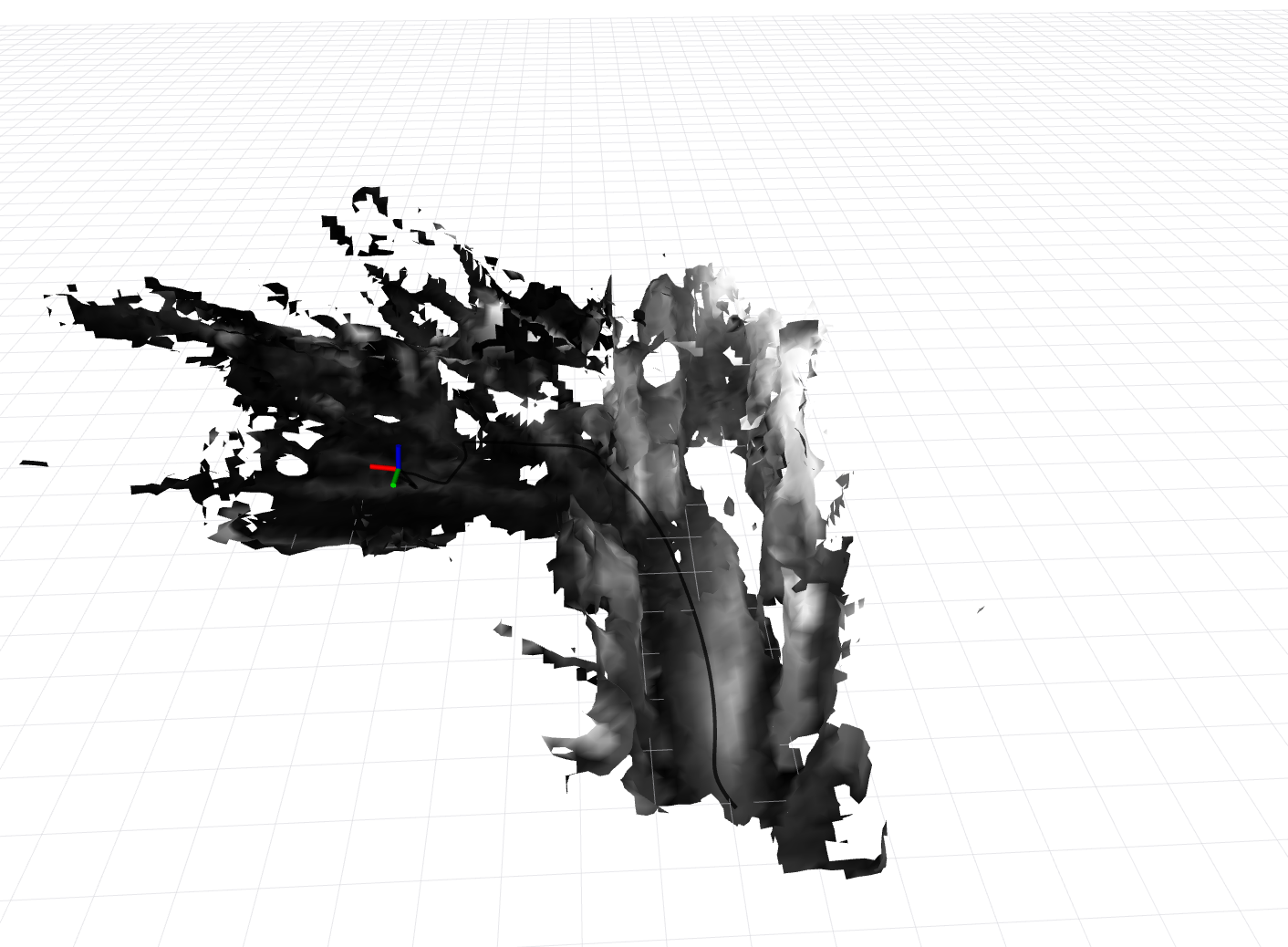}
    \caption{Exp. 4}
    \label{fig:exp4}
  \end{subfigure}
  \begin{subfigure}[b]{0.31\columnwidth}
    \includegraphics[width=1.0\columnwidth,trim=0 00 240 120 mm, clip=true]{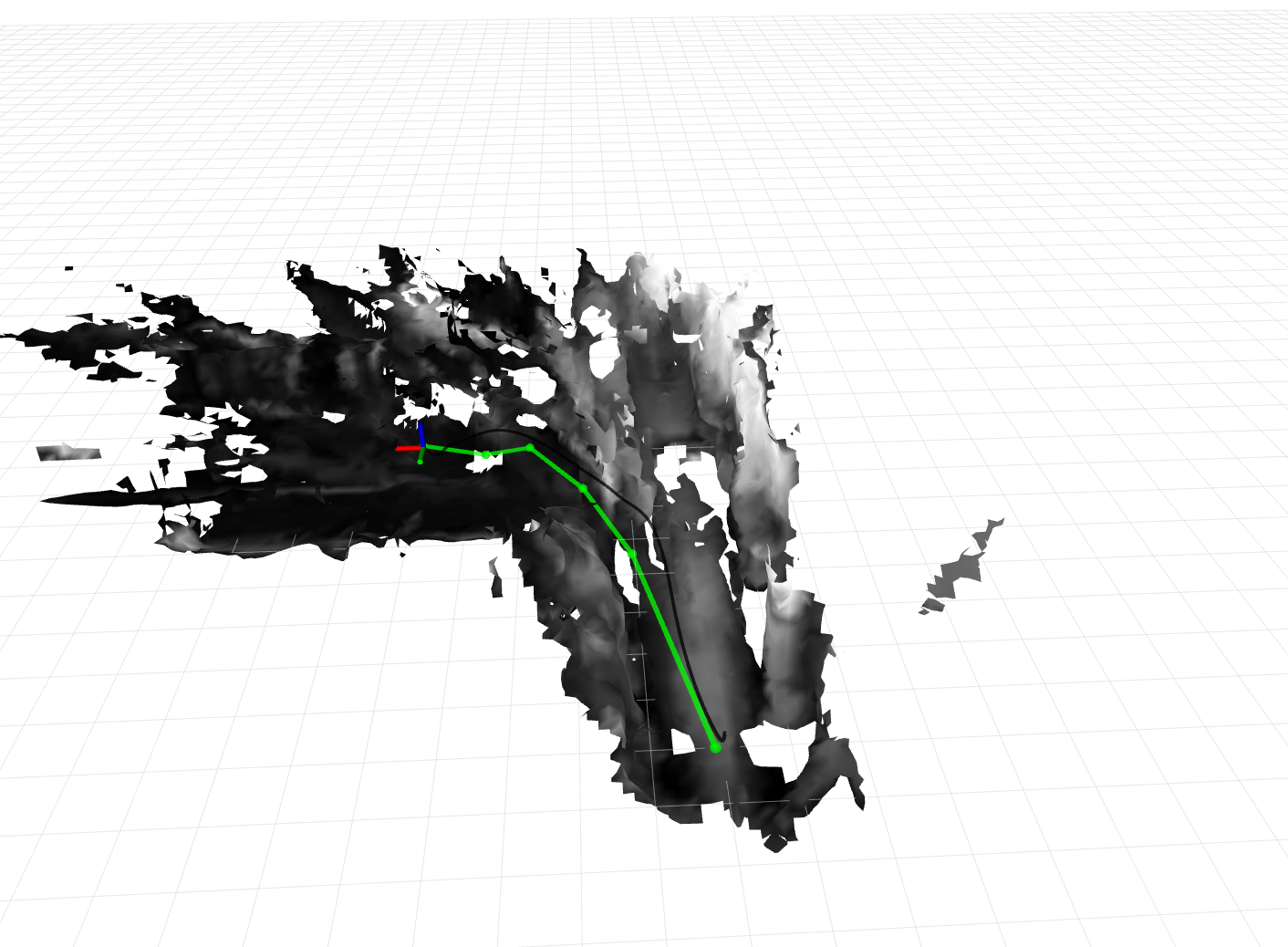}
    \caption{Exp. 5}
    \label{fig:exp5}
  \end{subfigure}
    \begin{subfigure}[b]{0.31\columnwidth}
    \includegraphics[width=1.0\columnwidth,trim=0 00 240 120 mm, clip=true]{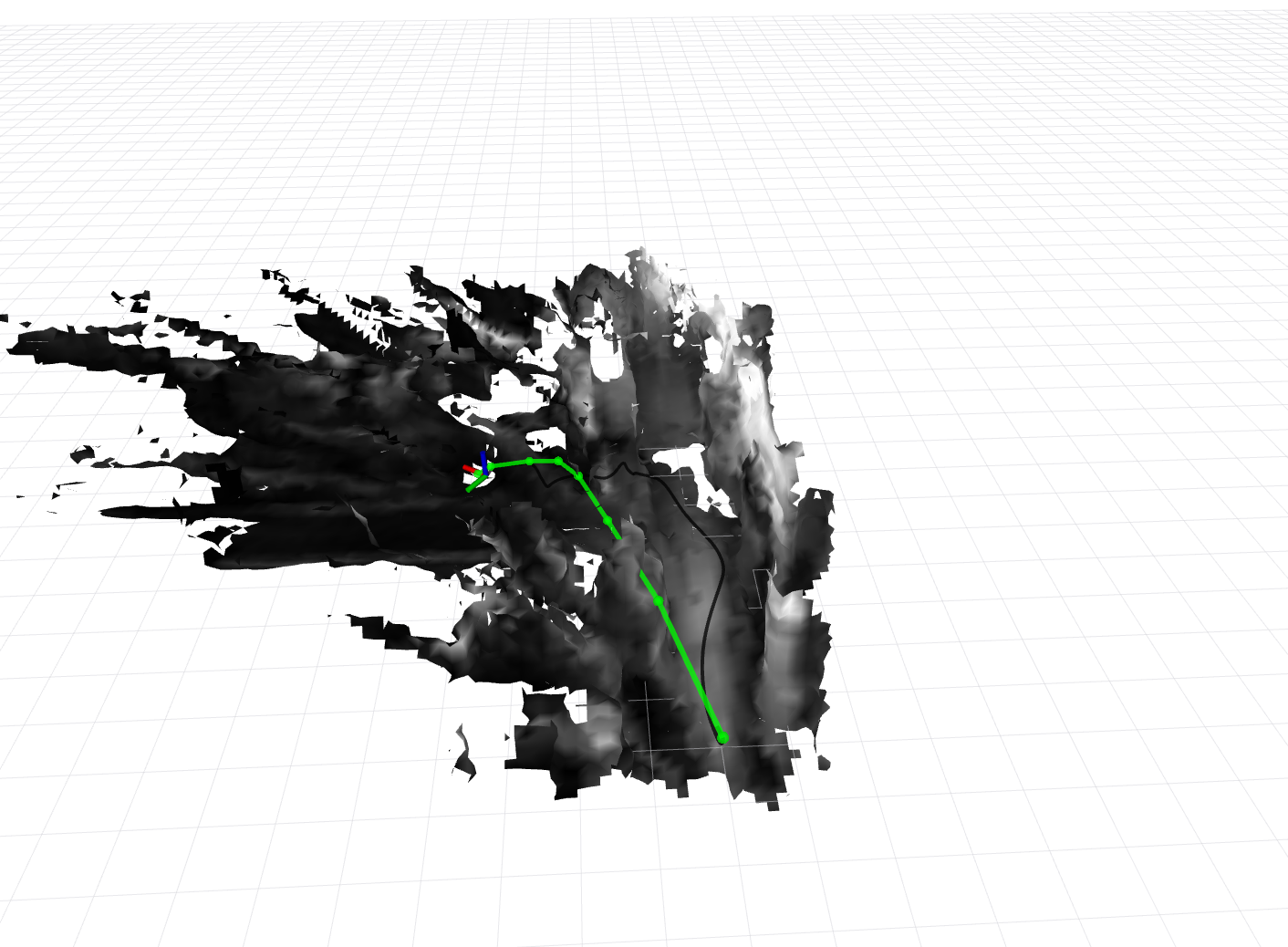}
    \caption{Exp. 6}
    \label{fig:exp6}
  \end{subfigure}
  \caption{Screenshots of the on-board map and planned paths (the local planner in black, global planner in green). The start and end locations are the same for all 3 experiments in each location. Note how in experiments 2 and 6, the global planner path goes directly through an obstacle (as it is not present in the global map), but the local planner is able to replan and recover.}
  \label{fig:experiment_results}
\end{figure}

A table describing all of the experiments is shown in \reftab{table:experiments}, and screenshots of the final paths flown are shown in \reffig{fig:experiment_results}.
As with all live experiments, they are best viewed on video at \url{http://youtu.be/xQTokwr1le0}.
The timings for each component in Experiment 1 are given in \reftab{table:timings}, averaged over the entire experiment.
Note that we use 15 cm voxels a 4 meter clear sphere, which contributes to the long update time of the ESDF integration.

\begin{table}[tbp]
  \centering
  \begin{tabular}{lr}
    \toprule
    \textbf{Step} & \textbf{Time} [ms] \\
    \midrule[1.5pt]
    TSDF Integration & 2.0\\
    ESDF Clear Radius & 5.2\\
    ESDF Integration & 97.5\\
    Shotgun Goal Search & 9.8 \\
    Loco Solve Time & 28.3 \\
    \bottomrule
  \end{tabular}
  \caption{Average timings for each step in mapping and planning in Experiment 1 on-board the platform.}
  \label{table:timings}
\end{table}

\begin{table}[tbp]
  \centering
  \begin{tabular}{cllllll}
    \toprule
    \textbf{\#} & \textbf{Location} & \textbf{Local Map} & \textbf{Global Map} & \textbf{Planner Forward} & \textbf{Planner Back} & \textbf{Obstacles} \\
    \midrule[1.5pt]
    1 & Office & New & None & Local & None &  None \\
    2 & Office & New & From Exp. 1  & Global & None & Pallet \\
    3 & Office & New & New & Local & Global & None  \\
    4 & Machine Hall & New & None & Local & None & None  \\
    5 & Machine Hall & New & New & Local & Global  & None \\
    6 & Machine Hall & New & From Exp. 5 & Global & Global & Trash bin \\
    \bottomrule
  \end{tabular}
  \caption{Summary of experimental setup, and which maps were used for local and global planning, and what obstacles were added.}
  \label{table:experiments}
\end{table}

In the spirit of complete honesty with the reader, while most of the experiments presented worked on the first attempt, a few did not.
Some did not succeed due to bugs in the planner (which have since been fixed), but a number failed for other reasons.
Of the 3 experiments that we ran after removing the last bugs, all of which were in the machine hall, One failed on the return flight due to the planner being too conservative around the added trash bin obstacle, and not being able to find its way back (stopping next to the trash bin). 
Two succeeded but with ``loops'' or $360\deg$ turns in the trajectory, causing the safety pilot concern over their ability to control the system in case of an emergency, and doing a voluntary abort.
We believe that these ``loops'' were due to creating too many closely-spaced waypoints, leading to numerical issues in the optimization.

Of all the failures, including those while debugging edge cases in the local planner, there were no collisions, which shows the advantages of a very conservative planning strategy: even if the planned paths are suboptimal for an outside observer or a safety pilot, no paths will ever be sent to the robot that will lead to a collision.

\section{Conclusions}
In this paper we have presented a complete system for mapping and planning with MAVs. The system was designed and implemented with a focus on GPS-denied environments and assuming no communication and therefore suitable for online processing.
Specifically, we focused on the problems of surveying for search and rescue in earthquake-damaged buildings and areas, where the environments are heavily cluttered and no prior map exists, and additionally on industrial inspection scenarios, where a prior map may exist but objects may shift between inspections, and we are required to operate very close to obstacles.
These scenarios led us to investigate \textit{conservative} mapping and planning strategies, which do not allow the robot to go into unknown space before observing it.

In particular, we have presented our findings on performing dense mapping and both local and global planning on-board Micro-Aerial Vehicles with narrow field-of-view vision sensors.
Our contributions focused on discussing practical problems and solutions to issues encountered while attempting to navigate such systems safely in real environments.
We extended our previous work by improving a global planning sparse topology generation algorithm, suggest methods in which our local replanning algorithm can also function for path smoothing, improved the success rate of our local planning, provided a full framework for local replanning that is able to handle both global paths and local waypoints in unknown space, and benchmark a variety of global and local planning methods, both on real and simulated data.
Most importantly, we described not only our mapping and planning approaches, but considerations that must be taken in other parts of the system for this approach to work, such as state estimation and controls, and made all of our code available online and open-source.
Finally, we validated the complete system in a series of platform experiments mirroring the industrial inspection use-case.

\bibliographystyle{apalike}
\bibliography{sources}

\end{document}